\pdfoutput=1

\documentclass[11pt,table]{article}

\usepackage[final]{acl}

\usepackage{times}
\usepackage{latexsym}

\usepackage[T1]{fontenc}

\usepackage[utf8]{inputenc}

\usepackage{microtype}

\usepackage{inconsolata}

\usepackage{graphicx}

\usepackage{booktabs, multirow}
\usepackage{multicol}
\usepackage{lscape}
\usepackage{makecell}
\usepackage{wrapfig}
\usepackage{scalerel,xparse}

\usepackage{caption}[=v1]
\usepackage{subcaption}

\usepackage[compact]{titlesec}
\NewDocumentCommand\lock{}{
    \includegraphics[scale=0.5]{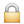}
}

\usepackage[utf8]{inputenc} 
\usepackage[T1]{fontenc}    
\usepackage{hyperref}       
\usepackage{url}            
\usepackage{booktabs}       
\usepackage{amsfonts}       
\usepackage{nicefrac}       
\usepackage{microtype}      
\usepackage{xcolor}         
\usepackage{graphicx}
\usepackage{amsmath}
\usepackage{placeins}

\newcommand{\squishlist}{
 \begin{list}{$\bullet$}
  { \setlength{\itemsep}{0pt}
     \setlength{\parsep}{2pt}
     \setlength{\topsep}{2pt}
     \setlength{\partopsep}{0pt}
     \setlength{\leftmargin}{1em}
     \setlength{\labelwidth}{1em}
     \setlength{\labelsep}{0.4em} } }

\newcommand{\squishend}{
  \end{list}  }

\title{Language Model Council: Democratically Benchmarking Foundation Models on Highly Subjective Tasks}

\author{
\\
 \textbf{Justin Zhao\textsuperscript{1}},
 \textbf{Flor Miriam Plaza-del-Arco\textsuperscript{2}},
 \textbf{Benjamin Genchel\textsuperscript{1}},
 \textbf{Amanda Cercas Curry\textsuperscript{3}}
\\
 \textsuperscript{1}Independent,
 \textsuperscript{2}Bocconi University,
 \textsuperscript{3}CENTAI Institute
\\
 \\
\url{https://llm-council.com}
}

\begin{document}
\maketitle
\begin{abstract}


As Large Language Models (LLMs) continue to evolve, evaluating them remains a persistent challenge. Many recent evaluations use LLMs as judges to score outputs from other LLMs, often relying on a single large model like GPT-4o. However, using a single LLM judge is prone to intra-model bias, and many tasks – such as those related to emotional intelligence, creative writing, and persuasiveness – may be too subjective for a single model to judge fairly. We introduce the \textbf{Language Model Council (LMC)}, where a group of LLMs collaborate to create tests, respond to them, and evaluate each other's responses to produce a ranking in a democratic fashion. Unlike previous approaches that focus on reducing cost or bias by using a panel of smaller models, our work examines the benefits and nuances of a fully inclusive LLM evaluation system. In a detailed case study on emotional intelligence, we deploy a council of 20 recent LLMs to rank each other on open-ended responses to interpersonal conflicts. Our results show that the LMC produces rankings that are more separable and more robust, and through a user study, we show that they are more consistent with human evaluations than any individual LLM judge. Using all LLMs for judging can be costly, however, so we use Monte Carlo simulations and hand-curated sub-councils to study hypothetical council compositions and discuss the value of the incremental LLM judge.




\end{abstract}

\section{Introduction}

\begin{figure}[t]
    \centering
    \includegraphics[width=\linewidth]{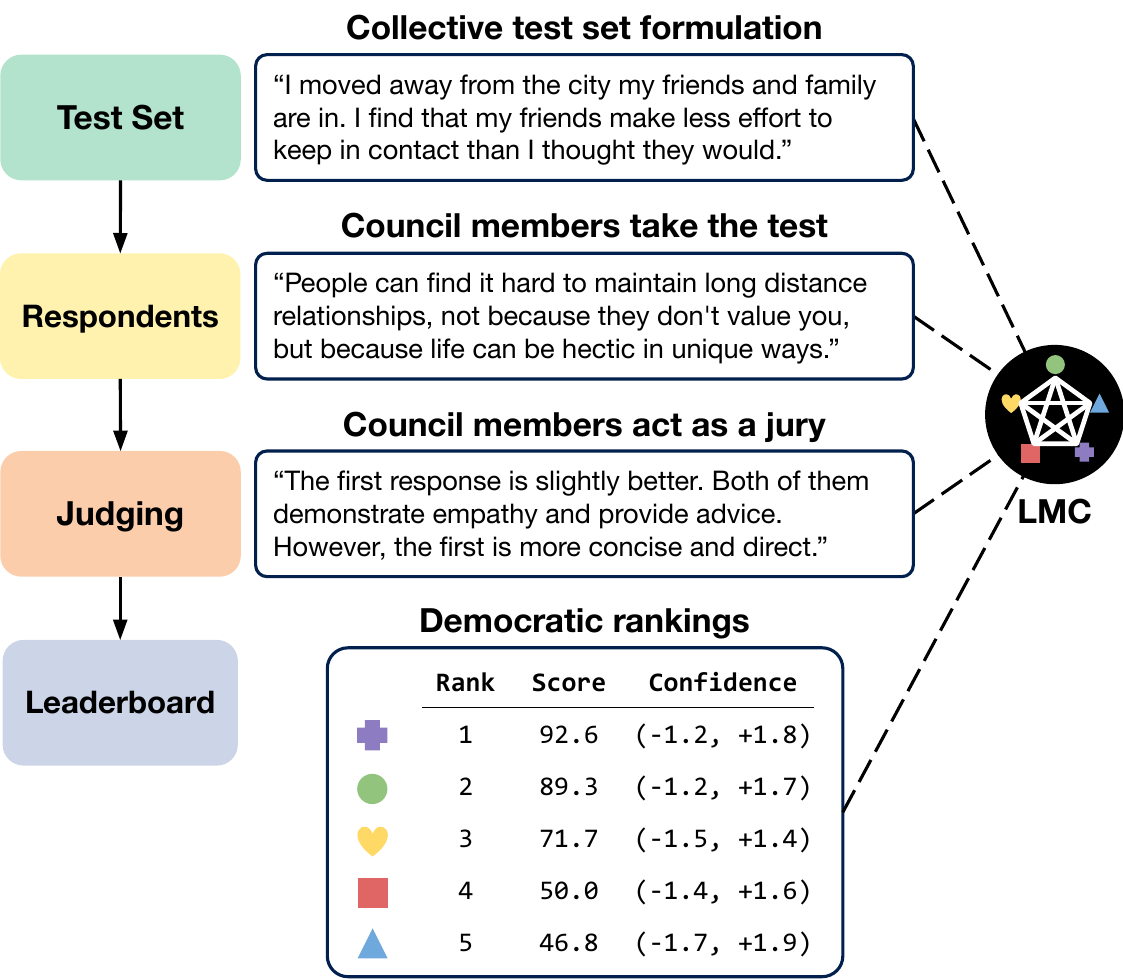}
    \caption{Overview of the Language Model Council (LMC) evaluation framework. By using the same LLMs for test set formulation, task completion, and judging, the framework offers an equitable way to achieve an inclusive, consensus-based ranking.}
    \label{fig:hero}
\end{figure}

As Large Language Models (LLMs) continue to advance, evaluating their outputs remains a significant challenge.
Manual human evaluations are time-consuming and expensive, motivating the need for automatic metrics \cite{novikova2017we, lowe-etal-2017-towards} and evaluation methods \cite[e.g.][]{zheng2024judging, li2024nlgeval, kocmi-federmann-2023-large, shen-etal-2023-large}. 
Conventional model evaluations rely on closed-ended questions that can be checked automatically such as MMLU \cite{5mmluhendrycks2021measuring}. However, these static benchmarks are vulnerable to data contamination \cite{ravaut2024contamination} and are often misaligned with human preferences in real-world, open-ended contexts \cite{3chatbotarenachiang2024chatbot}. For evaluating open-ended responses, automatic metrics like BLEU \cite{papineni-etal-2002-bleu}, ROUGE \cite{lin-2004-rouge}, and BLEURT \cite{sellam-etal-2020-bleurt} are common. These require reference responses that may be expensive to collect and yet may still fail to reflect human preferences beyond a quality threshold \cite{freitag-etal-2020-bleu}.




Arena-based methods enable reference-free evaluation by comparing LLMs in head-to-head matchups. The outcome of a battle between the responses of two LLMs can be evaluated based on objective measures \cite{bianchi2024negotiation}, ratings from strong model judges like GPT-4 \cite{dubois2024alpacafarm, 11arenahard2024}, or human judges \cite{3chatbotarenachiang2024chatbot}, with each outcome contributing to metrics like win rates or ELO
scores \cite{11arenahard2024}.


Strong LLM judges like GPT-4 can match both controlled and crowdsourced human preferences well, achieving over 80\% agreement, the same level of agreement between humans, in a general-domain, arena-based setting \cite{zheng2024judging}.
Unfortunately, it has also been observed that evaluator models tend to have their own biases, for example recognizing and preferring their own outputs over those of other models \cite{dubois2024lcalpacaeval, panickssery2024favor}, which can be mitigated by using a committee of LLM judges \cite[e.g.][]{verga2024poll, zhao2024autoarena}.


In a landscape of diverse LLM judges, can we account for each model's opinions and still establish a definitive ranking amongst them? This paper aims to contribute a unique perspective in the pursuit of a fully decentralized approach to evaluating LLMs. When ranking LLMs on competencies that may be too subjective for a single model to judge fairly, we investigate the dynamics and benefits of an inclusive evaluation network where all models contribute equally to the ranking process.


Inspired by democratic organizations in human society, we introduce the \textbf{Language Model Council (LMC)}, a framework for collective evaluation for a group of LLMs. The LMC operates in three stages:\textbf{ (1)} a test set is formulated with equal participation from all council members, \textbf{(2)} the test set is administered to all council members to
complete, and \textbf{(3)} the responses are evaluated by the council as a collective jury.

Our main contributions are as follows:

\squishlist
    \item[1. ] We propose the LMC, a flexible decentralized evaluation framework that uses LLMs to rank themselves in a democratic manner, and show that our method aligns with human rankings for an open-ended emotional intelligence task.
    \item[2. ] We define and analyze key measures of LLM judging dynamics including: separability, pairwise positional consistency, agreement, and affinity, for the largest ensemble of LLM judges to date.
    \item[3. ] We use Monte Carlo simulations and hand-crafted sub-councils to discuss the value of larger evaluation networks and the value of the incremental judge. 
\squishend

We release all data, code, leaderboard, and web demo at \url{https://llm-council.com}.




\section{Related Work}


Subjective tasks and label variability in humans have received increased attention in the NLP community 
\cite{alm2011subjective,30pdaiPerspectivistData,plank-2022-problem}. 
LLMs also exhibit variability in their outputs as they inherit inconsistencies and biases from human data \cite{hosking2023humannotgold, song2024personality, plaza-del-arco-etal-2024-wisdom, bai2024implicitbias, plazadelarco2024angry, koo2023biases}. 
While LLM judges might not replicate human disagreement patterns exactly \cite{lee2023dissentingvoices, dong2024personalizedllmjudge}, their differences in judgment are increasingly viewed as reflecting valid dissent. Notably, when humans disagreed with GPT-4, they considered its judgments reasonable in 75\% of cases and sometimes revised their own answers \cite{zheng2024judging}.


\textbf{LLM evaluation ensembles.} 
Recent work explores using LLMs as evaluators through structured interactions, committees, weighted voting, and role-playing techniques. Language-Model-as-an-Examiner \cite{bai2024examiner} uses LLMs to interact with candidates through follow-up fact-oriented queries in the knowledge domain. Auto Arena \cite{zhao2024autoarena} proposes an LLM committee to judge competitive multi-turn interactions between LLMs. PRD \cite{li2023prd} allows LLMs to discuss evaluations and assigns higher voting weights based on ability. PRE \cite{chu2024pre} selects a small group of reviewers to produce individual evaluations, then aggregates these evaluations through a chair model. DRPE \cite{wu2023large} uses multi-roleplayer prompting to simulate different roles with one base model, integrating them as votes for the final results. PoLL \cite{verga2024poll} enhances cost-effectiveness by replacing one large judge with multiple smaller judges. 


We build upon existing research by (1) focusing on a highly subjective case study on emotional intelligence where human agreement is inherently low, (2) emphasizing full inclusivity, where each LLM plays an equal role in determining the final rankings, and (3) 
engaging a large ensemble of diverse LLMs to study judging dynamics in greater depth. To our knowledge, this is the largest ensemble of LLM judges studied to date.
\section{Case Study: Using the LMC to Rank LLMs on Emotional Intelligence}
\label{sec:case_study}

The LMC framework consists of three stages: (1) test set formulation, (2) response gathering, and (3) collective judging (Figure \ref{fig:hero}). While the LMC framework is broadly applicable to a wide range of open-ended tasks, this paper presents a focused study on a subjective task of applying emotional intelligence (EI) in interpersonal conflict resolution.

See Appendix \ref{sec:all-prompts} for all prompts used in this case study.

\subsection{Why the EI domain?} 

Unlike objective benchmarks like in coding and math, emotional intelligence (EI) benchmarks are often designed with subjectivity in mind. For example, they often incorporate ratings from a survey of humans for a single ground truth answer \cite{wang2023seceu, sabour2024emobench} and, in the case of multiple-choice questions, enabling multiple correct answers by measuring cosine similarity against a weighted distribution of choices \cite{wang2023seceu, 21eqbenchpaech2023eqbench}. This thematic emphasis on multiple valid viewpoints in current EI benchmarks resonates with the LMC framework, which is itself designed to incorporate multiple LLM perspectives throughout the evaluation development process.

\subsection{Council member selection}

Our selection of LLM council members was guided by several key considerations, including their widespread adoption within the AI community, availability of technical reports, well-supported API access, and performance on benchmarks like MMLU \cite{5mmluhendrycks2021measuring} and Chatbot Arena \cite{3chatbotarenachiang2024chatbot}. We ensure a broad variety of LLMs by including models from \textit{eight} different organizations and \textit{four} countries, with a mix of open-source and closed-source models, small and large (Table \ref{tab:listofcouncilmembers}).

\subsection{Test set formulation}

To create a compelling, open-ended test set for EI, we build upon the EmoBench dataset, a publicly available, hand-crafted, theory-based English dataset designed for EI assessment \cite{sabour2024emobench}. EmoBench consists of 200 emotionally balanced, handcrafted scenarios, e.g., \textit{``Sarah found out that her younger brother is being bullied at school, but he begged her not to tell their parents.''} We solicit the council to expand EmoBench's concise scenarios into richly described dilemmas in the first person (see Figure \ref{fig:prompt-template-synthetic-expansion} for an example). Each of the 20 council members expands five scenarios, resulting in a test set of 100 dilemmas, similar in scale to MT-Bench (80 questions). We manually review all expansions for EI suitability.\footnote{Our manual review resulted in no omissions, though some submitted expansions required minor edits to remove preambles like 'Here is the expanded dilemma...'.}

Relying on a single LLM to generate the entire test set -- even a top performer like GPT-4o -- may introduce bias and limit perspectives. In a survey of 10 human respondents evaluating potential expansions for the test set, 51\% of the preferred expansions were those not authored by GPT-4o.\footnote{Human respondents were asked to choose, in a series of pairwise comparisons, which expanded dilemma would be better for an emotional intelligence test. Each comparison was between a response from GPT-4 and one from a randomly chosen council member (Figure \ref{fig:streamlit_rating_generation_dilemmas}).} Inclusively constructing test examples also mitigates the risk of any single LLM's generative idiosyncrasies \cite{ai4science2023delve} from dominating the test pool. 

An alternative approach would be to have all council members propose expansions for each scenario and select the best through voting. While this may be just as rigorous from a democratic perspective, given the generally high quality of expansions, we opted to use a balanced set of submitted expansions directly.

\subsection{Response gathering}

After expanding 100 dilemmas, each council member responds to every dilemma, yielding 2,000 total responses. To standardize response lengths across council members and preemptively minimize length bias in evaluation, the prompt suggests a 250-word limit (Figure \ref{fig:prompt-template-respond-to-dilemma}). Responses exceeding the limit are truncated at the nearest sentence within the limit.\footnote{Sentence splits are based on standard English end punctuation (. ! ?), as the experiment is conducted in English.} Despite the suggested word limit, some council members consistently generated shorter responses (Figure \ref{tab:mainexperiment}). These responses are left unchanged.

\subsection{Collective judging}
\label{sec:evaluation-settings}


\textbf{Arena-style pairwise comparisons with a single reference model.} We adopt the pairwise comparison setup of Chatbot Arena where responses are compared in head-to-head matchups (see prompt in Figure \ref{fig:prompt-template-sxs-granular-no-tie}). LLM rankings are determined by expected win rates using an ELO scoring system \cite{bai2022training, boubdir2023elo}, with Bradley-Terry (BT) coefficients~\citep{bradley1952rank} applied for improved statistical estimation. Following \cite{11arenahard2024}, confidence intervals are derived through 100 rounds of bootstrapping. Like \cite{dubois2024lcalpacaeval, 11arenahard2024}, we use a single reference model for all pairwise battles. However, instead of GPT-4, we use Qwen-1.5-32B. For details on how the reference model was chosen, refer to Appendix \ref{app:reference_model_selection}.


\paragraph{4-point preference scale.} We query all LLMs with a temperature of 0 and with granular comparison options without ties (A\textgreater{}\textgreater{}B, A\textgreater{}B, B\textgreater{}A, B\textgreater{}\textgreater{}A). We use Chain-of-Thought (CoT) prompting \cite{wei2022cot} to generate discussion before giving judgments. The reasoning behind all of these choices is detailed in Appendix \ref{sec:judge-callibration}.

\paragraph{Exhaustive position-swapping.} To minimize position bias from affecting the final ranking, we adopt a two-game setup, swapping model positions per query, resulting in $100*2=200$ judgments per model per judge. Following the implementation of BT coefficient calculation in the original codebase\footnote{\url{https://github.com/lm-sys/arena-hard-auto}}, inconsistent results after swapping are treated as ties and strong votes are counted as 3 separate wins.

\paragraph{Voting aggregations.} We consider 3 different voting aggregations for consolidating scores across multiple LLM judges on a per-battle basis: \textit{majority vote} (mode of all votes), \textit{mean pooling}\footnote{For mean pooling, we map ratings to a 4-point numeric scale (A\textgreater{}\textgreater{}B: 2, A\textgreater{}B: 1, B\textgreater{}\textgreater{}A: -1, B\textgreater{}\textgreater{}A: -2), take the mean rounded to the nearest whole value, and use the value corresponding to that whole number as the final rating.}, and \textit{no aggregation} (judgments across all battles are equally considered).

\subsection{Characterizing LLM judges}

We leverage the many-to-many interactions between LLMs to define \textit{key judging qualities} of LLMs in an ensemble setting such as the LMC.

\textbf{Separability} measures how confidently models can be distinguished in the final rankings. We adopt this metric from \cite{11arenahard2024}, which defines separability as the percentage of model pairs with non-overlapping confidence intervals, where higher separability indicates better differentiation. The success of score separation depends on three factors: judge discrimination, test discrimination, and participant abilities.  They all contribute, and any one of them could reduce the separation to zero. In the LMC framework, all three are influenced by non-deterministic LLMs, which makes it challenging to isolate which factor contributes most to low separability, for example. Our analysis of separability focuses on LLMs as judges, as the test set, participants, and responses remain constant across different LMC configurations.

\textbf{P}airwise \textbf{P}ositional \textbf{C}onsistency (\textbf{PPC}) measures how often a judge gives consistent results when the order of the two responses in a pairwise comparison is swapped. For instance, if the judge ranks A>B in one comparison and B>A when the positions are swapped (a \textit{rating couplet}), the judge is considered "consistent" because the preference remained the same independent of position. \textbf{Position bias} measures how much the LLM judge favors a specific position (either the first or second) and we define it to be $1 - ppc$. On the 4-point preference scale, a rating couplet is still considered consistent as long as the relative ranking remains consistent overall — fine-grained differences such as (A\textgreater{}\textgreater{}B, B>A) or (B\textgreater{}\textgreater{}A, A>B) are tolerated. Table \ref{tab:consistency} lists all possible couplets and their consistency mappings. \textbf{Conviction} is the raw percentage of strong votes (A\textgreater{}\textgreater{}B or B\textgreater{}\textgreater{}A).

\textbf{Affinity} between a judge and respondent is the score the respondent model receives under the judge’s jurisdiction. \textbf{Self-enhancement bias} is the difference between a model’s affinity to itself and the council’s score for that model. \textbf{Polarization} is the range of the highest and lowest assigned scores. \textbf{Length bias} is the $R^2$ of a linear regression model predicting score from average response length.

\textbf{Agreement} is measured using Cohen’s Kappa \cite{cohenkappamchugh2012interrater} between two judges’ ratings. Similar to position bias, we consider judges in agreement as long as they express the same relative preference, e.g. (A>B and A\textgreater{}\textgreater{}B) or (B>A and B\textgreater{}\textgreater{}) are in agreement. \textbf{Contrarianism} is measured as the disagreement between an LLM and the Council’s majority decision, reported as $1 - \kappa$.\footnote{Cohen's $\kappa$ ranges from -1 to 1. Subtracting from 1 is not particularly interpretable. Applying the negative makes it so that a higher score implies more disagreement and vice versa.}

\subsection{Human study}

To validate the LMC's evaluations, we conduct a human study mirroring that of the LMC's EI test. Human raters are asked to select the better response from a pair presented for each dilemma. The goal is to assess alignment with human preferences in overall ranking, rather than to model the exact distribution of preferences. 


We select nine LLM council members from our pool of 20 to be rated for this study (Figure \ref{fig:human-rankings}).\footnote{The decision to evaluate only nine models was driven solely by budget constraints. Given the cost of human evaluations (€9/hour), the number of models studied had to be limited. However, these nine models were carefully chosen to ensure diversity in model size, openness, and company origin.}
Participants were recruited via crowdsourcing on Prolific.\footnote{\url{https://www.prolific.com/}} A total of 102 participants took part in the study, with each response evaluated by an average of 11 raters, resulting in 1,343 total ratings. Further details on recruitment, quality control, and participant demographics are in Appendix \ref{app_sec:human_study}.

\section{Results and Findings}
\label{sec:results}

\begin{table*}
\centering
\scalebox{0.7}{
\begin{tabular}{l|ccc|ccc}\toprule
\multirow{3}{*}{} &\multicolumn{3}{c}{\multirow{2}{*}{\textbf{As a Respondent}}} &\multicolumn{2}{c}{\multirow{2}{*}{\textbf{As a Judge}}} \\
& & & & & \\\midrule
\textbf{LLM}&\textbf{Rank} &\textbf{Council EI Score} &\textbf{Avg. response length} &\textbf{Separability} &\textbf{Consistency} \\
\midrule
qwen1.5-110B-Chat &1 &65.6 (-1.2, 1.8) &\cellcolor[HTML]{b9d95a}233 &62.1\% &67.6\% \\
gpt-4o-2024-05-13 &2 &59.2 (-1.2, 1.7) &\cellcolor[HTML]{fefdfd}224 &60.5\% &50.8\% \\
gpt-4-turbo-2024-04-09 &3 &57.5 (-1.2, 1.7) &\cellcolor[HTML]{fefaf9}221 &57.9\% &38.5\% \\
gemini-1.0-pro &4 &50.6 (-1.2, 1.5) &\cellcolor[HTML]{e5f1c2}228 &30.5\% &34.8\% \\
claude-3-opus &5 &50.1 (-1.5, 1.4) &\cellcolor[HTML]{e5f1c2}228 &72.6\% &\textbf{74.6\%} \\
qwen1.5-32B-Chat &6 &50.0 (0.0, 0.0) &\cellcolor[HTML]{9eca1c}236 &\underline{25.3\%} &\underline{23.5\%} \\
qwen1.5-72B-Chat &7 &48.7 (-1.4, 1.6) &\cellcolor[HTML]{9eca1c}236 &37.9\% &26.9\% \\
llama-3-70b-chat &8 &45.1 (-1.5, 1.4) &\cellcolor[HTML]{fefdfd}224 &64.2\% &51.1\% \\
claude-3-sonnet &9 &42.5 (-1.5, 1.6) &\cellcolor[HTML]{f7fbeb}226 &52.1\% &39.7\% \\
dbrx-instruct &10 &38.8 (-1.5, 1.9) &\cellcolor[HTML]{b9d95a}233 &50.5\% &44.2\% \\
claude-3-haiku &11 &38.6 (-1.7, 2.2) &\cellcolor[HTML]{b0d446}234 &45.3\% &44.2\% \\
command-r-plus &12 &35.6 (-1.7, 1.7) &\cellcolor[HTML]{fefbfb}222 &61.1\% &52.9\% \\
command-r &13 &34.7 (-1.7, 1.5) &\cellcolor[HTML]{eef6d6}227 &45.8\% &54.5\% \\
mixtral-8x7b &14 &34.4 (-1.4, 1.5) &\cellcolor[HTML]{b9d95a}233 &56.8\% &58.6\% \\
mistral-large &15 &33.9 (-1.5, 1.3) &\cellcolor[HTML]{fbeae9}208 &\textbf{73.7\%} &72.5\% \\
llama-3-8b-chat &16 &30.0 (-1.4, 1.4) &\cellcolor[HTML]{fae9e8}207 &31.1\% &26.1\% \\
mistral-medium &17 &29.3 (-1.6, 1.5) &\cellcolor[HTML]{f5cfcc}185 &57.9\% &59.0\% \\
gpt-4-0613 &18 &26.9 (-1.4, 1.4) &\cellcolor[HTML]{f3c1bc}173 &64.7\% &53.6\% \\
gpt-3.5-turbo-0125 &19 &18.2 (-1.1, 1.1) &\cellcolor[HTML]{f6d1ce}187 &55.8\% &57.7\% \\
gemini-1.5-pro &20 &11.6 (-0.9, 0.8) &\cellcolor[HTML]{e67c73}115 &60.0\% &52.3\% \\
\midrule
Average Judge & & & &\textit{53.3\%} &\textit{49.2\%} \\
\midrule
LMC (majority vote) & & & &73.7\% &\cellcolor[HTML]{9eca1c}\textbf{75.3\%} \\
LMC (mean pooling) & & & &74.7\% &68.5\% \\
LMC (no aggregation) & & & &\cellcolor[HTML]{9eca1c}\textbf{90.5\%} &52.3\% \\
\bottomrule
\end{tabular}
}
\caption{The LMC promotes equal participation as respondents and judges. The Council EI rank and scores are derived from the “council (no aggregation) setting,” where ratings from all LLMs are tallied equally, without aggregation or modification. Under various aggregation algorithms, the council is more separable and more consistent than individual LLM judges.}\label{tab:mainexperiment}
\end{table*}

Table \ref{tab:mainexperiment} presents the main results of our LMC EI case study, with key insights summarized below.

\paragraph{
Qwen-110B outranks GPT-4o in an unexpected upset.} Like other benchmarks, larger models within the same family tend to outrank their smaller or older versions. However, unlike other benchmarks, Qwen-1.5-110B (\#20 on Chatbot Arena) scores highest on our EI task, followed by GPT-4o (\#1 on Chatbot Arena).\footnote{The scores and rankings referened for Chatbot Arena were those as of May 2024. The Qwen-1.5 models have been dropped from Chatbot Area since the Qwen-2.5 family of models have been added.} This is surprising, as Qwen-1.5-110B does not typically outperform GPT-4o. One possible reason for this outcome is the use of Qwen-1.5-32B as the reference model (Appendix \ref{app:reference_model_selection}). Because all of Qwen-1.5-110B's responses are compared to the responses of a strictly smaller variant of the same family, this could result in an outsized advantage for Qwen-1.5-110B in the evaluation overall. This raises an interesting possibility of \textit{successor bias} in arena-style evaluations: the choice of reference model may inadvertently favor its successors in the same arena.



\paragraph{Judges prioritize actionability, clarity, and structure when expressing preferences.}
Using chain-of-thought (CoT) prompting, judges provided detailed reasoning for their preferences. We analyzed 1,000 reasoning traces and identified common themes (Appendix \ref{sec:qualitativeanalysis}). 



\paragraph{
Judges disfavor models that produce responses significantly shorter than the suggested word limit.} Despite a suggested 250-word limit, some models generated much shorter responses, even though decoding parameters allowed for longer output. Models that adhered to the limit, using 220+ words on average, performed better, while all of the models in the bottom four positions averaged less than 200 words. Notably, Gemini-1.5-pro placed last with an average response length of just 115 words, far worse than its predecessor Gemini-1.0-pro (4th place) with 228 words on average. LLM judges bias towards longer responses, but if we exclude the models that went well under the limit, length bias becomes insignificant (Table \ref{tab:lengthbiascorrected}).


\begin{figure}[ht]
    \centering
    \includegraphics[width=1.0\linewidth]{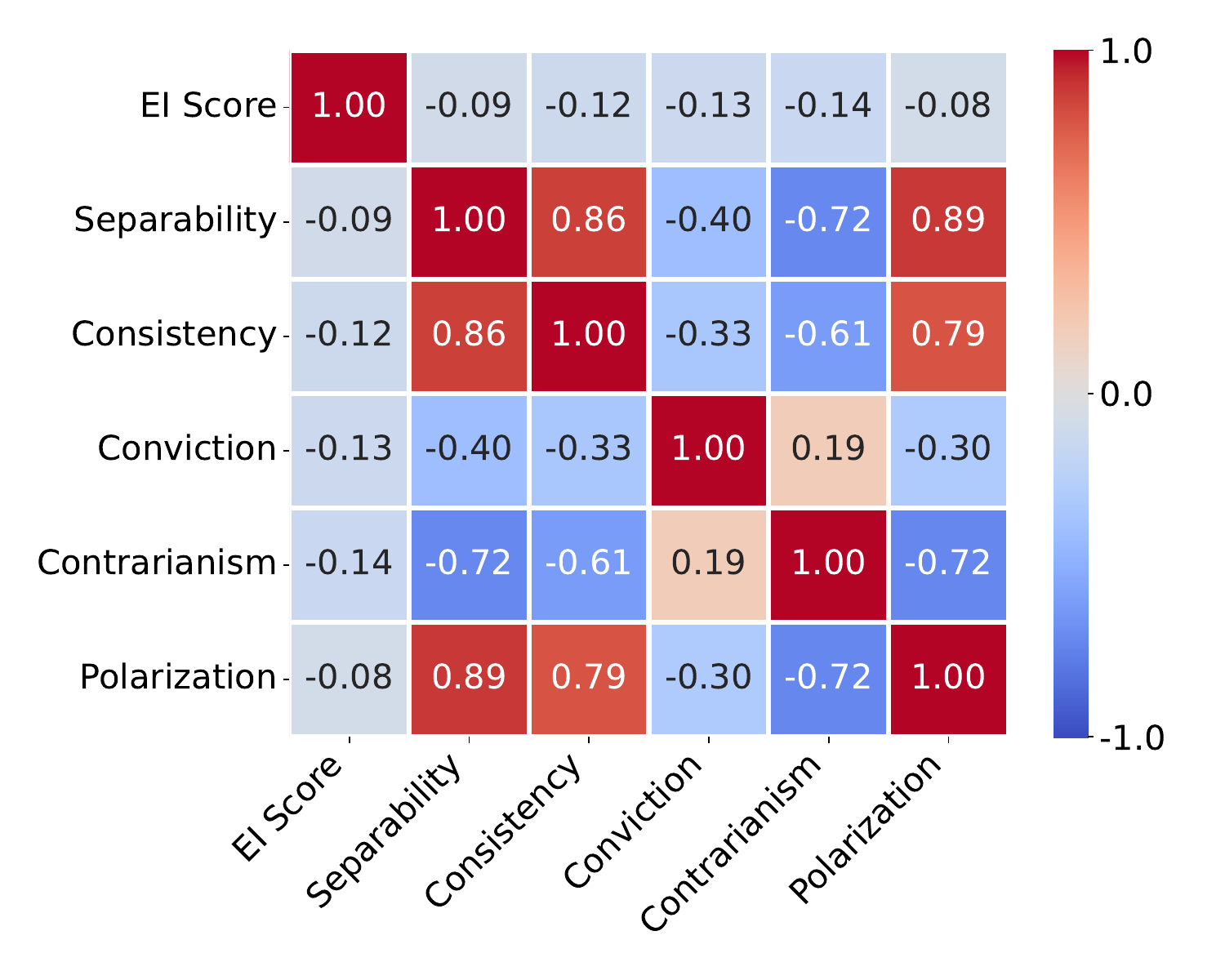}
    \vspace{-10mm}
    \caption{Spearman correlation between EI score and key judging qualities across 20 LLM council members.}
    \label{fig:keyjudgingqualitiescorr}
\end{figure}

\paragraph{
LLM success in the EI task does not correlate with its judging ability.} Performance on the EI task had only a weak correlation with any of the key judging qualities (Figure \ref{fig:keyjudgingqualitiescorr}), which suggests that the ability to perform well in the task and the ability to judge others’ responses are distinct skills.

\paragraph{
Consistent judging, neutral voting patterns, and lower contrarianism correlate with higher separability.} Judges with more consistent votes and neutral voting patterns tend to achieve higher separability. Judges with high conviction — those that express strong preferences (A\textgreater{}\textgreater{}B or B\textgreater{}\textgreater{}A) frequently — are negatively correlated with separability, suggesting that overly strong votes tend to introduce noise. For example, claude-3-opus, despite expressing strong preferences only 0.1\% of the time, achieves the second-highest separability (72.6\%). Higher contrarianism, which measures how often a judge disagrees with the majority decision, also correlates with lower separability. Qwen-1.5-72B, the most contrarian judge, has only 38.9\% separability.

\paragraph{LLMs exhibit slight self-bias, but this effect is mitigated within the full Council.} 
Out of the 20 LLMs in our study, 12 exhibited positive self-enhancement bias, meaning they rated themselves more favorably than the Council's overall score for them. When self-graded battles are excluded, the overall rankings remain similar (Figure \ref{fig:council_rankings_with_and_without_self_grading}). This indicates that while individual models carry self-enhancement bias, the ensembling of the LMC effectively neutralizes these biases.

\begin{figure}
    \includegraphics[width=\linewidth]{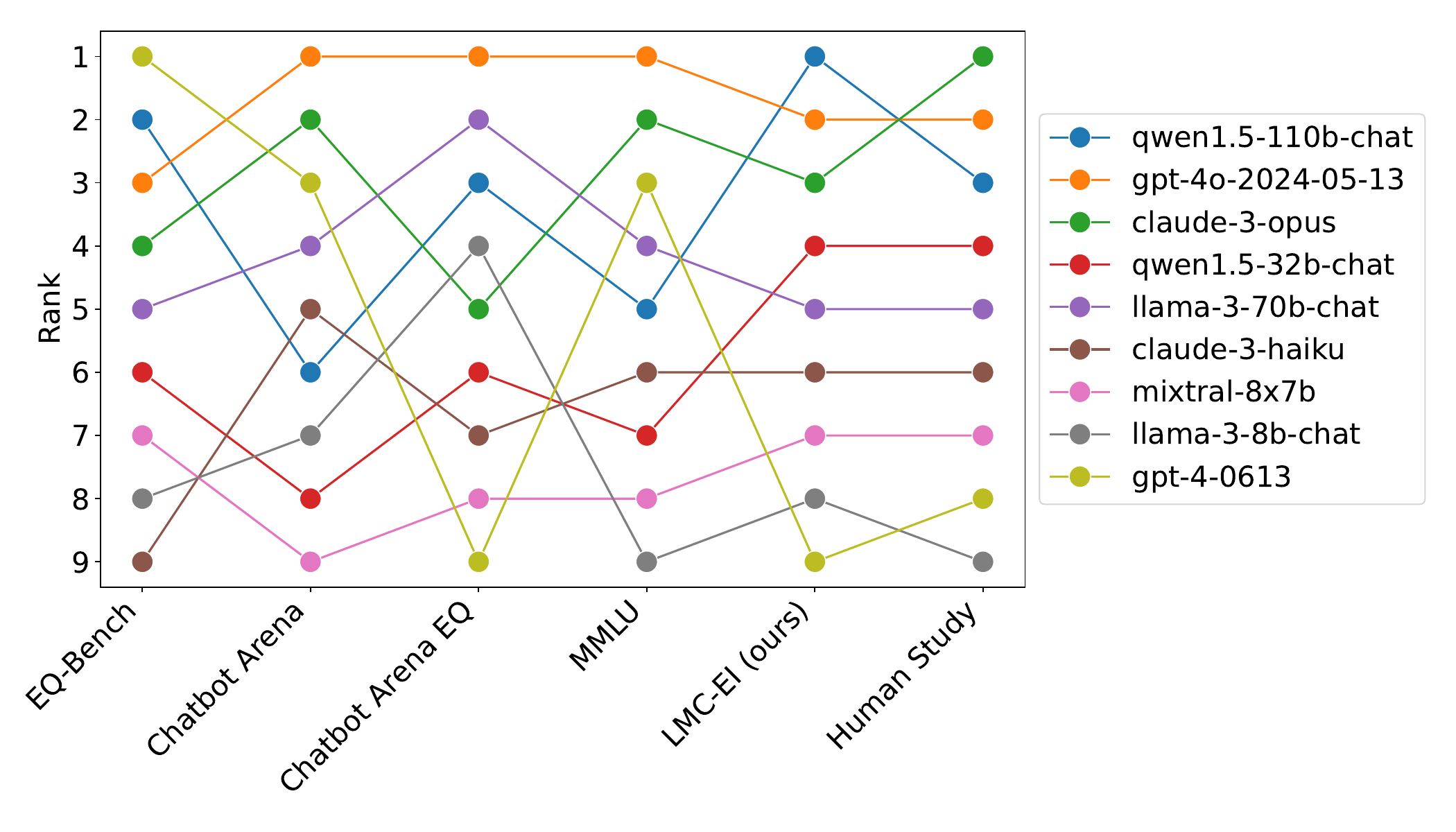}
    \caption{LLM rankings from different benchmarks.}
    \label{fig:human-rankings}
\end{figure}

\paragraph{
Agreement among LMC members, among humans, and between the LMC and humans is similar.} Figure \ref{fig:human-rankings} shows that the rankings produced by LMC members and humans are consistent, with small variations. Both groups agree on the top-performing models and the lowest-ranked ones. The level of agreement between the LMC and humans is roughly the same as the agreement within human evaluators (51.9\%).


\begin{table}
    \centering
        \centering
        \scriptsize
        \begin{tabular}{c|cccc}
        \toprule
        & \textbf{Human} & \textbf{GPT-4o} & \textbf{LMC-A} & \textbf{LMC-M} \\
        \midrule
        Human & 51.9\% & 51.4\% & 52.2\% & 54.2\% \\
        GPT-4o & 51.4\% & -- & 60.2\% & 78.6\% \\
        LMC-A & 52.3\% & 60.2\% & 56.4\% & 67.4\% \\
        LMC-M & \textbf{54.2}\% & 56.4\% & 67.4\% & -- \\
        \bottomrule
        \end{tabular}
        \caption{Agreement between humans and the LMC on the LMC's EI task. "C-A" denotes a body of 20 individual LLMs while "C-M" is the Council with majority aggregation.}
        \label{tab:humanagreement}
\end{table}

\begin{figure}[ht]
    \includegraphics[width=\linewidth]{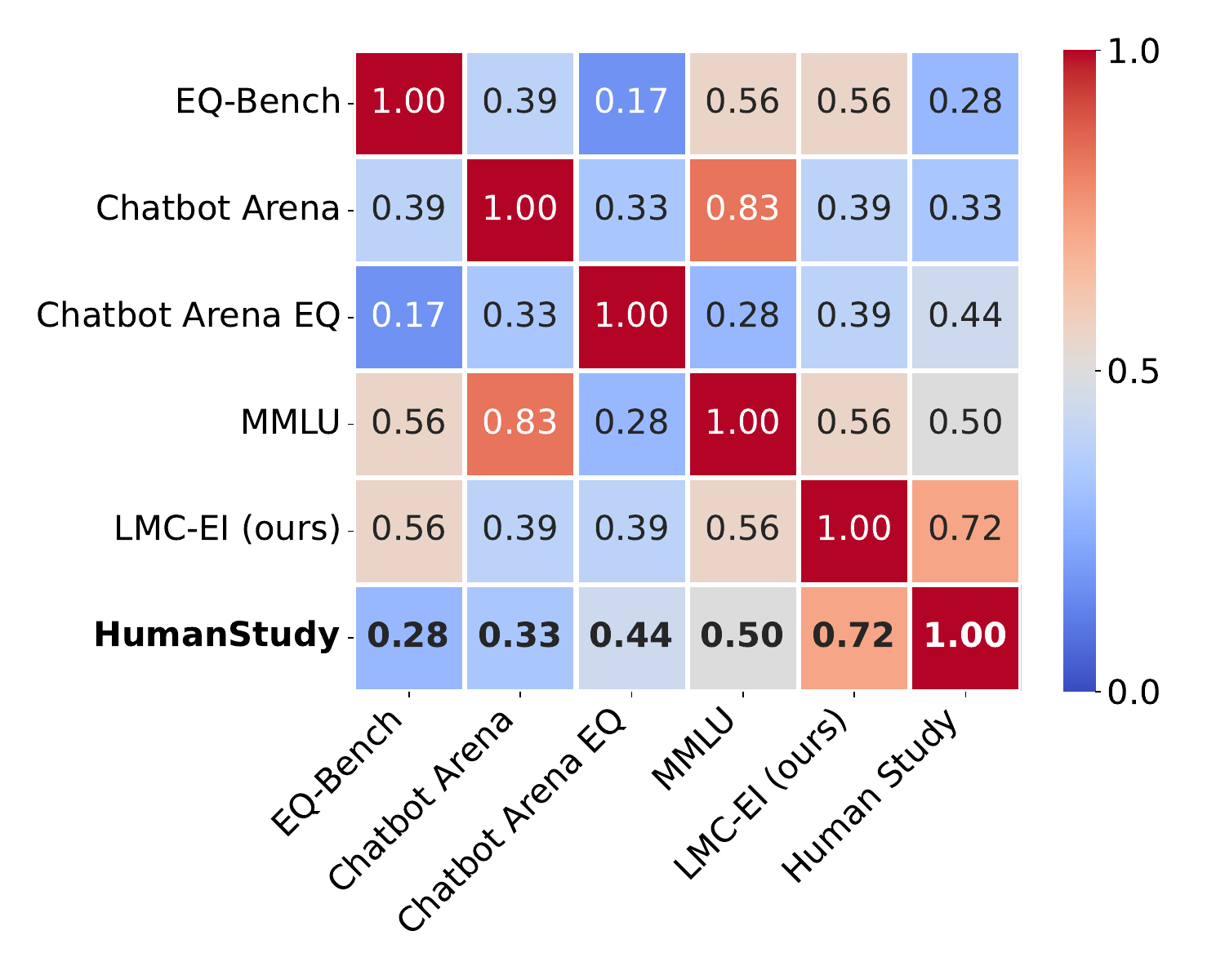}
    \vspace{-8mm}
\caption{Kendall-Tau correlation between benchmark scores and human study scores for nine LLMs (see Appendix \ref{app_sec:human_study}).}
\label{fig:leaderboardcomparison}
\end{figure}

\paragraph{
The LMC’s rankings align more closely with human evaluations than other benchmarks or individual judges.}
The LMC achieves the highest correlation with human-established rankings, outperforming other benchmarks, including those from similar domains like the EI-specific subset of Chatbot Arena (Appendix \ref{app:comparison_to_other_leaderboards_details}) and EQ-Bench \cite{21eqbenchpaech2023eqbench} (Figure \ref{fig:leaderboardcomparison}). 
While a few individual LLM judges, like dbrx-instruct, achieve similar correlation scores (e.g., 0.917), they do so with significantly lower separability (50.5\% vs. 90.5\% for the LMC). This suggests that the LMC’s collective judgment not only aligns more closely with human preferences but also provides clearer distinctions between model performance, making it a more reliable method for ranking models in our case study.

\paragraph{Additional findings.} The 20x20 LLM interactions generate a wealth of data, some of which is not included in the main paper for brevity.  Additional insights can be found in Appendix \ref{app:details_on_main_experiment}.



\section{Discussion}
\label{sec:discussion}


One of the most important questions about using a Language Model Council is whether it is worth the trouble. Collecting more opinions costs more energy, and parsing, tallying, and storing them add logistical costs. What is the value of another opinion, much less \textit{everyone's} opinion?

From a cost perspective, a closely related question is how many examples should be judged in the first place. A smaller test set not only lowers costs but also simplifies the task for any human reviewers — reading and evaluating 10 examples is far more manageable than 10,000.


If we assume that the main goal is to establish good relative rankings, we narrow the focus of quantifying the success of a Council by measuring the \textit{significance} and \textit{stability} of the final ranking. For significance, we look at separability (as in the main experiment), which measures how well models can be distinguished based on non-overlapping confidence intervals. For stability, we create a new metric, \textbf{M}ean \textbf{E}xpected \textbf{R}ank \textbf{V}ariance (\textbf{MERV}), defined as the expected ordinal swing of the average respondent's rank (Appendix \ref{app:merv}). A MERV of 3 means an average respondent's rank is expected to change up to 3 positions in a new trial. Lower MERV indicates that the benchmark has more stable rankings, with MERV of 0 signifying perfect deterministic-like stability.

\subsection{Monte Carlo simulations}

To study the dynamics of separability and stability with differently-sized councils and test sets irrespective of the inclusion of any specific LLM, we use a Monte Carlo procedure to simulate many random hypothetical councils and test sets. Our Monte Carlo simulation procedure is as follows: 

\squishlist
    \item[  (1)] For a given council size $c$ and test set size $t$, randomly sample $c$ LLMs to form a council $C$ and $t$ examples to form a test set $T$. Sampling is performed with replacement.
    \item[  (2)] Find the associated judgments from the main experiment for the specific ($C$, $T$) configuration to determine the scores and relative rankings for all LLMs.
    \item[  (3)] Repeat for 100 trials.\footnote{We use 100 to be consistent with the number of rounds of bootstrapping in the main experiment. For calculating separability in Monte Carlo simulations, the trials themselves are used to bootstrap confidence intervals directly.}
    \item[  (4)] After all 100 trials are complete, tally the results: for stability, observe fluctuations in rankings for each LLM to compute MERV, and for significance, report the mean separability.
\squishend

\begin{figure*}[ht]
    \centering
    \begin{subfigure}[b]{0.24\linewidth}
        \centering
        \includegraphics[width=\linewidth]{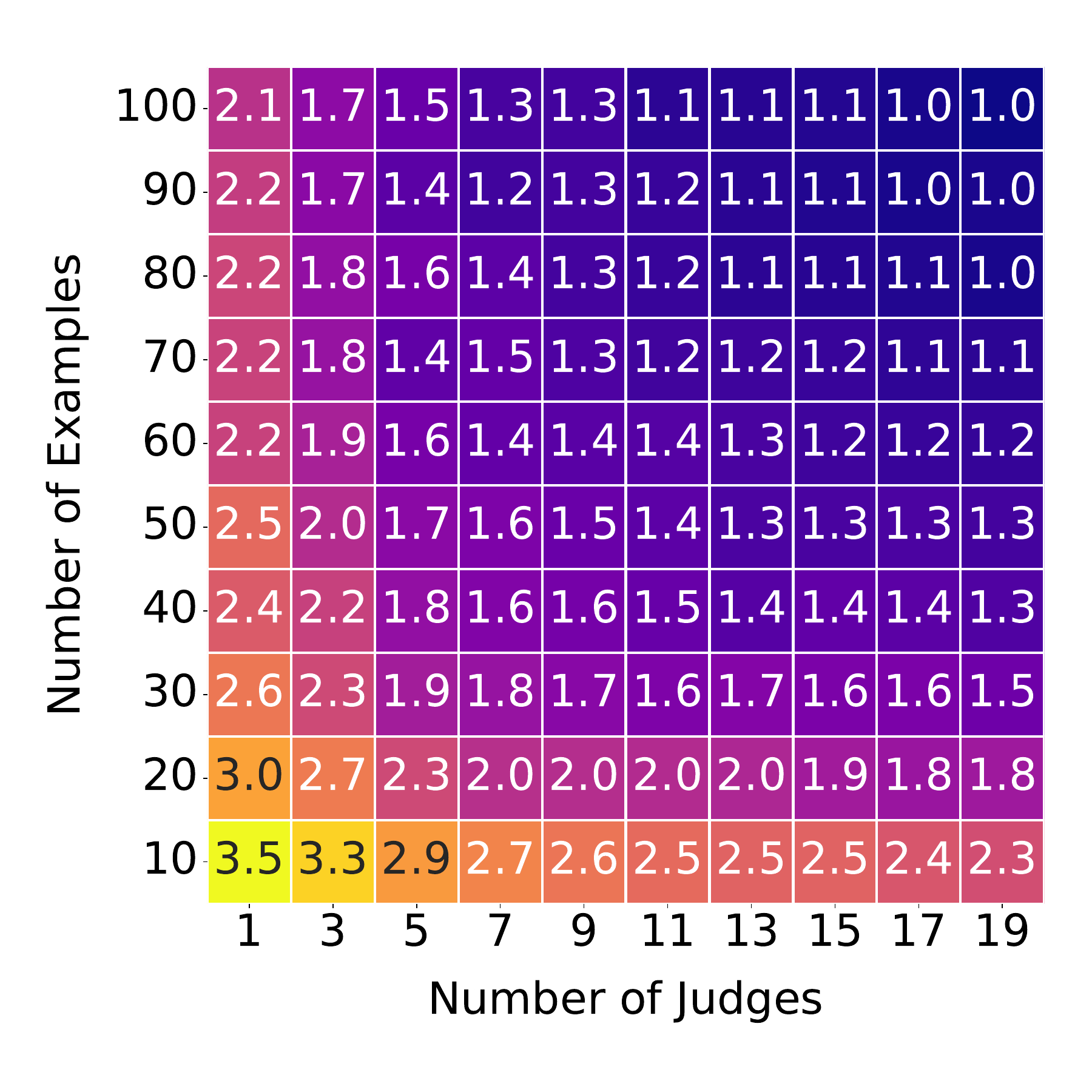}
        \caption{MERV}
        \label{fig:merv_heatmap}
    \end{subfigure}
    \begin{subfigure}[b]{0.24\linewidth}
        \centering
        \includegraphics[width=\linewidth]{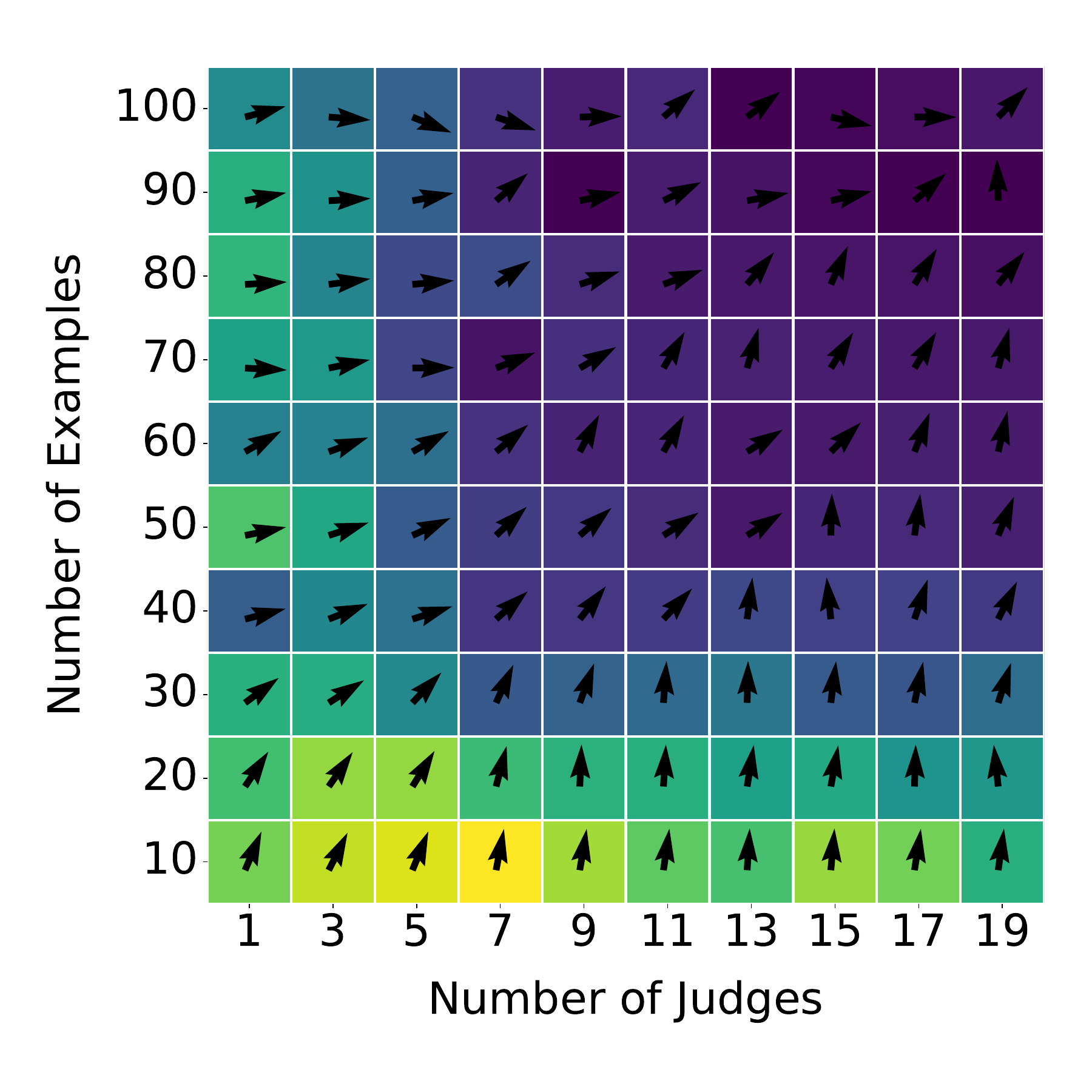}
        \caption{MERV gradients}
        \label{fig:merv_gradient_heatmap}
    \end{subfigure}
    \begin{subfigure}[b]{0.24\linewidth}
        \centering
        \includegraphics[width=\linewidth]{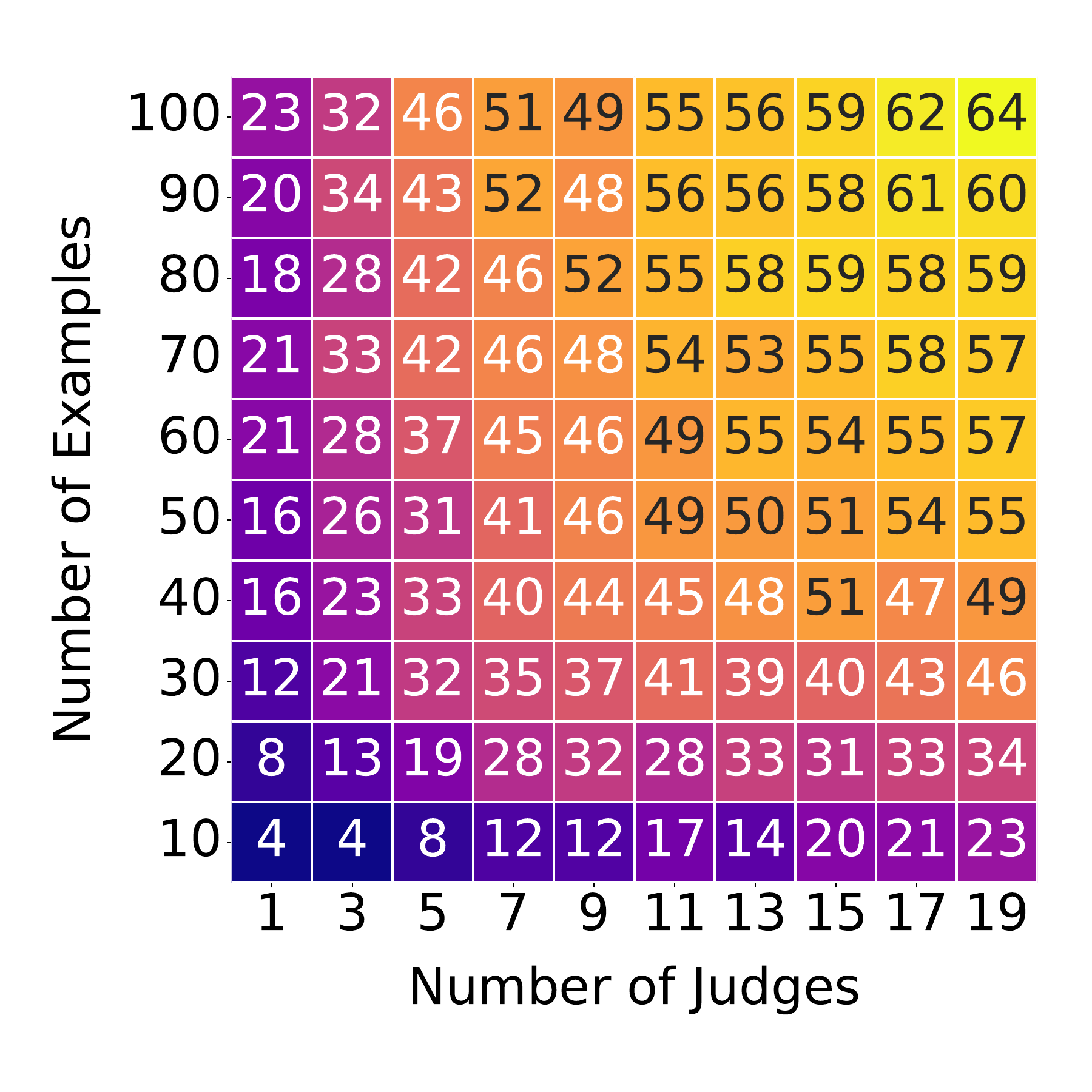}
        \caption{Separability $\mu$}
        \label{fig:separability_heatmap}
    \end{subfigure}
    \begin{subfigure}[b]{0.24\linewidth}
        \centering
        \includegraphics[width=\linewidth]{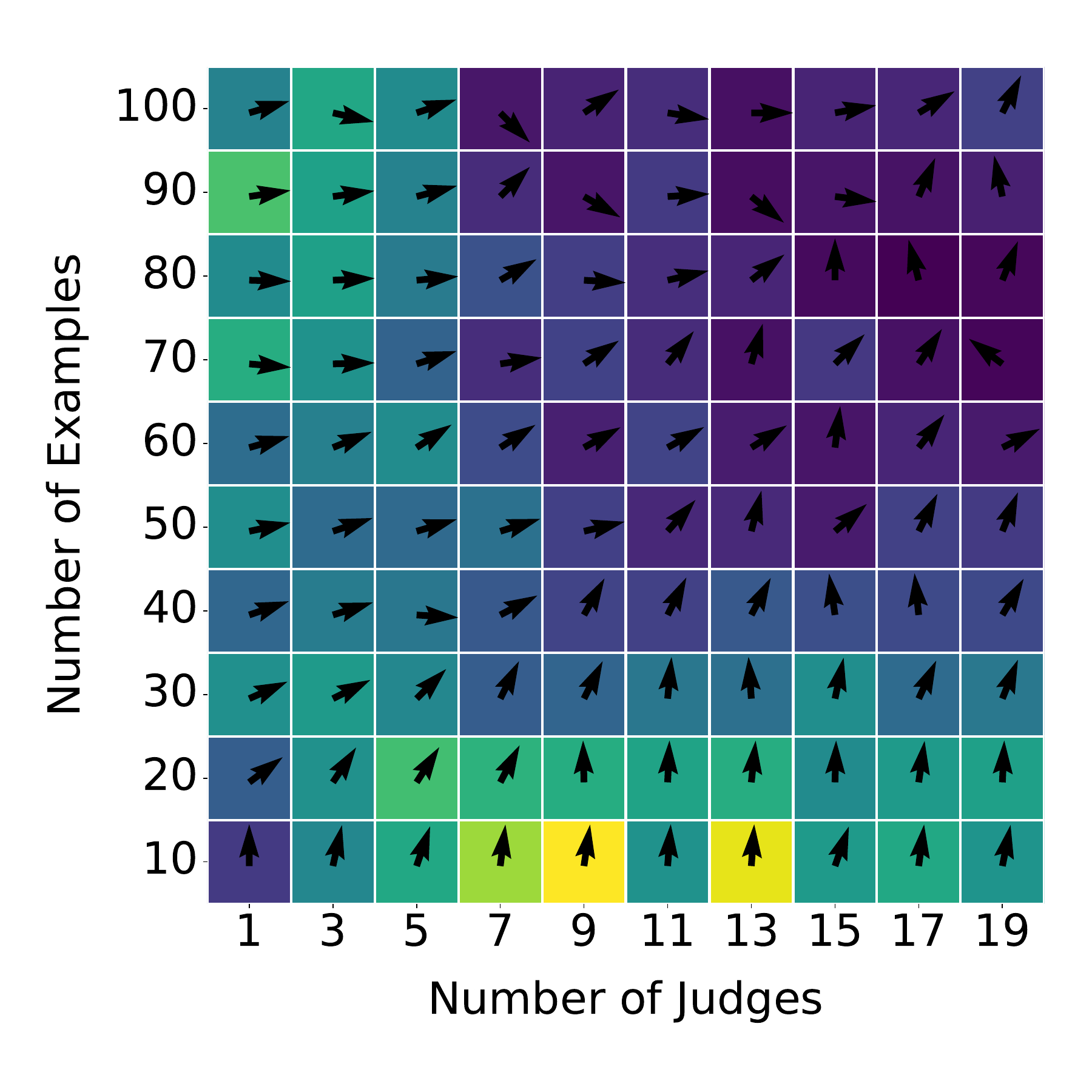}
        \caption{Separability $\mu$ gradients}
        \label{fig:separability_gradient_heatmap}
    \end{subfigure}
    \caption{Measurements of rank stability (MERV) ((a) and (b)) and separability ((c) and (d)) averaged over 100 randomized trials for various numbers of judges and examples. (a) and (c) display raw metric values while (b) and (d) display the gradient magnitude (colors) and direction (arrows). The gradient calculation follows a Manhattan distance approach where row-wise and column-wise gradients are linearly combined to reflect the discreteness of changes between adjacent squares, highlighting the incremental impact of adding another judge or more examples.}
    
    \label{fig:incremental_judge}
\end{figure*}

Figure \ref{fig:incremental_judge} shows results for a sweep $c \in \{1, 3, 5, \dots, 19\}, \quad t \in \{10, 20, 30, \dots, 100\}$.

\paragraph{A nuanced trade-off between the size of the test set and the number of judges.} Both separability and stability improve as the number of test examples and the number of LLM judges increase, with the best scores achieved when both are maximized. However, based on the gradient maps for MERV (Figure \ref{fig:merv_gradient_heatmap}) and separability (Figure \ref{fig:separability_gradient_heatmap}), the added benefit of including either an additional judge or more test examples diminishes significantly in a concentric shape that starts \textasciitilde50 examples and \textasciitilde9 judges. The gradients in this darker zone indicate where the utility of any additional opinion—whether in the form of new test data or new LLM judges—becomes marginal.

\begin{figure}
    \centering
    \includegraphics[width=1.0\linewidth]{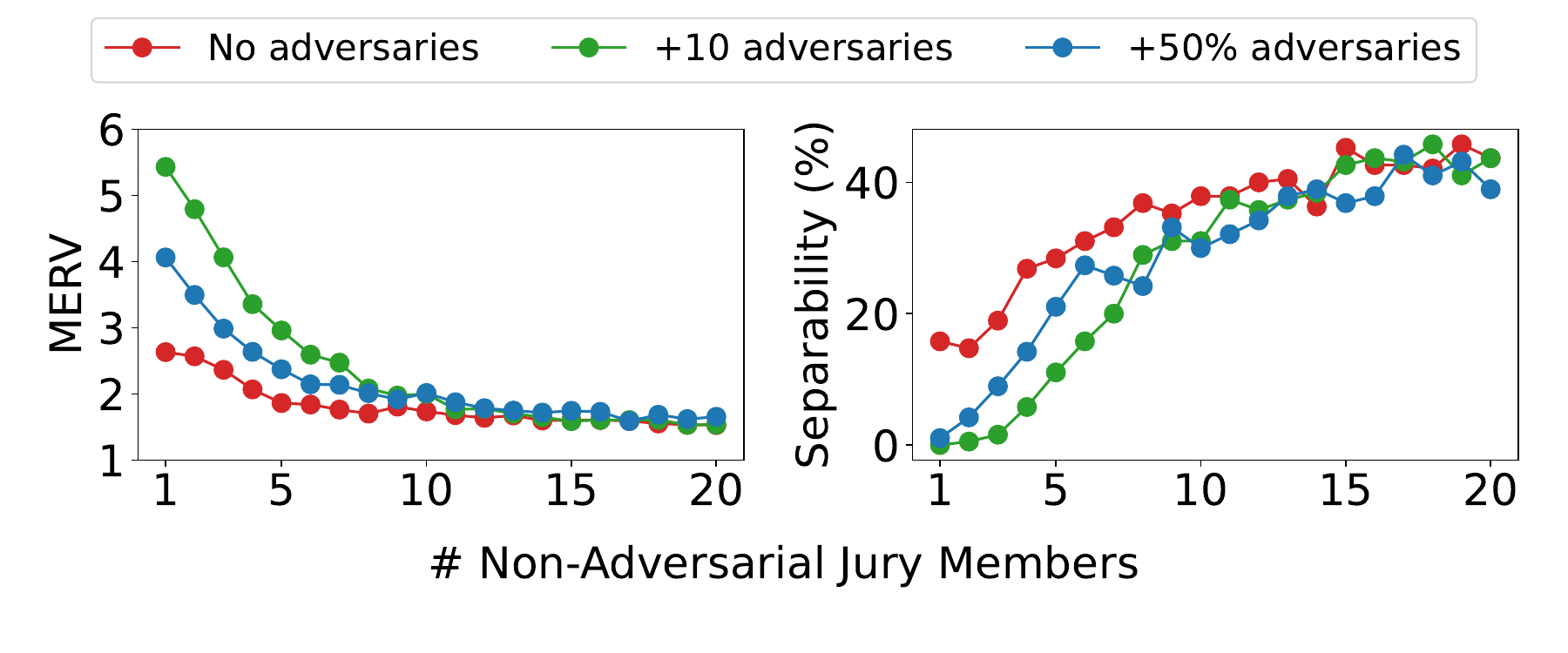}
    \vspace{-8mm}
    \caption{MERV and separability for Monte Carlo simulations for ($t$=30) with adversarial judges.}
    \label{fig:jury_ablation_combined}
\end{figure}

\paragraph{Larger councils are more robust to adversarial judges, with diminishing marginal utility.}

As the size of the LMC grows to include many different LLMs, it may become difficult to verify the quality of every member, particularly on subjective tasks. With the same Monte Carlo procedure, we experiment with a simulation setting with \textit{adversarial judges}. An adversarial judge is a fake LLM council member that returns ratings at random.\footnote{Other adversarial algorithms such as always voting for the first position were not explored.} In Figure \ref{fig:jury_ablation_combined}, we find that on both separability and stability, larger councils reduce the negative impact of adversarial LLM judges. This robustness continues to strengthen as the size of the council grows, even when maintaining the same ratio of adversarial judges to real judges, albeit with diminishing marginal returns.

\paragraph{What is the value of the incremental judge?} With respect to stability and separability, it depends. When test data is scarce, adding more test examples yields greater benefits than increasing the number of judges. However, once the test set exceeds 20-30 examples, introducing an additional judge becomes more valuable. 

Larger councils also demonstrate greater resilience to adversarial judges. As the council size increases, the influence of any single unreliable judge diminishes, reducing the risk of significant disruption to the study. Consequently, strict selection criteria for council members become less critical in larger configurations.





\subsection{Oligarchical councils}
\label{sec:oligarchical_councils}


If a task does benefit from multiple perspectives and the evaluation budget allows for a limited number of opinions, \textit{whose opinions should be included?} Can a subset of judges (or a single judge) effectively represent the fully democratic LMC?


We compare the rankings of three hand-curated sub-councils: \textit{flagships}, \textit{smalls}, and \textit{top-4}.\footnote{We refer to the sub-council as an ``oligarchical'' council because they rely on a small subset of LLM judges to determine rankings, akin to oligarchies in human society.} See Table \ref{fig:oligarchalcouncilmembers} for detailed sub-council membership.



While full council participation yields the highest scores for human agreement and statistical significance, our analysis finds that smaller sub-councils can still produce rankings aligned with human judgments while maintaining strong separability. Notably, \textit{smalls}, a council composed of the smallest LLMs, achieves a separability of 71\%—exceeding the average judge’s 53.3\%—and a Spearman correlation of 0.88 with human rankings, only slightly below the full council’s 0.92 (Figure \ref{fig:leaderboard_comparison_expanded}).

However, council composition remains a crucial factor. For instance, \textit{top-4} achieves the same correlation with human rankings as \textit{smalls} (0.88) but with significantly higher separability (79\%). While smaller councils can perform well, considerations such as separability, robustness, human alignment, and bias mitigation should be carefully weighed in council design.

\begin{figure}[ht]
    \includegraphics[width=\linewidth]{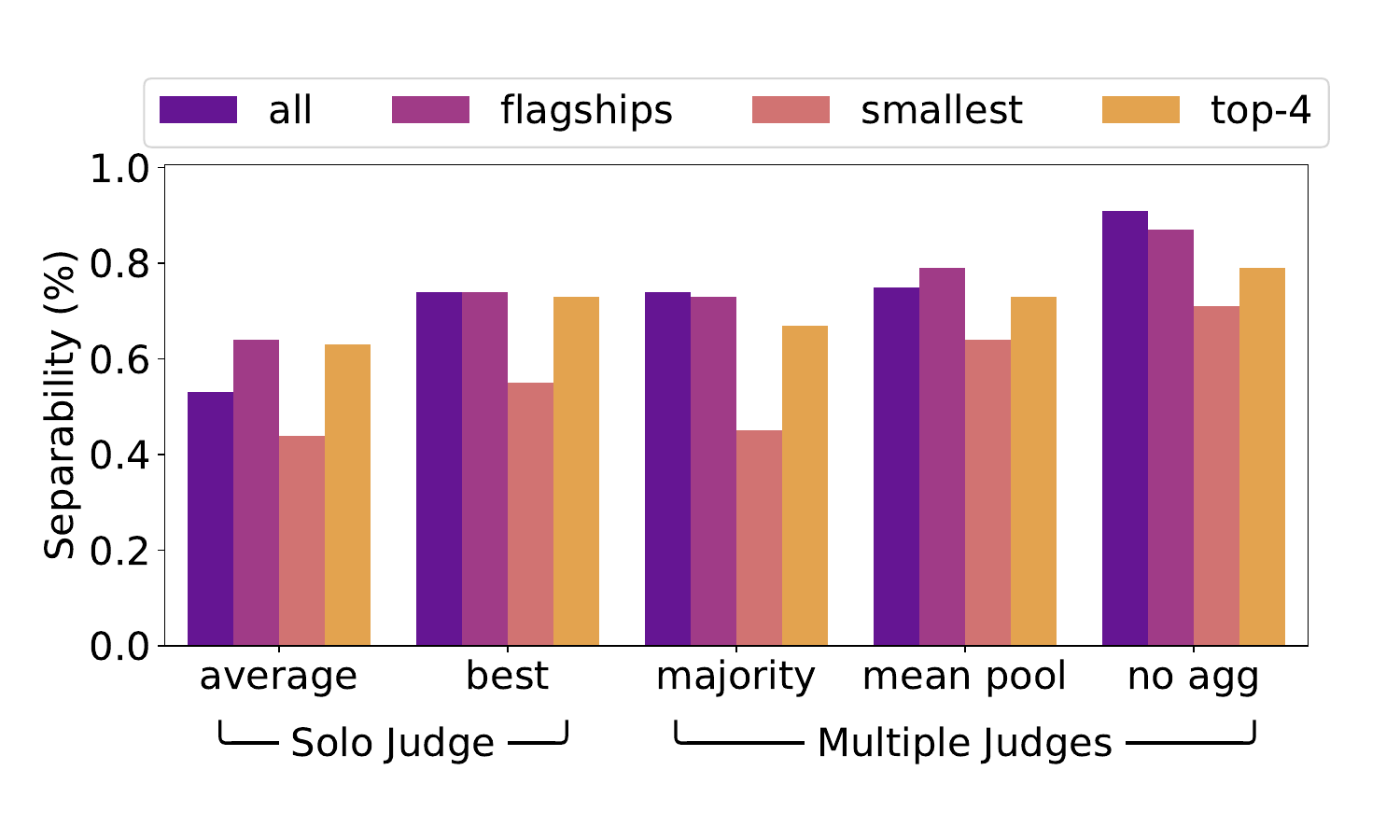}
    \vspace{-8mm}
\caption{Separability scores achieved by different council compositions and aggregation methods. Higher separability is better.}
\label{fig:oligarchicalseparability}
\end{figure}
\section{Conclusion}

In this paper, we introduce the Language Model Council (LMC), a flexible, decentralized evaluation framework for ranking LLM agents through democratic participation. Applying the LMC to an emotional intelligence task with 20 LLMs, we demonstrate that the LMC can produce highly separable rankings that align more closely with human judgments than other benchmarks or individual judges. Through both Monte Carlo simulations and hand-curated sub-councils, we find that while larger councils provide benefits, they are incrementally diminishing, and the majority of key qualities—such as ranking significance, stability, and alignment with human evaluations—can be achieved with a smaller, well-chosen ensemble of judges. As humans increasingly rely on LLMs to evaluate other LLMs, we hope the LMC framework, along with insights from our case study, offers a valuable foundation for developing reliable yet inclusive LLM evaluations, even for highly subjective tasks.

\section*{Limitations}
\label{sec:limitations}

\paragraph{Generalizability of the LMC.} Although we present one detailed case study focused on EI, the Language Model Council (LMC) framework is broadly applicable to a wide range of open-ended tasks. The core mechanism of tallying preferences through arena-style pairwise comparisons is inherently adaptable to various types of prompts and tasks \cite{3chatbotarenachiang2024chatbot}. However, framing new subjective tasks, such as those related to aesthetics or politics, in a form suitable for technical evaluation still requires careful design. In our case study, we were responsible for the technical formulation of the EI task, and the task examples were seeded from EmoBench \cite{sabour2024emobench}, a human-crafted dataset that we chose.

The least generalizable aspect of the LMC is likely the first step: formulating the test set in a collaborative way. For tasks with fixed or human-authored test sets, it may be undesirable or unclear how to implement participation from multiple LLMs. In such cases, this step could be omitted or delegated to a single strong LLM. Whether any LLM can generate meaningful test sets for narrow tasks in a fully unsupervised, domain-generic manner—subjective or otherwise—remains an open area of exploration \cite{huggingface_infinite_dataset_hub}.

\paragraph{Single-turn interactions and English-only evaluation.} Our case study evaluates EI based on single, self-contained interactions and was conducted entirely in English. However, many tasks may be better assessed through extended conversations, multiple sessions, multiple modalities, or in multiple languages, to reflect a broader range of human interaction dynamics, all of which are not explored in this paper.




\paragraph{Reproducibility challenges.} LLMs are inherently stochastic, meaning the same model can produce different ratings even with temperature set to 0 \cite{152334hNondeterminismGPT4}. Reproducibility is further complicated by closed-weight models like GPT-4, which may receive undisclosed updates. All responses in our study were collected in May 2024, but serverless providers, like Together\footnote{\url{https://www.together.ai/}}, may update their APIs or discontinue support for certain models, as happened with Qwen-1.5-32B (replaced by Qwen-2.5). For open source models, changes in deployment hardware or GPU configuration can introduce additional variability, making exact replication of results difficult.

\paragraph{LLM statelessness.} We assume that LLMs, being memory-less, can serve as both respondents and judges simultaneously. In contrast, humans would struggle to judge their own responses impartially due to memory retention. As LLMs evolve to incorporate memory—such as retaining recent prompts and responses \cite{openaimemory}—the risk of self-enhancement bias may increase. If LLMs begin to remember their own responses during evaluation, disabling self-grading may become the default approach to ensure fairness.




\paragraph{Inclusive democracy does not guarantee fairness.} While the LMC effectively neutralizes biases individual LLMs, it is still not immune to systemic biases within the evaluation framework itself. Our case study suggests that using a single reference model may have inadvertently favored its successors within the arena. To mitigate such biases, we recommend conducting dry runs with individual LLM judges to detect and correct mechanical biases in evaluation design before scaling to a full council.

\paragraph{Diversity of opinions in LLMs versus humans.} The question of whether the distribution of an ensemble of LLM judgments is a good approximation for general human diversity \cite{dong2024personalizedllmjudge, hosking2023humannotgold} is secondary to our focus on the alignment of \textit{final rankings}, which is the aggregated expression of the collection of abundantly dissenting opinions (from humans or LLMs), and thus where we assert the utility of the LMC's methodology. In our EI case study, we found that the LMC's final ranking aligned \textit{better} with human preferences compared to other benchmarks and the rankings produced by nearly all individual LLM judges.

We also recognize that the human judgments collected in our study may not fully represent the diversity of opinions within the broader population \cite{elangovan2024considers}, nor do they reflect authentic, first-hand judgments. Emotional responses are highly individual, and real interpersonal conflicts are shaped by personal experiences and social factors that may be difficult for anyone other than the person experiencing the conflict to fully evaluate. For both humans and LLMs, we can only make a deliberate effort to gather judgments from some accessible variety of relevant profiles. This approach mirrors the rationale behind the design of the LMC in our case study, which was similarly formed with variety in mind.

\section*{Acknowledgements}
\label{sec:acknowledgements}


We thank Sahand Sabour for creating EmoBench and insightful discussions about emotionally rich synthetic data. We thank Alex Tamkin for proposing the idea to measure the value of the incremental judge. We thank Mitchell Gordon for his suggestion to qualitatively analyze why certain responses are preferred. We thank David So for his ideas on calibration, repeatability, and model affinity. We thank Federico Bianchi and Sam Paech for their idea of measuring oligarchical councils and for their feedback on prompt design. Sam Paech provided insightful discussions on separability, voting aggregation, length bias, and the relationship to other leaderboards, and he advised us throughout the project. Finally, we thank Predibase for their support and funding this research.

Flor Miriam Plaza-del-Arco is supported by the European Research Council (ERC) under the European Union’s Horizon 2020 research and innovation program (grant agreement No. 949944, INTEGRATOR).
She is a member of the MilaNLP group and the Data and Marketing Insights Unit of the Bocconi Institute for Data Science and Analysis (BIDSA). Amanda Cercas Curry is a former member of MilaNLP and was supported by the same grant while working on this study.

{
\bibliography{custom}
}

\clearpage
\appendix
\clearpage
\section{Additional Findings}
\label{app:details_on_main_experiment}

\begin{table*}[t]
\centering
\scriptsize
\begin{tabular}{lllllll}
\toprule
\textbf{Council Composition} &\textbf{Separability} &\textbf{Conviction} &\textbf{Consistency} &\textbf{Polarization} 
&\textbf{Length bias} \\
\midrule
all &\textbf{0.92 (+0.01)} &0.05 (+0.04) &\textbf{1.0 (+0.48)} &0.81 (+0.27) &0.35 (-0.19) \\
flagships &0.90 (+0.03) &0.04 (+0.03) &\textbf{1.0 (+0.48)} &\textbf{0.85 (+0.22)} &0.29 (-0.17) \\
smalls &0.81 (+0.10) &\textbf{0.11 (-0.21)} &\textbf{1.0 (+0.74)} &0.73 (+0.26) &0.44 (-0.25) \\
top-4 &0.86 (+0.07) &0.04 (+0.03) &\textbf{1.0 (+0.48)} &0.84 (+0.24) &\textbf{0.25 (-0.13)} \\
\bottomrule
\end{tabular}
\caption{Key judging qualities from councils (no aggregation) for hand-selected sub-councils (Table \ref{tab:listofcouncilmembers}) with only positionally consistent votes (positionally inconsistent ratings are filtered out prior to ranking). The value in parentheses represents the change compared to using all votes without a first filtering step.}
\label{tab:council_composition}
\end{table*}

\subsection*{A note on proactively remove inconsistent LLM judge ratings}
In the two-game setup during pairwise comparisons, we gather ratings for models in both positions, allowing us to identify and potentially remove inconsistent ratings before calculating ELO scores.

Automatically removing inconsistent votes affects the weighting of LLM judges, as those with more inconsistent ratings will have fewer votes counted and thus less influence overall. However, this could also be argued as a positive outcome, as positional consistency is often a sign that the judgment was noisy or arbitrary, especially when tie ratings are not permitted. By comparison, some arena-based evaluation systems like \cite{zheng2024judging} allow judges to declare a tie between responses, which are subsequently excluded from ELO scoring.

In our EI case study, we chose to retain all inconsistent votes, allowing downstream processes like aggregation, Bradley-Terry scoring, and bootstrapping to promulgate any diminished influence caused by inconsistent voting.

Since inconsistent votes are a potential source of noise, however, we include metrics when only consistent votes are considered. For the hand-selected sub-councils analyzed in Section \ref{sec:oligarchical_councils}, Table \ref{tab:council_composition} shows the changes in key judging qualities when considering only consistent votes. Table \ref{tab:lengthbiascorrected} shows the impact of inconsistent vote pre-filtering on length bias.

\newpage
\subsection*{Extended judging profiles and references to larger visualizations.}

\begin{figure*}[t]
\centering
\includegraphics[width=\linewidth]{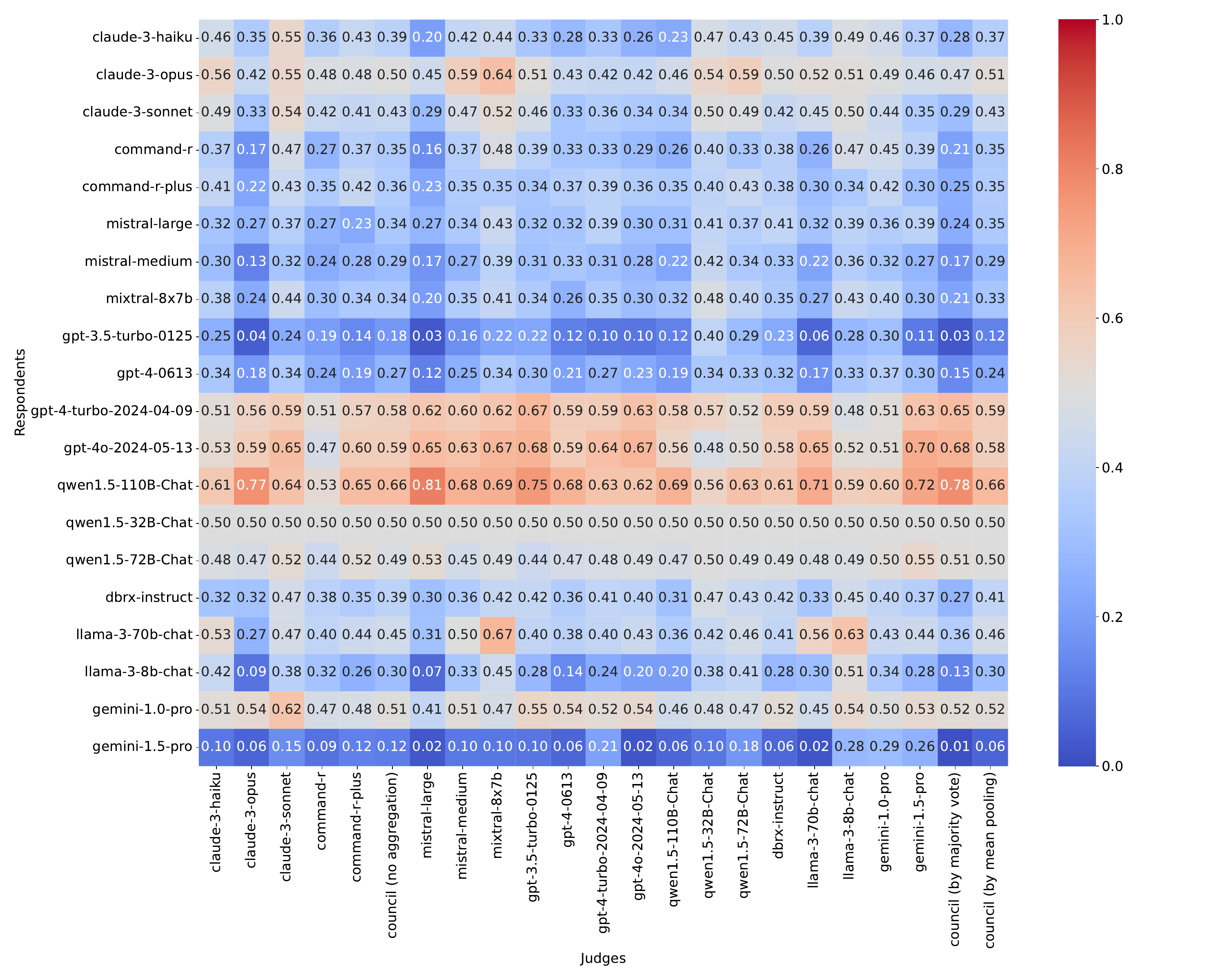}
\caption{Heatmap of the affinities of LLM judges to LLM respondents. The relatively consistent horizontal bands in the heatmap suggest a clear consensus on the preferred LLM participants. Some judges, like mistral-large, exhibit high polarization, with a significant difference between their highest and lowest-rated LLMs. In contrast, judges like llama-3-8b display a narrower range of affinity. The highest affinity expressed by any LLM comes from mistral-large for Qwen-1.5-110B, while the strong horizontal blue band for gemini-1.5-pro indicates it was a consensus low performer.}
\label{fig:raw-affinities}
\end{figure*}

\begin{figure*}[t]
\centering
\includegraphics[width=\linewidth]{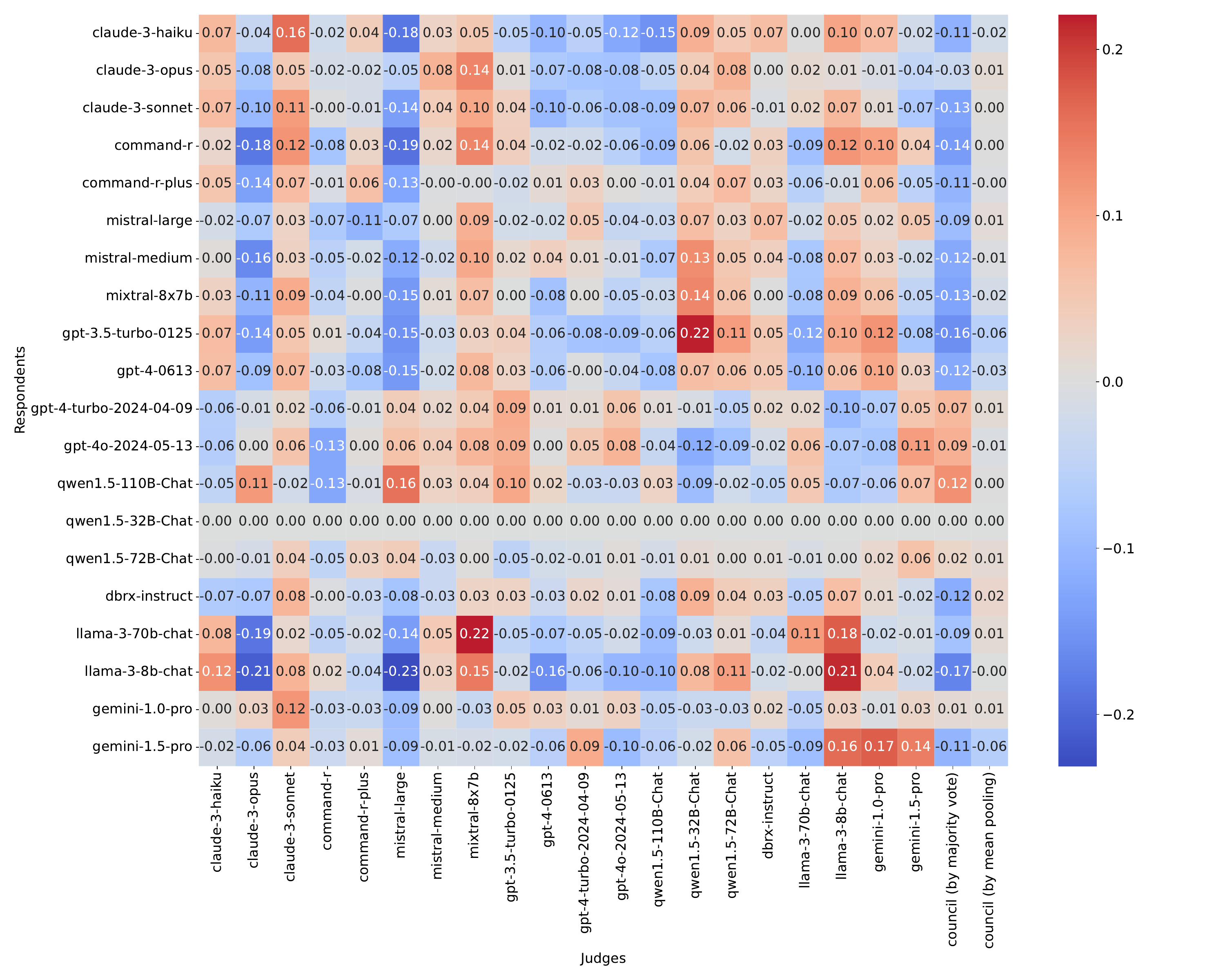}
\caption{This heatmap shows the normalized affinities of LLM judges toward LLM respondents, with the LMC's consensus affinity subtracted out. Self-enhancement bias is now visible along the diagonal, where most LLMs exhibit some level of bias—though not all. Interestingly, six LLMs, including Claude-3-Opus and mistral-large, display negative self-enhancement bias, rating their own responses lower than the council's consensus. For instance, mixtral-8x7b shows a strong preference for llama3-70b’s responses (+0.22 points above the consensus), but this affection is not reciprocated—llama3-70b actually rates mixtral-8x7b -0.08 points below the LMC’s consensus. Mistral-large is a particularly critical judge, rating 15 out of 20 LLMs more harshly than the LMC’s consensus. In contrast, Claude-3-Sonnet is much more favorable, expressing negative affinity for only one LLM, qwen-1.5-110B. The family blocks along the diagonal also reveal patterns of family-enhancing or self-deprecation bias. The llama3 family shows the highest family-enhancing bias, while the mistral family is more divided. Mistral-large and mistral-medium disproportionately rate their fellow family members, including themselves, more negatively, whereas mixtral-8x7b shows positive family bias.}
\label{fig:raw-affinities-council-normalized}
\end{figure*}

\begin{figure*}[t]
\centering
\includegraphics[width=\linewidth]{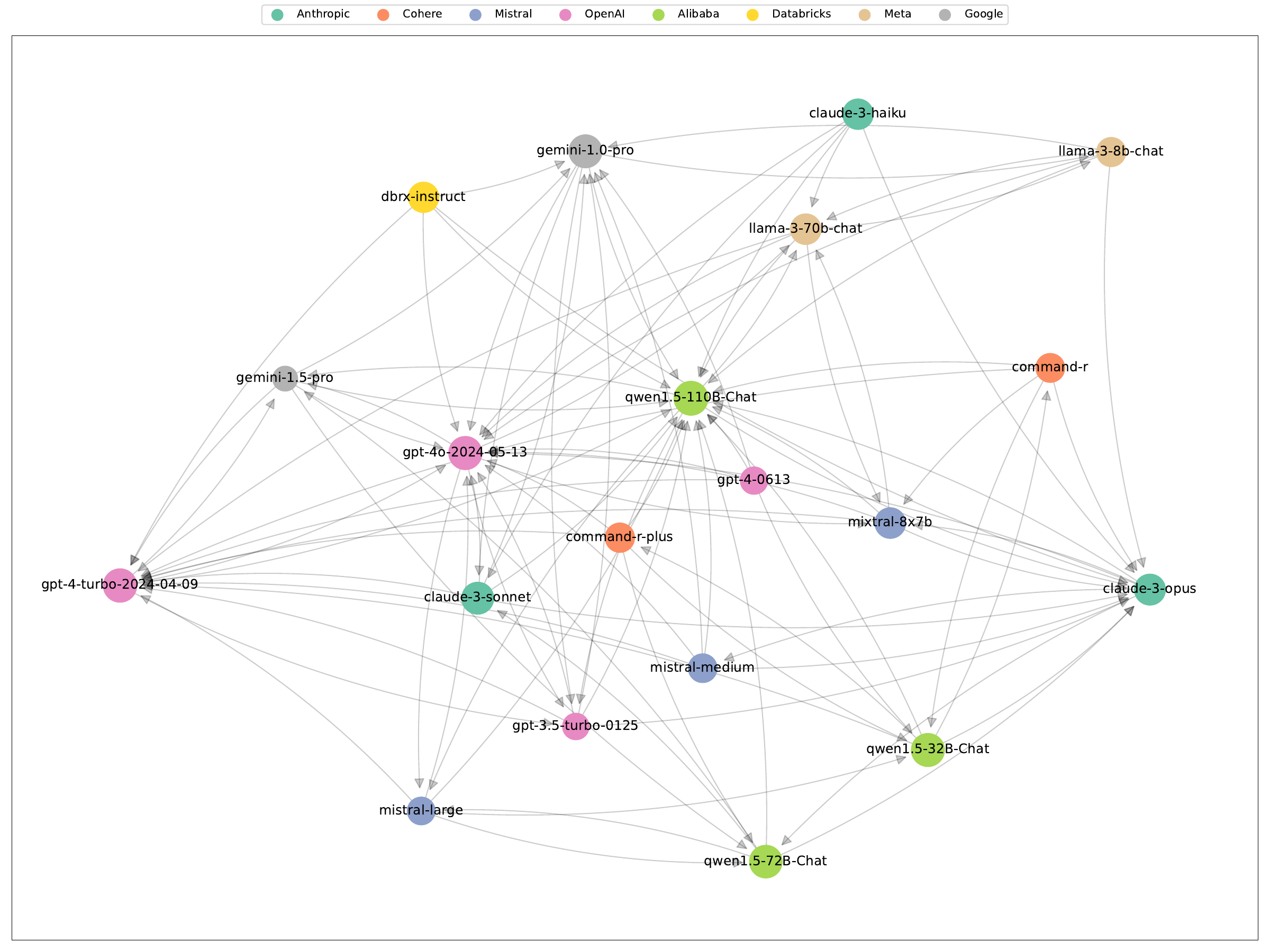}
\caption{Graph of top 5 affinities. An edge exists from LLM $a$ to LLM $b$ if $\text{affinity}(a, b)$ is in the top 5 affinities for LLM $a$. This view allows us to identify "popular" LLMs, "hipster" LLMs, and "LLM friendships" (where two LLMs have strong mutual affinity for each other). Popular LLMs, such as GPT-4o, have many arrows pointing toward them, while unpopular LLMs like dbrx-instruct have none. Qwen-1.5-32B and command-r form a "friendship" with mutual strong affinities, though Qwen-1.5-32B also receives incoming edges from three other LLMs, making it more widely liked. Some LLMs have only one "fan," others have several, and some have none. Of course, this analysis is somewhat contrived, as affinities are continuous values and using the top 5 as a cutoff is arbitrary. Nevertheless, it offers an interesting starting point for studying the dynamics and patterns that emerge when an arbitrary affinity threshold is established.}
\label{fig:top-5-affinities}
\end{figure*}

\begin{figure*}[t]
\centering
\includegraphics[width=\linewidth]{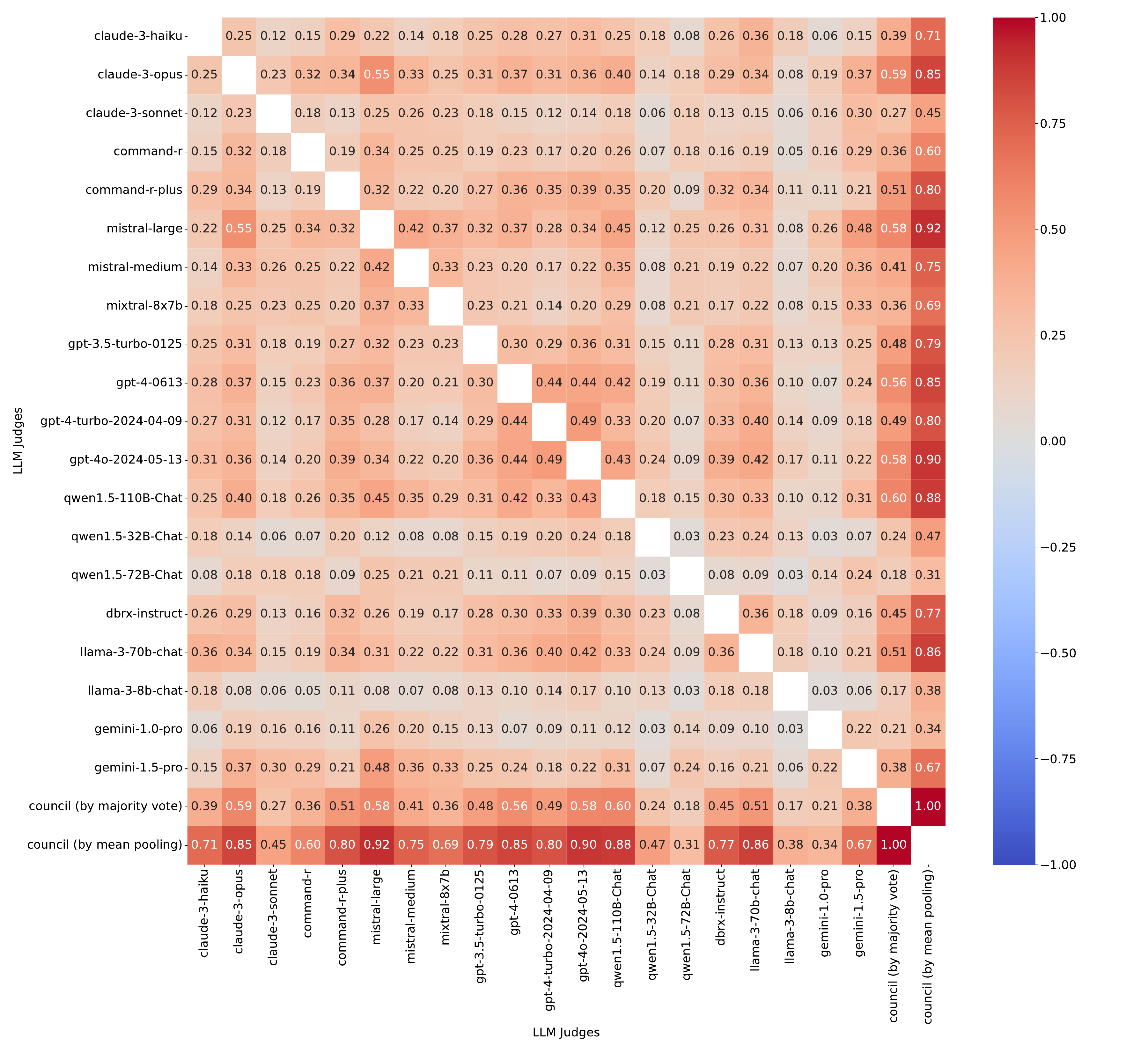}
\caption{Heatmap of LLM judge Cohen's $\kappa$ sidewise agreement scores. Since these are sidewise agreement scores, alignment on fine-grained ratings is not captured. However, the strong red band across the council rows indicates that the council is functioning as expected, representing a meaningful majority consensus. Inter-family agreement appears high overall, except within the Qwen-1.5 family, which shows lower scores of 0.03 and 0.08. In contrast, the OpenAI family demonstrates the highest inter-family agreement, with scores ranging from 0.30 to 0.49. Overall, LLM judges tend to be agreeing (no negative scores), though Llama-3-8b, Qwen-1.5-72b, and Gemini-1.0-Pro have the lowest agreement scores on the board. Interestingly, mistral-large and Claude-3-Opus show notably higher agreement scores than any other LLM pair.}
\label{fig:sidewiseagreement}
\end{figure*}

\begin{figure*}[t]
\centering
\includegraphics[width=\linewidth]{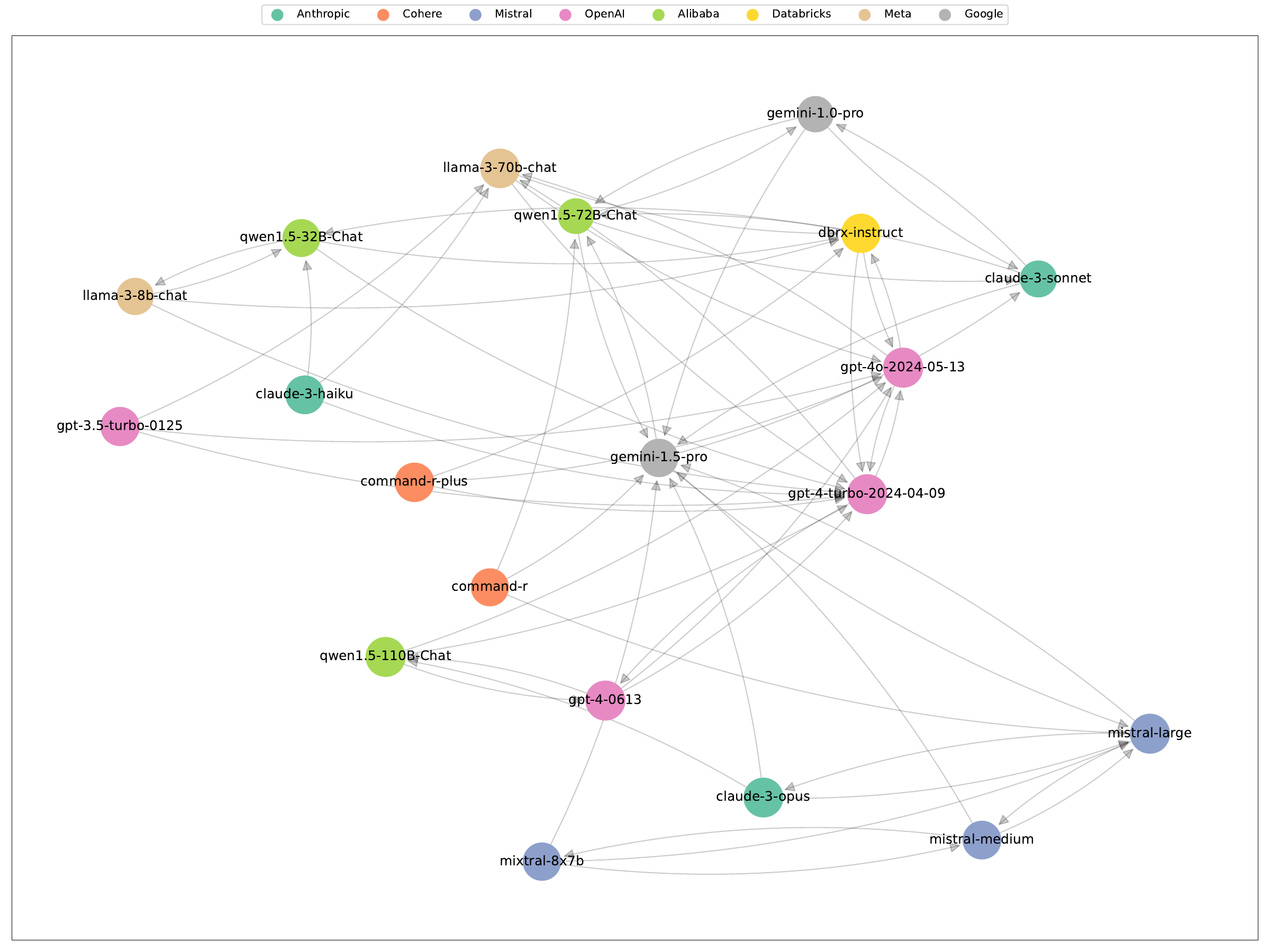}
\caption{Graph of top 5 agreement. An edge exists from LLM $a$ to LLM $b$ if $\text{agreement}(a, b)$ is in the top 5 agreement scores for LLM $a$. This visualization helps identify representative LLMs—those with many arrows pointing toward them are the ones that many other LLMs agree with. It also reveals communities of LLMs that tend to align with each other. While families exhibit high agreement in Figure \ref{fig:sidewiseagreement}, this graph shows fewer arrows within families, suggesting that certain non-family LLMs achieve higher agreement. A reverse graph, showing the bottom 5 agreement scores, could highlight contrarian LLMs. Overall, this approach helps identify the LLMs most agreed upon by other council members, which can be useful when selecting a representative sub-council.}
\label{fig:top-5-agreement}
\end{figure*}

\begin{figure*}[t]
\centering
\includegraphics[width=\linewidth]{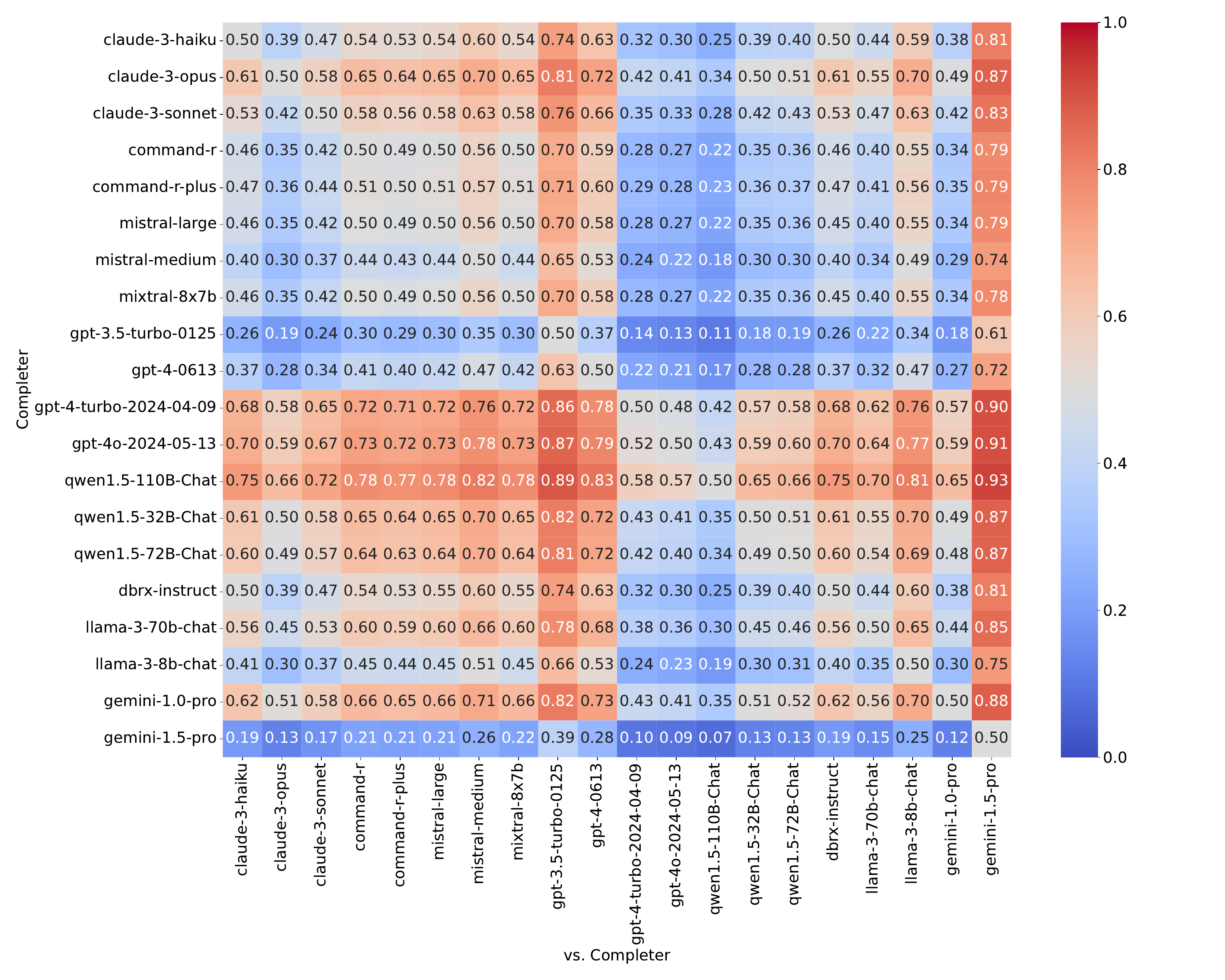}
\caption{Heatmap of the estimated LLM vs. LLM win rates. One notable outcome of using the Bradley-Terry \cite{bradley1952rank} estimation with a single reference model is the elimination of the "rock-paper-scissors" effect, where LLMs might have disproportionately favorable matchups. With a single reference model, the estimated win rates across all model pairs remain perfectly consistent, ensuring that while win rates vary, the distributional variance of these rates remains fixed. This design promotes stable relative rankings, which is desirable for evaluation purposes. However, it likely deviates from real-world scenarios where head-to-head win rates would be more heterogeneous, with different LLMs having specific dynamic advantages over others.}
\label{fig:faceoffwinrates}
\end{figure*}

The 20x20 LLM interactions generate a wealth of data that spans multiple pages. For ease of reference, all large tables and figures are compiled here:

\begin{itemize}
    \item Table \ref{tab:judge-extended-profile-bias} shows measures of bias for individual judges and the LMC as a whole.
    \item Table \ref{tab:judge-extended-profile-agreement} shows measures of agreement for individual LLM judges.
    \item Table \ref{tab:judge-extended-profile-affinity} shows polarization and affinity for individual LLM judges.
    \item Figure \ref{fig:raw-affinities} shows a heatmap of the affinities between LLM judges and LLM respondents.
    \item Figure \ref{fig:raw-affinities-council-normalized} shows a heatmap of the normalized affinities between LLM judges and LLM respondents (the LMC's consensus affinity subtracted out).
    \item Figure \ref{fig:top-5-affinities} shows a graph consisting of each LLM judge's top 5 affinities.
    \item Figure \ref{fig:sidewiseagreement} shows a heatmap of Cohen's $\kappa$ sidewise agreement scores.
    \item Figure \ref{fig:top-5-agreement} shows a graph consisting of each LLM judge's top 5 most agreeable other LLMs.
    \item Figure \ref{fig:faceoffwinrates} shows a heatmap of the estimated LLM vs. LLM win rates.
\end{itemize}

\begin{table*}[t]
\centering
\scalebox{0.6}{
\begin{tabular}{lcc|cc}
\toprule
& \multicolumn{2}{c}{\textbf{Models with $>$200 words}} & \multicolumn{2}{c}{\textbf{All models}} \\
\cmidrule{2-5}
& \makecell{\textbf{All}\\ \textbf{ votes}} & \makecell{\textbf{Consistent}\\ \textbf{ votes}} & \makecell{\textbf{All}\\ \textbf{ votes}} & \makecell{\textbf{Consistent}\\ \textbf{ votes}} \\
\midrule
Average judge & \textbf{0.158} & 
\textbf{0.143} & 0.502 & 0.319 \\
\midrule
LMC (majority) & 0.116 & 0.129 & 0.365 & 0.354 \\
LMC (mean pool) & 0.112 & 0.139 & \textbf{0.592} & \textbf{0.389} \\
LMC (no aggregation) & 0.125 & 0.106 & 0.545 & 0.347 \\
\bottomrule
\end{tabular}
}
\vspace{2mm}
\caption{Length bias with and without models with responses $<$200 words.}
\label{tab:lengthbiascorrected}
\end{table*}

\begin{figure*}[ht]
\centering
\includegraphics[width=\linewidth]{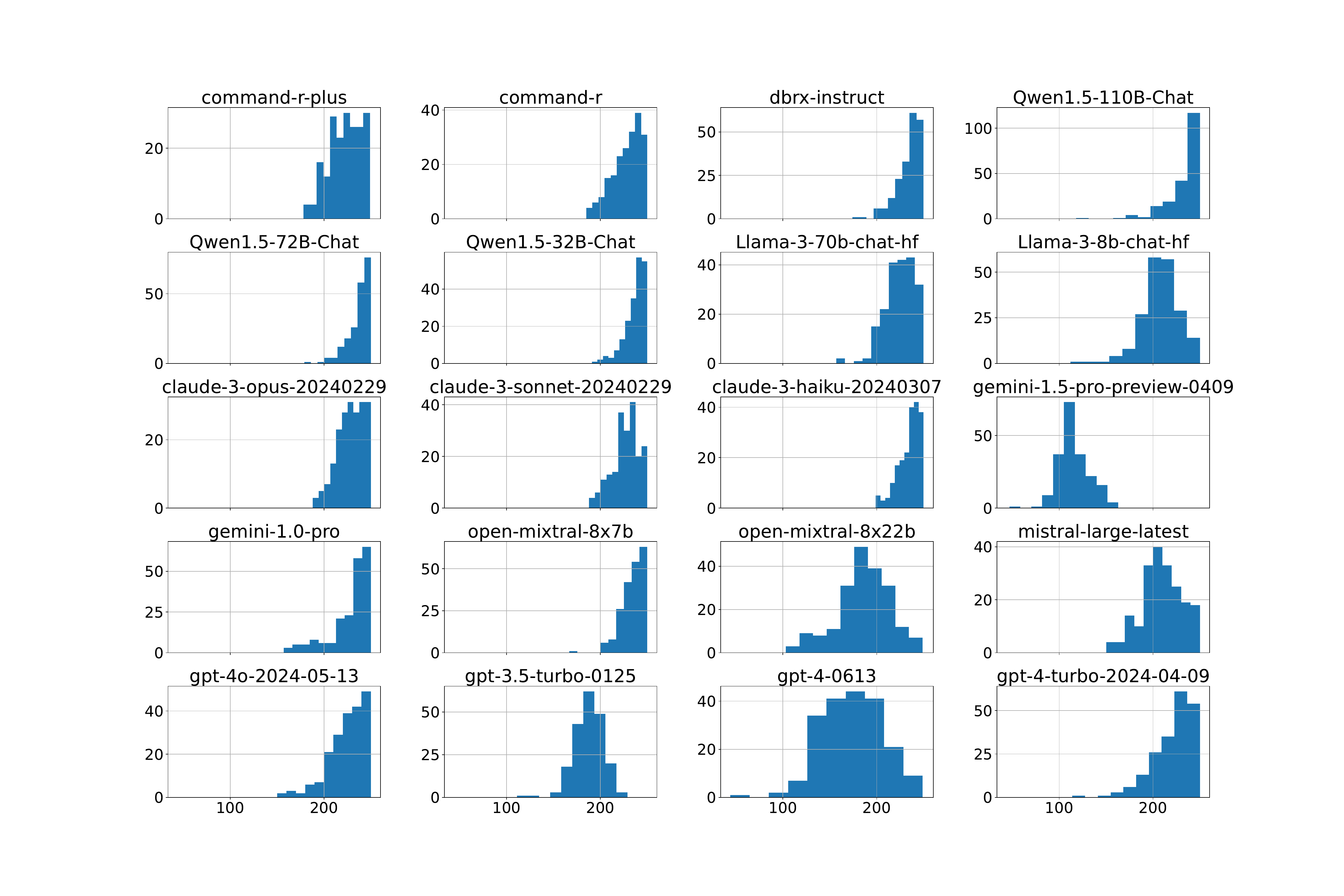}
\caption{Distribution of response lengths for 20 LLMs on our EI task, measured in number of tokens.}
\label{fig:respondent-response-length-distributions}
\end{figure*}

\begin{figure*}
\centering
\includegraphics[width=\linewidth]{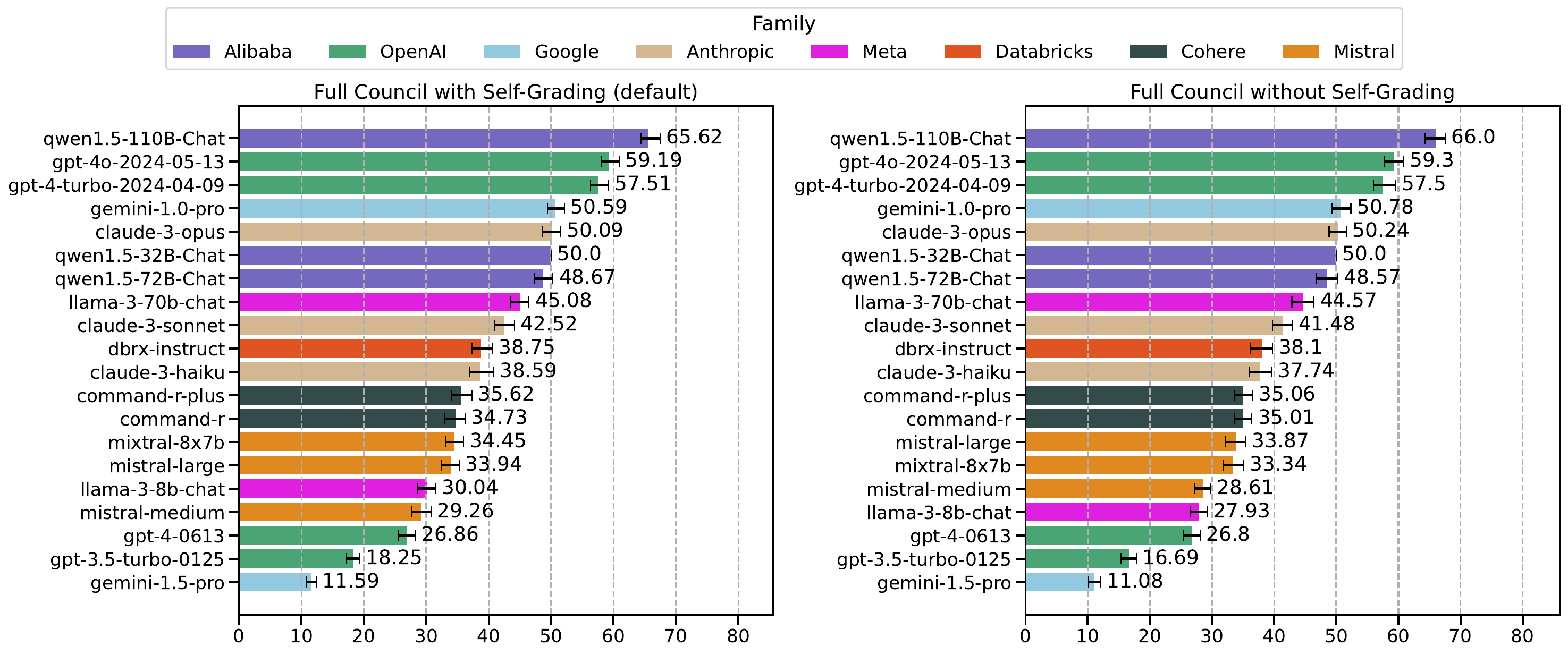}
\caption{Comparison of full LMC rankings (no aggregation), with self-grading (left, default) permitted and with self-grading disabled (right). The rankings are broadly identical, confirming that the ensemble of LLM judges of the Language Model Council mitigates self-enhancement bias.}
\label{fig:council_rankings_with_and_without_self_grading}
\end{figure*}

\begin{table*}[ht]
\centering
\scalebox{0.55}{
\begin{tabular}{c|cccc|cccc}
\toprule
& \multicolumn{4}{c|}{\underline{All votes}} & \multicolumn{4}{c}{\underline{Consistent votes}} \\
\textbf{LLM} & \textbf{Position bias} \textbf{(first)} & \textbf{Position bias (second)} & \textbf{Self bias} & \textbf{Length
bias} & \textbf{Position bias} \textbf{(first)} & \textbf{Position bias (second)} & \textbf{Self bias} & \textbf{Length
bias} \\
\midrule
qwen1.5-110B-Chat & 26.6\% & 5.8\% & 0.03 & 0.44 & 0.00\% & 0.00\% & -0.04 & 0.31 \\
gpt-4o-2024-05-13 & 47.5\% & 1.7\% & 0.08 & 0.45 & 0.00\% & 0.00\% & 0.13 & 0.22 \\
gpt-4-turbo-2024-04-09 & 59.0\% & 2.5\% & 0.01 & 0.35 & 0.00\% & 0.00\% & 0.11 & 0.18 \\
gemini-1.0-pro & 5.2\% & 60.0\% & -0.01 & 0.52 & 0.00\% & 0.00\% & -0.02 & 0.35 \\
claude-3-opus & 9.2\% & 16.2\% & -0.08 & 0.36 & 0.00\% & 0.00\% & 0 & 0.37 \\
qwen1.5-32B-Chat & 75.5\% & 1.0\% & 0.00 & 0.77 & 0.00\% & 0.00\% & -0.11 & 0.28 \\
qwen1.5-72B-Chat & 0.4\% & 72.7\% & 0.00 & 0.60 & 0.00\% & 0.00\% & 0.07 & 0.31 \\
llama-3-70b-chat & 46.9\% & 1.9\% & 0.11 & 0.50 & 0.00\% & 0.00\% & 0.24 & 0.3 \\
claude-3-sonnet & 4.0\% & 56.3\% & 0.11 & 0.66 & 0.00\% & 0.00\% & 0.24 & 0.48 \\
dbrx-instruct & 52.0\% & 3.8\% & 0.03 & 0.63 & 0.00\% & 0.00\% & 0.04 & 0.32 \\
claude-3-haiku & 52.1\% & 3.7\% & 0.07 & 0.62 & 0.00\% & 0.00\% & 0.17 & 0.34 \\
command-r-plus & 45.1\% & 2.1\% & 0.06 & 0.55 & 0.00\% & 0.00\% & 0.13 & 0.39 \\
command-r & 7.4\% & 38.1\% & -0.08 & 0.62 & 0.00\% & 0.00\% & -0.1 & 0.42 \\
mixtral-8x7b & 8.2\% & 33.2\% & 0.07 & 0.52 & 0.00\% & 0.00\% & 0.15 & 0.4 \\
mistral-large & 4.4\% & 23.1\% & -0.07 & 0.32 & 0.00\% & 0.00\% & -0.07 & 0.21 \\
llama-3-8b-chat & 71.7\% & 2.2\% & 0.21 & 0.51 & 0.00\% & 0.00\% & 0.55 & 0.28 \\
mistral-medium & 11.8\% & 29.2\% & -0.02 & 0.53 & 0.00\% & 0.00\% & 0.01 & 0.41 \\
gpt-4-0613 & 37.8\% & 8.6\% & -0.06 & 0.42 & 0.00\% & 0.00\% & -0.04 & 0.23 \\
gpt-3.5-turbo-0125 & 32.7\% & 9.6\% & 0.04 & 0.42 & 0.00\% & 0.00\% & 0 & 0.29 \\
gemini-1.5-pro & 1.6\% & 46.1\% & 0.14 & 0.26 & 0.00\% & 0.00\% & -0.02 & 0.29 \\
\midrule
Average Judge & 30.0\% & 20.9\% & 0.03 & 0.50 & 0.0\% & 0.0\% & 0.07 & 0.32 \\
\midrule
council (by majority vote) & 21.5\% & 3.2\% & & 0.36 & 3.10\% & 0.10\% & & 0.35 \\
council (by mean pooling) & 26.5\% & 5.0\% & & 0.59 & 1.80\% & 0.90\% & & 0.39 \\
council (no aggregation) & 1.6\% & 46.1\% & & 0.54 & 0.00\% & 0.00\% & & 0.35 \\
\bottomrule
\end{tabular}
}
\vspace{5mm}
\caption{LMC judging profile relates for bias, with and without consistent votes.}
\label{tab:judge-extended-profile-bias}
\end{table*}

\begin{table*}[t]
\centering

\scalebox{0.55}{
\begin{tabular}{c|ccc|ccc}
\toprule
& \multicolumn{3}{c|}{\underline{All votes}} & \multicolumn{3}{c}{\underline{Consistent votes}} \\
\textbf{LLM} & \textbf{Contrarianism}
& \textbf{Agrees most with} & \textbf{Disagrees most with} & \textbf{Contrarianism} & \textbf{Agrees most with} & \textbf{Disagrees most with} \\
\midrule
qwen1.5-110B-Chat & 19.2\% & gpt-4o-2024-05-13 & qwen1.5-72B-Chat & 8.30\% & qwen1.5-72B-Chat & llama-3-8b-chat \\
gpt-4o-2024-05-13 & 18.8\% & gpt-4-turbo-2024-04-09 & qwen1.5-72B-Chat & 5.20\% & gemini-1.5-pro & llama-3-8b-chat \\
gpt-4-turbo-2024-04-09 & 21.4\% & gpt-4o-2024-05-13 & qwen1.5-72B-Chat & 5.90\% & gemini-1.5-pro & llama-3-8b-chat \\
gemini-1.0-pro & 43.3\% & qwen1.5-72B-Chat & qwen1.5-32B-Chat & 17.90\% & gpt-4o-2024-05-13 & llama-3-8b-chat \\
claude-3-opus & 20.6\% & mistral-large & llama-3-8b-chat & 13.80\% & qwen1.5-72B-Chat & llama-3-8b-chat \\
qwen1.5-32B-Chat & 32.2\% & llama-3-8b-chat & qwen1.5-72B-Chat & 9.70\% & gpt-4-turbo-2024-04-09 & llama-3-8b-chat \\
qwen1.5-72B-Chat & 46.6\% & claude-3-sonnet & qwen1.5-32B-Chat & 7.80\% & qwen1.5-110B-Chat & llama-3-8b-chat \\
llama-3-70b-chat & 22.3\% & gpt-4-turbo-2024-04-09 & qwen1.5-72B-Chat & 8.20\% & gemini-1.5-pro & gemini-1.0-pro \\
claude-3-sonnet & 40.1\% & qwen1.5-72B-Chat & qwen1.5-32B-Chat & 13.10\% & gpt-4-turbo-2024-04-09 & command-r \\
dbrx-instruct & 24.5\% & gpt-4-turbo-2024-04-09 & qwen1.5-72B-Chat & 9.50\% & gpt-4o-2024-05-13 & llama-3-8b-chat \\
claude-3-haiku & 27.6\% & llama-3-70b-chat & qwen1.5-72B-Chat & 13.00\% & llama-3-70b-chat & qwen1.5-32B-Chat \\
command-r-plus & 22.8\% & gpt-4-turbo-2024-04-09 & qwen1.5-72B-Chat & 8.80\% & gemini-1.5-pro & llama-3-8b-chat \\
command-r & 33.5\% & gemini-1.5-pro & llama-3-8b-chat & 15.30\% & gpt-4-turbo-2024-04-09 & llama-3-8b-chat \\
mixtral-8x7b & 33.5\% & gemini-1.5-pro & llama-3-8b-chat & 15.90\% & qwen1.5-72B-Chat & gemini-1.0-pro \\
mistral-large & 21.2\% & claude-3-opus & llama-3-8b-chat & 6.00\% & gemini-1.5-pro & llama-3-8b-chat \\
llama-3-8b-chat & 36.0\% & qwen1.5-32B-Chat & qwen1.5-72B-Chat & 25.70\% & llama-3-70b-chat & gpt-4-turbo-2024-04-09 \\
mistral-medium & 30.5\% & mistral-large & llama-3-8b-chat & 12.20\% & qwen1.5-72B-Chat & llama-3-8b-chat \\
gpt-4-0613 & 20.3\% & gpt-4-turbo-2024-04-09 & qwen1.5-72B-Chat & 7.90\% & gpt-4o-2024-05-13 & llama-3-8b-chat \\
gpt-3.5-turbo-0125 & 25.1\% & gpt-4o-2024-05-13 & qwen1.5-72B-Chat & 12.80\% & gpt-4o-2024-05-13 & llama-3-8b-chat \\
gemini-1.5-pro & 33.2\% & mistral-large & llama-3-8b-chat & 4.00\% & gpt-4-turbo-2024-04-09 & llama-3-8b-chat \\
\midrule
Average Judge & 28.6\% & & & 11.1\% & & \\
\midrule
council (by majority vote) & & gpt-4o-2024-05-13 & qwen1.5-72B-Chat & & gemini-1.5-pro & llama-3-8b-chat \\
council (by mean pooling) & & mistral-large & qwen1.5-72B-Chat & 0.10\% & gemini-1.5-pro & llama-3-8b-chat \\
council (no aggregation) & & & & & & \\
\bottomrule
\end{tabular}
}
\vspace{5mm}
\caption{LMC judging profiles related to agreement.}
\label{tab:judge-extended-profile-agreement}
\end{table*}

\begin{table*}[t]
\centering

\scalebox{0.55}{
\begin{tabular}{c|ccc|ccc}
\toprule
& \multicolumn{3}{c|}{\underline{All votes}} & \multicolumn{3}{c}{\underline{Consistent votes}} \\
\textbf{LLM} & \textbf{Polarization} & \textbf{Lowest affinity for} & \textbf{Highest affinity for} & \textbf{Polarization} & \textbf{Lowest affinity for} & \textbf{Highest affinity for} \\
\midrule
qwen1.5-110B-Chat & 62.6 & gemini-1.5-pro & qwen1.5-110B-Chat & 78.30\% & gemini-1.5-pro & qwen1.5-110B-Chat \\
gpt-4o-2024-05-13 & 65.4 & gemini-1.5-pro & gpt-4o-2024-05-13 & 84.80\% & gemini-1.5-pro & gpt-4o-2024-05-13 \\
gpt-4-turbo-2024-04-09 & 54.5 & gpt-3.5-turbo-0125 & gpt-4o-2024-05-13 & 88.20\% & mistral-medium & gpt-4o-2024-05-13 \\
gemini-1.0-pro & 31.0 & gemini-1.5-pro & qwen1.5-110B-Chat & 72.30\% & gemini-1.5-pro & qwen1.5-110B-Chat \\
claude-3-opus & 73.0 & gpt-3.5-turbo-0125 & qwen1.5-110B-Chat & 93.30\% & gemini-1.5-pro & qwen1.5-110B-Chat \\
qwen1.5-32B-Chat & 46.7 & gemini-1.5-pro & gpt-4-turbo-2024-04-09 & 87.50\% & gpt-3.5-turbo-0125 & qwen1.5-110B-Chat \\
qwen1.5-72B-Chat & 45.7 & gemini-1.5-pro & qwen1.5-110B-Chat & 84.90\% & gemini-1.5-pro & qwen1.5-110B-Chat \\
llama-3-70b-chat & 68.3 & gemini-1.5-pro & qwen1.5-110B-Chat & 89.30\% & gemini-1.5-pro & qwen1.5-110B-Chat \\
claude-3-sonnet & 49.7 & gemini-1.5-pro & gpt-4o-2024-05-13 & 82.60\% & gemini-1.5-pro & gpt-4o-2024-05-13 \\
dbrx-instruct & 54.5 & gemini-1.5-pro & qwen1.5-110B-Chat & 85.00\% & gemini-1.5-pro & qwen1.5-110B-Chat \\
claude-3-haiku & 51.1 & gemini-1.5-pro & qwen1.5-110B-Chat & 79.40\% & gemini-1.5-pro & qwen1.5-110B-Chat \\
command-r-plus & 52.5 & gemini-1.5-pro & qwen1.5-110B-Chat & 78.40\% & gemini-1.5-pro & qwen1.5-110B-Chat \\
command-r & 44.4 & gemini-1.5-pro & qwen1.5-110B-Chat & 53.80\% & gemini-1.5-pro & gemini-1.0-pro \\
mixtral-8x7b & 59.4 & gemini-1.5-pro & qwen1.5-110B-Chat & 81.30\% & gemini-1.5-pro & qwen1.5-110B-Chat \\
mistral-large & 78.8 & gemini-1.5-pro & qwen1.5-110B-Chat & 92.00\% & gpt-3.5-turbo-0125 & qwen1.5-110B-Chat \\
llama-3-8b-chat & 34.9 & gpt-3.5-turbo-0125 & llama-3-70b-chat & 76.20\% & gpt-3.5-turbo-0125 & llama-3-70b-chat \\
mistral-medium & 58.0 & gemini-1.5-pro & qwen1.5-110B-Chat & 81.90\% & gemini-1.5-pro & qwen1.5-110B-Chat \\
gpt-4-0613 & 62.0 & gemini-1.5-pro & qwen1.5-110B-Chat & 86.00\% & gemini-1.5-pro & qwen1.5-110B-Chat \\
gpt-3.5-turbo-0125 & 65.6 & gemini-1.5-pro & qwen1.5-110B-Chat & 84.20\% & gemini-1.5-pro & qwen1.5-110B-Chat \\
gemini-1.5-pro & 61.7 & gpt-3.5-turbo-0125 & qwen1.5-110B-Chat & 90.00\% & gemini-1.5-pro & gpt-4o-2024-05-13 \\
\midrule
Average Judge & 56.0 & & & 82.47\% & & \\
\midrule
council (by majority vote) & 77.0 & gemini-1.5-pro & qwen1.5-110B-Chat & 82.50\% & gemini-1.5-pro & qwen1.5-110B-Chat \\
council (by mean pooling) & 60.3 & gemini-1.5-pro & qwen1.5-110B-Chat & 80.50\% & gemini-1.5-pro & qwen1.5-110B-Chat \\
council (no aggregation) & 54.0 & gemini-1.5-pro & qwen1.5-110B-Chat & 81.10\% & gemini-1.5-pro & qwen1.5-110B-Chat \\
\bottomrule
\end{tabular}
}
\vspace{5mm}
\caption{LMC judging profiles related to affinity.}
\label{tab:judge-extended-profile-affinity}
\end{table*}

\begin{table*}[ht]
\scriptsize
\begin{tabular}{lllccccl}
\toprule
\textbf{Country} & \textbf{Organization} & \textbf{LLM} & \makecell{\textbf{Release}\\\textbf{Date}} & \makecell{\textbf{Chat Arena}\\ \textbf{Elo}} & \makecell{\textbf{MMLU}\\
\textbf{(5-shot)}} & \textbf{Size} & \textbf{License} \\ \midrule
United States & Open AI & gpt-4o-2024-05-13 \cite{gpt4o} & 05/24 & 1287 & 88.7 & \lock & Proprietary \\
United States & Open AI & gpt-4-turbo-04-09 \cite{gpt4-turbo} & 04/24 & 1256 & \lock  & \lock & Proprietary \\
United States & Open AI & gpt-4-0613 \cite{gpt4-0613} & 06/23 & 1246 & 86.4 & \lock & Proprietary \\
United States & Open AI & gpt-3.5-turbo-0125 \cite{gpt35} & 01/24 & 1102 & 70.0 & \lock & Proprietary \\
France & Mistral & mistral-large-latest \cite{mistralLarge} & 02/24 & 1156 & 81.2 & \lock & Proprietary \\
France & Mistral & open-mixtral-8x22b \cite{mixtral22b} & 04/24 & 1146 & 77.8 & 176 B & Apache 2.0 \\
France & Mistral & open-mixtral-8x7b \cite{mixtral} & 12/23 & 1114 & 70.6 & 56 B & Apache 2.0 \\
United States & Meta & llama-3-70b-chat-hf \cite{llama3} & 04/24 & 1208 & 82.0 & 70 B & Llama 3 Community \\
United States & Meta & llama-3-8b-chat-hf \cite{llama3} & 04/24 & 1153 & 68.4 & 8 B & Llama 3 Community \\
United States & Google & gemini-1.5-pro-preview-0409 \cite{gemini15} & 05/24 & 1268 & 81.9 & \lock & Proprietary \\
United States & Google & gemini-1.0-pro \cite{gemini10} & 04/24 & 1208 & 71.8 & \lock & Proprietary \\
United States & Databricks & dbrx \cite{databricksIntroducingDBRX} & 03/24 & 1103 & 73.7 & 132 B & DBRX LICENSE \\
Canada & Cohere & command-r-plus \cite{command-r-plus} & 04/24 & 1189 & 75.7 & 104 B & CC-BY-NC-4.0 \\
Canada & Cohere & command-r \cite{command-r} & 04/24 & 1147 & 68.2 & 35 B & CC-BY-NC-4.0 \\
United States & Anthropic & claude-3-opus-20240229 \cite{claude3} & 03/24 & 1248 & 86.8 & \lock & Proprietary \\
United States & Anthropic & claude-3-sonnet-20240229 \cite{claude3} & 03/24 & 1201 & 79.0 & \lock & Proprietary \\
United States & Anthropic & claude-3-haiku-20240307 \cite{claude3} & 03/24 & 1178 & 75.2 & \lock & Proprietary \\
China & Alibaba & qwen1.5-110B-chat \cite{qwen} & 02/24 & 1164 & 80.2 & 100 B & Qianwen LICENSE \\
China & Alibaba & qwen1.5-72B-chat \cite{qwen} & 02/24 & 1152 & 77.4 & 72 B & Qianwen LICENSE \\
China & Alibaba & qwen1.5-32B-chat \cite{qwen} & 02/24 & 1126 & 74.3 & 32 B & Qianwen LICENSE \\
\bottomrule
\end{tabular}
\vspace{3mm}
\caption{20 council members used for experiments in this work. We include models from eight different organizations across four countries, with a mix of open and closed-source models, small and large models. To our knowledge, this is the largest panel of LLM judges studied to date.}
\label{tab:listofcouncilmembers}
\end{table*}


\begin{table*}[ht]
\centering
\scriptsize
\small
\begin{tabular}{lcccc}
\toprule
\textbf{LLM} & \textbf{All} & \textbf{Flagships} & \textbf{Smalls} & \textbf{Top-4} \\
\midrule
gpt-4o-2024-05-13 & {\includegraphics[width=0.3cm, height=0.3cm]{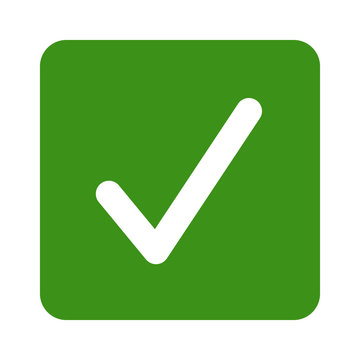}} & \includegraphics[width=0.3cm, height=0.3cm]{figures/checkbox.jpg} & & \includegraphics[width=0.3cm, height=0.3cm]{figures/checkbox.jpg} \\
gpt-4-turbo-04-09 & \includegraphics[width=0.3cm, height=0.3cm]{figures/checkbox.jpg} & & & \includegraphics[width=0.3cm, height=0.3cm]{figures/checkbox.jpg} \\
gpt-4-0613 & \includegraphics[width=0.3cm, height=0.3cm]{figures/checkbox.jpg} & & & \\
gpt-3.5-turbo-0125 & \includegraphics[width=0.3cm, height=0.3cm]{figures/checkbox.jpg} & & & \\
mistral-large-latest & \includegraphics[width=0.3cm, height=0.3cm]{figures/checkbox.jpg} & \includegraphics[width=0.3cm, height=0.3cm]{figures/checkbox.jpg} & & \\
open-mixtral-8x22b & \includegraphics[width=0.3cm, height=0.3cm]{figures/checkbox.jpg} & & & \\
open-mixtral-8x7b & \includegraphics[width=0.3cm, height=0.3cm]{figures/checkbox.jpg} & & \includegraphics[width=0.3cm, height=0.3cm]{figures/checkbox.jpg} & \\
llama-3-70b-chat-hf & \includegraphics[width=0.3cm, height=0.3cm]{figures/checkbox.jpg} & \includegraphics[width=0.3cm, height=0.3cm]{figures/checkbox.jpg} & & \\
llama-3-8b-chat-hf & \includegraphics[width=0.3cm, height=0.3cm]{figures/checkbox.jpg} & & \includegraphics[width=0.3cm, height=0.3cm]{figures/checkbox.jpg} & \\
gemini-1.5-pro-preview-0409 & \includegraphics[width=0.3cm, height=0.3cm]{figures/checkbox.jpg} & \includegraphics[width=0.3cm, height=0.3cm]{figures/checkbox.jpg} & & \includegraphics[width=0.3cm, height=0.3cm]{figures/checkbox.jpg} \\
gemini-1.0-pro & \includegraphics[width=0.3cm, height=0.3cm]{figures/checkbox.jpg} & & \includegraphics[width=0.3cm, height=0.3cm]{figures/checkbox.jpg} & \\
dbrx & \includegraphics[width=0.3cm, height=0.3cm]{figures/checkbox.jpg} & \includegraphics[width=0.3cm, height=0.3cm]{figures/checkbox.jpg} & \includegraphics[width=0.3cm, height=0.3cm]{figures/checkbox.jpg} & \\
command-r-plus & \includegraphics[width=0.3cm, height=0.3cm]{figures/checkbox.jpg} & \includegraphics[width=0.3cm, height=0.3cm]{figures/checkbox.jpg} & & \\
command-r & \includegraphics[width=0.3cm, height=0.3cm]{figures/checkbox.jpg} & & \includegraphics[width=0.3cm, height=0.3cm]{figures/checkbox.jpg} & \\
claude-3-opus-20240229 & \includegraphics[width=0.3cm, height=0.3cm]{figures/checkbox.jpg} & \includegraphics[width=0.3cm, height=0.3cm]{figures/checkbox.jpg} & & \includegraphics[width=0.3cm, height=0.3cm]{figures/checkbox.jpg} \\
claude-3-sonnet-20240229 & \includegraphics[width=0.3cm, height=0.3cm]{figures/checkbox.jpg} & & & \\
claude-3-haiku-20240307 & \includegraphics[width=0.3cm, height=0.3cm]{figures/checkbox.jpg} & & \includegraphics[width=0.3cm, height=0.3cm]{figures/checkbox.jpg} & \\
qwen1.5-110B-chat & \includegraphics[width=0.3cm, height=0.3cm]{figures/checkbox.jpg} & \includegraphics[width=0.3cm, height=0.3cm]{figures/checkbox.jpg} & & \\
qwen1.5-72B-chat & \includegraphics[width=0.3cm, height=0.3cm]{figures/checkbox.jpg} & & & \\
qwen1.5-32B-chat & \includegraphics[width=0.3cm, height=0.3cm]{figures/checkbox.jpg} & & \includegraphics[width=0.3cm, height=0.3cm]{figures/checkbox.jpg} & \\
\bottomrule
\end{tabular}
\vspace{3mm}
\caption{Additional council variations consisting of a hand-picked subset of LLMs. \textbf{Flagships}: the largest LLM from each organization; \textbf{Smalls}: the smallest LLMs from each organization; and \textbf{Top-4}, the top 4 LLMs according to Chatbot Arena as of May 2024.}
\label{fig:oligarchalcouncilmembers}
\end{table*}

\clearpage
\section{LLM Judge Calibration}
\label{sec:judge-callibration}

To understand the reliability and natural variability of LLM model judges and to help us decide evaluation settings, we run a calibration exercise prior to the main experiment.

\begin{figure*}[ht]
    \centering
    \scalebox{0.9}{
        \includegraphics[width=\linewidth]{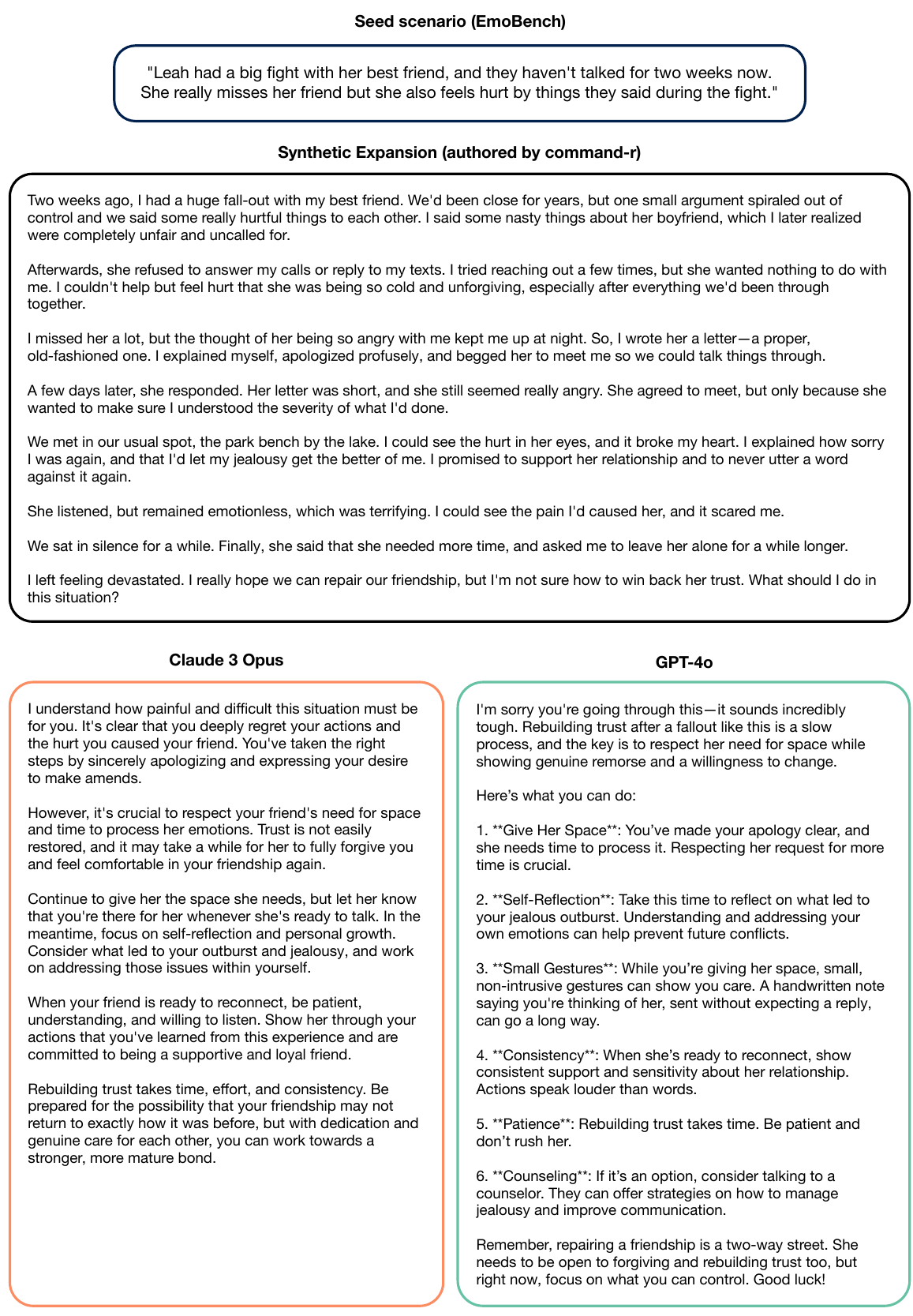}
    }
    \caption{The scenario, synthetic expansion, and responses used for pairwise comparison calibration. Three possible responses are evaluated: one from Claude Opus, one from GPT-4o, and a generic response: “I’m sorry it sounds like you are going through a rough time. I wish you the best.”}
    \label{fig:Judge Calibration}
\end{figure*}

We collect pairwise preference ratings on three responses to the same interpersonal conflict using different temperatures and pairwise comparison options. Two responses are competitive, and one is intentionally generic to serve as a ranking baseline (Figure \ref{fig:Judge Calibration}). We measure \textbf{Invariability} and \textbf{P}airwise \textbf{P}ositional \textbf{C}onsistency (\textbf{PPC}), defined below.

\newpage
\subsection{Invariability}

\begin{center}
    \textit{How reliably does the model give the same preference with the same pair of responses in the same order?}
\end{center}

Let:

\begin{itemize}
    \item \( P \) be the set of all pairs of responses.
    \item \( R_{i,j} \) be the result of the \( j \)-th repetition of the pairwise comparison of the \( i \)-th pair \( (x_i, y_i) \) in the same order.
    \item \( n \) be the number of repetitions.
\end{itemize}

For each pair \( (x_i, y_i) \), we perform \( n \) comparisons, resulting in a set of results \( \{R_{i,1}, R_{i,2}, \ldots, R_{i,n}\} \).

Define the mode of the set \( \{R_{i,1}, R_{i,2}, \ldots, R_{i,n}\} \) as \( \text{mode}(R_i) \).

The frequency of the mode for the \( i \)-th pair is given by:
\[ f_i = \frac{\sum_{j=1}^{n} \mathbb{I}(R_{i,j} = \text{mode}(R_i))}{n} \]
where \( \mathbb{I} \) is the indicator function, which is 1 if the condition inside is true, and 0 otherwise.

The invariability is then defined as the average of \( f_i \) over all pairs in \( P \):
\[ invariability = \frac{1}{|P|} \sum_{i \in P} f_i \]

\newpage
\subsection{Pairwise Positional Consistency (PPC)}
\label{sec:ppc}

\begin{center}
    \textit{How reliably does the model give a consistent preference with the same pair of responses in swapped order?}
\end{center}

A rating couplet consists of a single rating of a pair of responses and then a rating of the same pair of responses in swapped order. For multiple repetitions of the same pair of responses in both orders, we take the percentage of consistent couplets over all possible rating couplets to factor out spuriously inconsistent couplets.

Let:

\begin{itemize}
    \item \( P \) be the set of all pairs of responses.
    \item \( R_{i,j} \) be the result of the \( j \)-th repetition of the pairwise comparison of the \( i \)-th pair \( (x_i, y_i) \) in the same order.
    \item \( R_{i',j} \) be the result of the \( j \)-th repetition of the pairwise comparison of the \( i \)-th pair \( (y_i, x_i) \) in swapped order.
    \item \( n \) be the number of repetitions.
\end{itemize}

For each pair \( (x_i, y_i) \), we perform \( n \) comparisons in both the original and swapped orders, resulting in two sets of results: \( \{R_{i,1}, R_{i,2}, \ldots, R_{i,n}\} \) and \( \{R_{i',1}, R_{i',2}, \ldots, R_{i',n}\} \).

We define a consistency function \( \text{are\_consistent}(R_{i,j}, R_{i',k}) \) which returns 1 if the results \( R_{i,j} \) and \( R_{i',k} \) are consistent (i.e., the model gives a consistent answer for both orders), and 0 otherwise based on reference table Figure \ref{tab:consistency}.

Consistency is then defined as the average consistency over all pairs \( (i, j) \in P \) and repetitions:

\begin{small}
\[ ppc = \frac{1}{|P| \cdot n^2} \sum_{i \in P} \sum_{j=1}^{n} \sum_{k=1}^{n} \text{are\_consistent}(R_{i,j}, R_{i',k}) \]
\end{small}

This is equivalent to the percentage of consistent couplets over all possible rating couplets.

\newpage
\subsection{Experiment}

\begin{figure*}[ht]
\centering
\includegraphics[width=\linewidth]{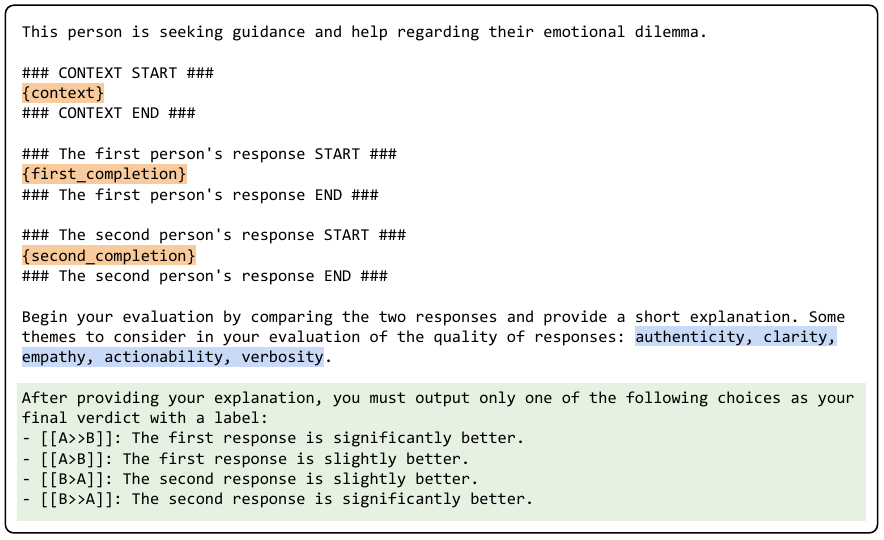}
\caption{Prompt used for pairwise comparison between responses.}
\label{fig:prompt-template-sxs-granular-no-tie}
\end{figure*}

Each LLM judge is prompted 5 times with the original pairwise comparison prompt (Figure \ref{fig:prompt-template-sxs-granular-no-tie}) and 5 times with a trivially reworded version of the prompt.\footnote{Trivial rewording involves changing the first sentence of the judging prompt (Figure \protect\ref{fig:prompt-template-sxs-granular-no-tie}) to: \textit{"This person is experiencing an emotional dilemma and is seeking guidance and help."}} This is repeated for the swapped order of responses.

For a single pair of responses, there are 10 repetitions (5 repetitions for each prompt $*$ 2 prompts) in one order and 10 reps in the swapped order. Thus, there are $10 * 10 = 100$ possible rating couplets, which forms the denominator for the calculation of PPC.

The \texttt{are\_consistent} function for consistency metrics is based on the mapping defined in Table \ref{tab:consistency}.

\begin{table*}[t]
\centering

\scalebox{0.75}{
\begin{tabular}{cccccc}
\toprule
\textbf{Rating}                                    & \textbf{Order-swapped rating}                      & \multicolumn{1}{l}{\textbf{Consistent}} & \multicolumn{1}{l}{\textbf{Inconsistent}} & \multicolumn{1}{l}{\textbf{Biased towards first}} & \multicolumn{1}{l}{\textbf{Biased towards second}} \\
\midrule
A\textgreater{}\textgreater{}B or A\textgreater{}B & A\textgreater{}\textgreater{}B or A\textgreater{}B & FALSE                                   & TRUE                                      & TRUE                                              & FALSE                                              \\
A\textgreater{}\textgreater{}B or A\textgreater{}B & B\textgreater{}\textgreater{}A or B\textgreater{}A & TRUE                                    & FALSE                                     & FALSE                                             & FALSE                                              \\
A\textgreater{}\textgreater{}B or A\textgreater{}B & A$\sim$=B                                          & FALSE                                   & TRUE                                      & TRUE                                              & FALSE                                              \\
B\textgreater{}\textgreater{}A or B\textgreater{}A & A\textgreater{}\textgreater{}B or A\textgreater{}B & TRUE                                    & FALSE                                     & FALSE                                             & FALSE                                              \\
B\textgreater{}\textgreater{}A or B\textgreater{}A & B\textgreater{}\textgreater{}A or B\textgreater{}A & FALSE                                   & TRUE                                      & FALSE                                             & TRUE                                               \\
B\textgreater{}\textgreater{}A or B\textgreater{}A & A$\sim$=B                                          & FALSE                                   & TRUE                                      & FALSE                                             & TRUE                                               \\
A$\sim$=B                                          & A\textgreater{}\textgreater{}B or A\textgreater{}B & FALSE                                   & TRUE                                      & TRUE                                              & FALSE                                              \\
A$\sim$=B                                          & B\textgreater{}\textgreater{}A or B\textgreater{}A & FALSE                                   & TRUE                                      & FALSE                                             & TRUE                                               \\
A$\sim$=B                                          & A$\sim$=B                                          & TRUE                                    & FALSE                                     & FALSE                                             & FALSE                                   \\
        \bottomrule
\end{tabular}
}
\vspace{5mm}
\caption{Reference table for categorizing a couplet of order-swapped ratings of the same set of items, (A, B) vs. (B, A). Consistency is still counted as long as the overall side of the preference is consistent. Position-inconsistent ratings are either biased towards the first or second position.}
\label{tab:consistency}
\end{table*}

We test 3 different temperatures ($0.0$, $0.5$, $1.0$) and $4$ different sets of pairwise comparison options:

\squishlist
    \item Coarse preferences with tie option 
    
    (A\textgreater B, B\textgreater A, A=B)
    
    \item Coarse preferences without tie option 
    
    (A\textgreater B, B\textgreater A)
    
    \item Granular preferences with tie option 
    
    (A\textgreater{}\textgreater{}B, A\textgreater B, B\textgreater A, B\textgreater{}\textgreater{} A, A=B)
    
    \item Granular preferences without tie option 
    
    (A\textgreater{}\textgreater{}B, A\textgreater B, B\textgreater A, B\textgreater{}\textgreater{}A)
\squishend

\begin{figure*}[ht]
\centering
\includegraphics[width=\linewidth]{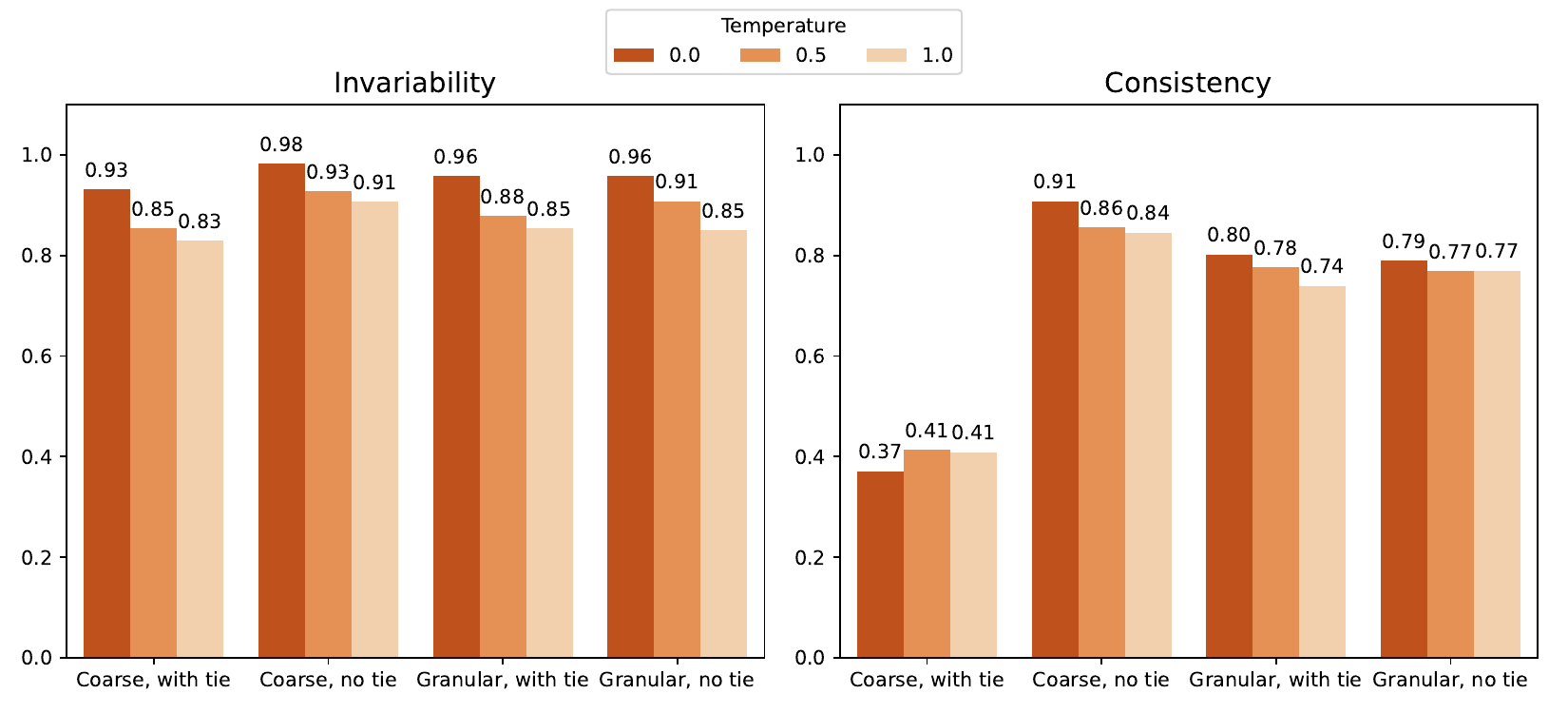}
\caption{Calibration scores for invariability (left) and pairwise positional consistency (PPC) (right), averaged over 20 LLMs and 10 repetitions for each under different pairwise comparison options.}
\label{fig:combined_consistency}
\end{figure*}

\newpage
\subsection{Results}

\textbf{temperature=0 is superior for reliable and consistent judgments}. The best average invariability and pairwise positional consistency across all 20 LLMs on the council is achieved with $temperature=0$. To our surprise, only $13/20$ models produce perfectly invariant ratings across all repetitions, even with $temperature=0$.

\textbf{Coarse or granular rating options?} Under $temperature=0$, the difference in invariability between using coarse and granular rating options is small (0.98 vs. 0.96). The difference in PPC is more stark (0.91 vs. 0.80), though still tolerable. We decide to proceed with granular rating options for the main experiment to maintain parity with Arena Hard \cite{11arenahard2024} and to give more weight to strong preferences in the final ELO calculation.

\textbf{To include or not include a tie}? Excluding the tie option slightly improves invariability and PPC at some temperatures, with negligible negative impact using $temperature=0$.

\textbf{Full findings.} Figure \ref{fig:combined_consistency} shows calibration scores for invariability and PPC, averaged over 20 LLMs and 10 repetitions for each under different pairwise comparison options. Table \ref{tab:calibrationfulltable} shows a detailed breakdown per LLM, using granular pairwise comparison options without ties and with $temperature=0$.

\newpage
\subsection{To CoT or not to CoT?}

Newer research suggests that Chain-of-Thought (CoT) prompting may degrade LLM performance on non-math and non-symbolic reasoning tasks, which may include ratings for simple pairwise comparisons \cite{sprague2024cotcotchainofthoughthelps}.

We assert our use of CoT prompting for two main reasons:

\begin{enumerate}
    \item It aligns with conventions in prior literature and arena-based LLM evaluation settings \cite{11arenahard2024, 3chatbotarenachiang2024chatbot}..
    \item CoT prompts generate reasoning traces, which we analyzed (Appendix \ref{sec:qualitativeanalysis}) to better understand the rationale behind the rankings.
\end{enumerate}

\subsection{Conclusion}

Our calibration study concludes with the decision to use granular comparison options without a tie to "force" judges to choose a side, thereby better distinguishing models, and with temperature 0.


\begin{table*}[ht]
\centering
\scalebox{0.75}{
\begin{tabular}{c|ccccc}
\toprule
LLM & Invariability & Conviction & Consistency & Position bias & Position bias \\
& & (strong votes) & & (first) & (second) \\
\midrule
claude-3-haiku & 100.0\% & 50.0\% & 50.0\% & 50.0\% & 50.0\% \\
claude-3-opus & 100.0\% & 50.0\% & 100.0\% & 0.0\% & 0.0\% \\
claude-3-sonnet & 100.0\% & 50.0\% & 100.0\% & 0.0\% & 0.0\% \\
command-r & 100.0\% & 50.0\% & 50.0\% & 50.0\% & 50.0\% \\
command-r-plus & 100.0\% & 50.0\% & 100.0\% & 0.0\% & 0.0\% \\
mistral-large & 100.0\% & 50.0\% & 50.0\% & 0.0\% & 0.0\% \\
mistral-medium & 100.0\% & 50.0\% & 50.0\% & 0.0\% & 0.0\% \\
mixtral-8x7b & 100.0\% & 25.0\% & 50.0\% & 0.0\% & 0.0\% \\
gpt-3.5-turbo-0125 & 82.5\% & 50.0\% & 95.0\% & 0.0\% & 0.0\% \\
gpt-4-0613 & 100.0\% & 50.0\% & 100.0\% & 0.0\% & 0.0\% \\
gpt-4-turbo-2024-04-09 & 92.5\% & 50.0\% & 100.0\% & 0.0\% & 0.0\% \\
gpt-4o-2024-05-13 & 95.0\% & 50.0\% & 100.0\% & 0.0\% & 0.0\% \\
qwen1.5-110B-Chat & 100.0\% & 50.0\% & 100.0\% & 0.0\% & 0.0\% \\
qwen1.5-32B-Chat & 95.0\% & 25.0\% & 100.0\% & 0.0\% & 0.0\% \\
qwen1.5-72B-Chat & 100.0\% & 50.0\% & 50.0\% & 0.0\% & 0.0\% \\
dbrx-instruct & 92.5\% & 50.0\% & 65.0\% & 50.0\% & 50.0\% \\
llama-3-70b-chat & 100.0\% & 50.0\% & 100.0\% & 0.0\% & 0.0\% \\
llama-3-8b-chat & 82.5\% & 50.0\% & 50.0\% & 50.0\% & 50.0\% \\
gemini-1.0-pro & 75.0\% & 25.0\% & 69.5\% & 0.0\% & 0.0\% \\
gemini-1.5-pro & 100.0\% & 50.0\% & 100.0\% & 0.0\% & 0.0\% \\
\bottomrule
\end{tabular}
}
\vspace{5mm}
\caption{Judging calibration results for 20 LLMs with using granular comparison options without a tie, with $temperature=0$. This is the same setting that was used for the paper's primary case study (Section \ref{sec:case_study}).}
\label{tab:calibrationfulltable}
\end{table*}

\clearpage
\section{Details on Reference Model Selection for the EI Case Study}
\label{app:reference_model_selection}
\raggedbottom
\begin{figure*}
\centering
\includegraphics[width=\linewidth]{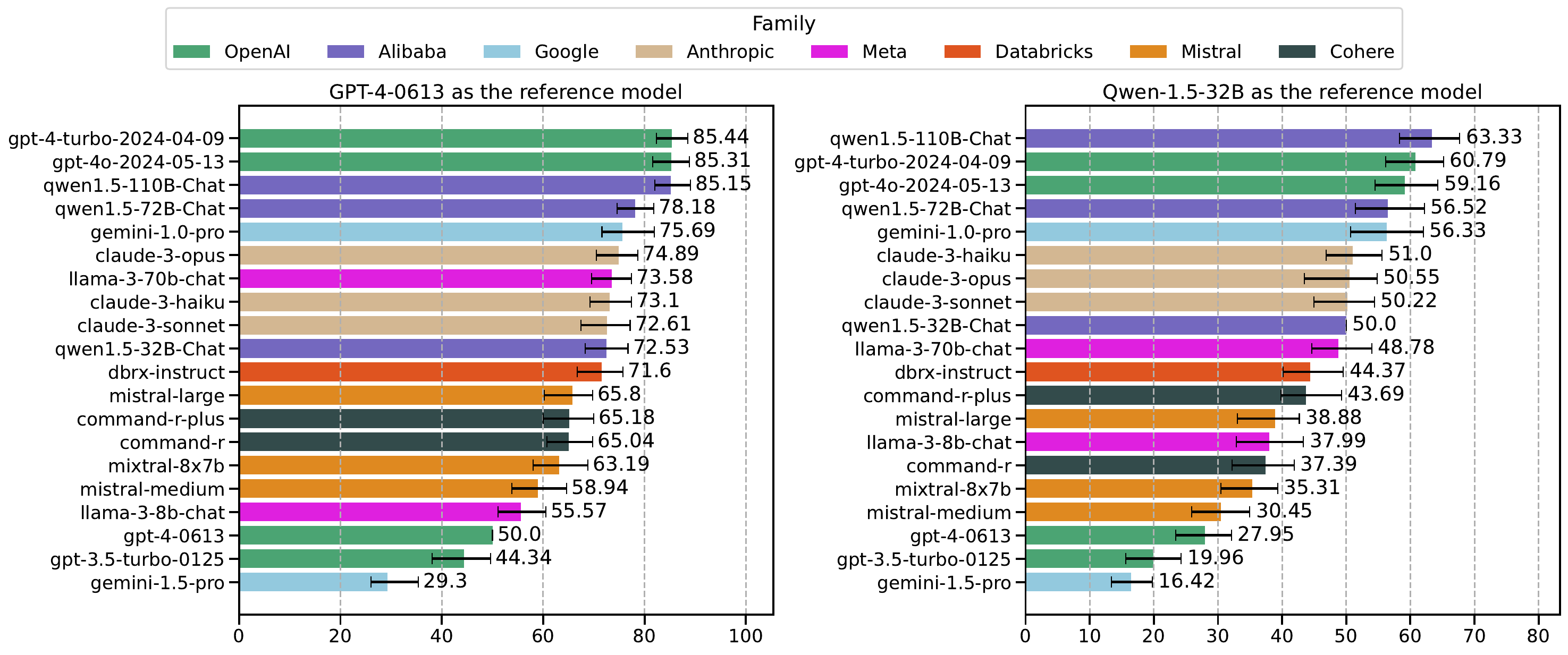}
\caption{Rankings from the dry run with 5\% of the data with GPT-4-0613 (left) or Qwen-1.5-32B-Chat (right) as the reference model. When the reference model is uncompetitive, separability among top performing models is poor.}
\label{fig:reference_model_selection_gpt4}
\end{figure*}

\subsection{Understanding the Bradley-Terry Procedure}

In a naive arena procedure with pairwise comparisons, every model's response is paired with every other model's response, requiring $O(n^2)$ comparisons for $n$ models. This approach is resource-intensive and impractical for large $n$. To circumvent the need for a quadratic number of comparisons, the Bradley-Terry algorithm~\cite{bradley1952rank} can be employed to determine expected win rates among a group of models, even without direct head-to-head battles between every pair.

The Bradley-Terry model offers a statistical method to estimate the relative strengths or ``abilities'' of items (models) based on pairwise comparison data. The key components of the model are:

\begin{itemize}
    \item \textbf{Skill Parameters}: Each model $i$ is assigned a positive real-valued parameter $\pi_i$, representing its skill or ability.
    \item \textbf{Win Probability}: The probability that model $i$ beats model $j$ is given by:
    \begin{equation}
    P(i \text{ beats } j) = \frac{\pi_i}{\pi_i + \pi_j}
    \end{equation}
\end{itemize}

\subsection*{Estimating Skill Parameters with Incomplete Data}

Even without direct comparisons between every pair of models, we can estimate the skill parameters $\{\pi_i\}$ using available comparison data through the following steps:

\begin{enumerate}
    \item \textbf{Collect Pairwise Comparisons}: Perform a subset of all possible pairwise comparisons, resulting in observed outcomes (which model won against which).
    \item \textbf{Set Up Likelihood Equations}: The likelihood of the observed data, given the skill parameters, is formulated based on the Bradley-Terry probabilities.
    \item \textbf{Maximum Likelihood Estimation (MLE)}:
    \begin{itemize}
        \item \textbf{Objective}: Find the set of skill parameters $\{\pi_i\}$ that maximize the likelihood of the observed data.
        \item \textbf{Process}: Solve the likelihood equations derived from the comparisons to estimate the $\{\pi_i\}$.
    \end{itemize}
    \item \textbf{Compute Expected Win Rates}:
    \begin{itemize}
        \item With the estimated skill parameters, calculate the expected probability that any model $i$ beats any model $j$ using:
        \begin{equation}
        P(i \text{ beats } j) = \frac{\pi_i}{\pi_i + \pi_j}
        \end{equation}
        \item \textbf{Note}: This computation is valid even for pairs of models that were not directly compared.
    \end{itemize}
\end{enumerate}

The Bradley-Terry algorithm enables us to estimate the expected win rates between all pairs of models without requiring a prohibitive number of direct comparisons by:

\begin{itemize}
    \item Assigning a skill parameter to each model.
    \item Using observed pairwise comparisons to estimate these parameters via maximum likelihood estimation.
    \item Calculating the probabilities of any model defeating another using the estimated parameters.
\end{itemize}

\subsection{What is the reference model?}

The reference model is the model whose responses are used across all pairwise comparisons. In this way, the reference model serves as a shared anchor for all models to be evaluated against. For example, if we have models W, X, Y, and Z, and use model Z as the reference, the pairwise comparisons are: (W vs. Z), (X vs. Z), and (Y vs. Z). The use of a reference model requires $O(n)$ comparisons for $n$ models.


While the Bradley-Terry algorithm doesn't require a shared reference model, using one ensures a more consistent win rate estimation across all model pairs.

\newpage
\subsection{Selecting our reference model}

In our arena-based LMC EI case study, we use a single reference model, following the approach of other arena-based benchmarks like Chatbot Arena Hard \cite{11arenahard2024} and Alpaca Eval \cite{dubois2024lcalpacaeval}, which used GPT-4-0314 and GPT-4-turbo, respectively. It is unclear how \cite{11arenahard2024} and \cite{dubois2024lcalpacaeval} chose their reference models, but we will explain our choice.

We initially used GPT-4-0613 from OpenAI as our reference model. In a dry run, we observed poor separability in ELO scores (Figure \ref{fig:reference_model_selection_gpt4}). Other models won very often against GPT-4-0613, (we believe this was due to its average response length of 173 words, which was much less than the suggested 250-word limit (Table \ref{tab:mainexperiment})). When the reference model is uncompetitive, ELO scores for other models get inflated, reducing ranking separability.

This led us to a key realization that for better separability, the reference model should have a varied mix of wins and losses against other models. We chose Qwen-1.5-32B as an alternative because it ranked mid-range in the initial dry run (Figure \ref{fig:reference_model_selection_gpt4}). Redoing the dry run with Qwen-1.5-32B improved separability substantially, so we proceeded to use it for the main experiment.

A more systematic approach to selecting the reference model—like having more randomized matchups, or using multiple reference models—could strengthen our arena-based LLM evaluation methods. However, this is beyond our paper's scope and budget.

The choice of Qwen-1.5-32B as our reference model is not cherrypicking (the opposite actually). We believe that the risk of invalidating our key findings due to this choice is low. However, we speculate in our main \hyperref[sec:results]{Results} that choosing the smaller Qwen model may have given an outsized advantage to larger models in the same family (Qwen-1.5-110B in particular).

\clearpage
\section{Human Evaluation}\label{app_sec:human_study}

\begin{figure*}[t]
\centering
\frame{\includegraphics[width=\linewidth]{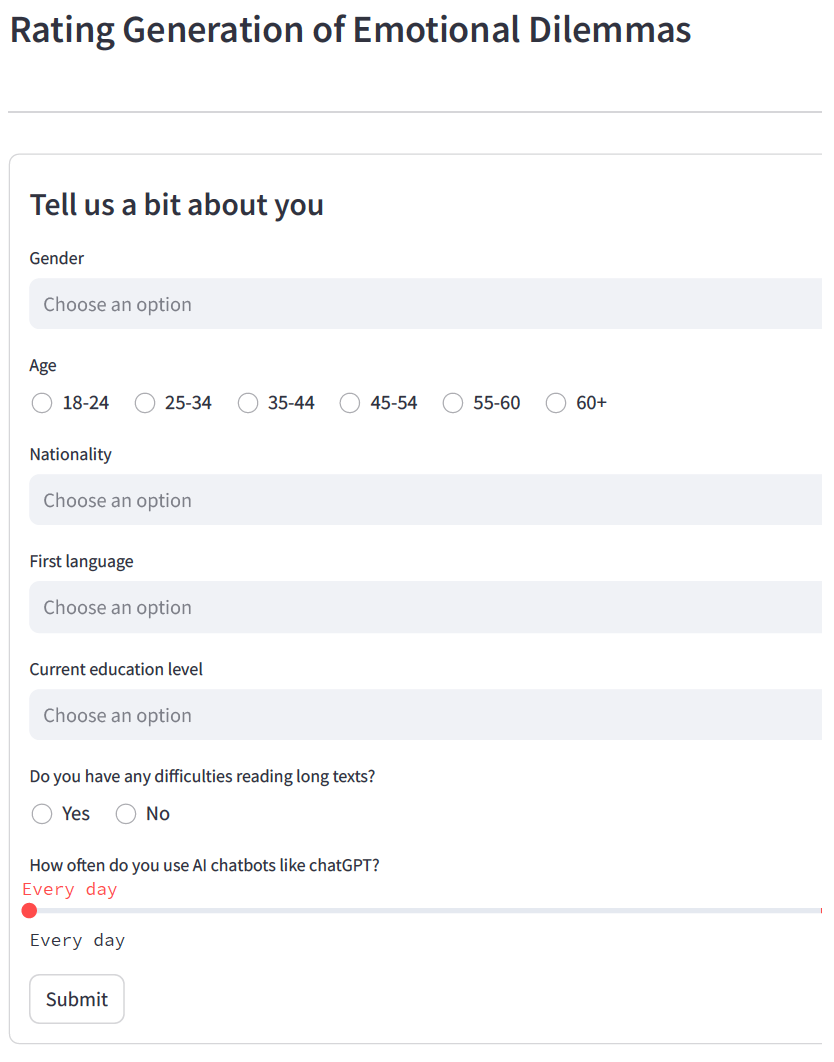}}
\caption{Participant demographic questionnaire.}
\label{fig:streamlit_demographics}
\end{figure*}

During registration for our experiments, all candidates provided their demographic details (see Figure \ref{fig:streamlit_demographics}). Additionally, we required each candidate to complete a questionnaire measuring their level of empathy, sourced from \cite{jolliffe2006development}.
All candidates were informed of the purpose of our study. 142 participants completed the survey but after removing those who failed attention checks, 102 participants remain. 
Each dilemma pair and response was rated by 11 participants on average, after removing malicious participants. Each participant was compensated \textsterling9.00 per hour. 

\paragraph{Participant demographics:} All participants are over 18 years old. Our sample is made up of 53 women, 46 men, and one non-binary identifying individual. 84 of our participants were from the United Kingdom, 14 from the United States and two from other English-speaking countries; all were native English speakers. With regards to their use of AI chatbots, 23 report using them every day or nearly every day, 48 sometimes, four rarely and only four report never using them. None report having difficulties reading long texts.

\paragraph{Data quality assurance:} Because the task is both difficult and subjective, we take a two-fold approach to ensure quality data: (1) we ask participants to provide demographics which we cross-reference with data from Prolific; and (2) we use two repeated dilemmas as test questions, checking for self-agreement. We allow participants to shift slightly to account for the lack of ties: a participant may slightly prefer one response then another, but not prefer one strongly then prefer a different response the following time. We remove data from workers who lack this consistency. This results in 102 unique participants in the final set.

\begin{figure*}[t]
\centering
\frame{\includegraphics[width=\linewidth]{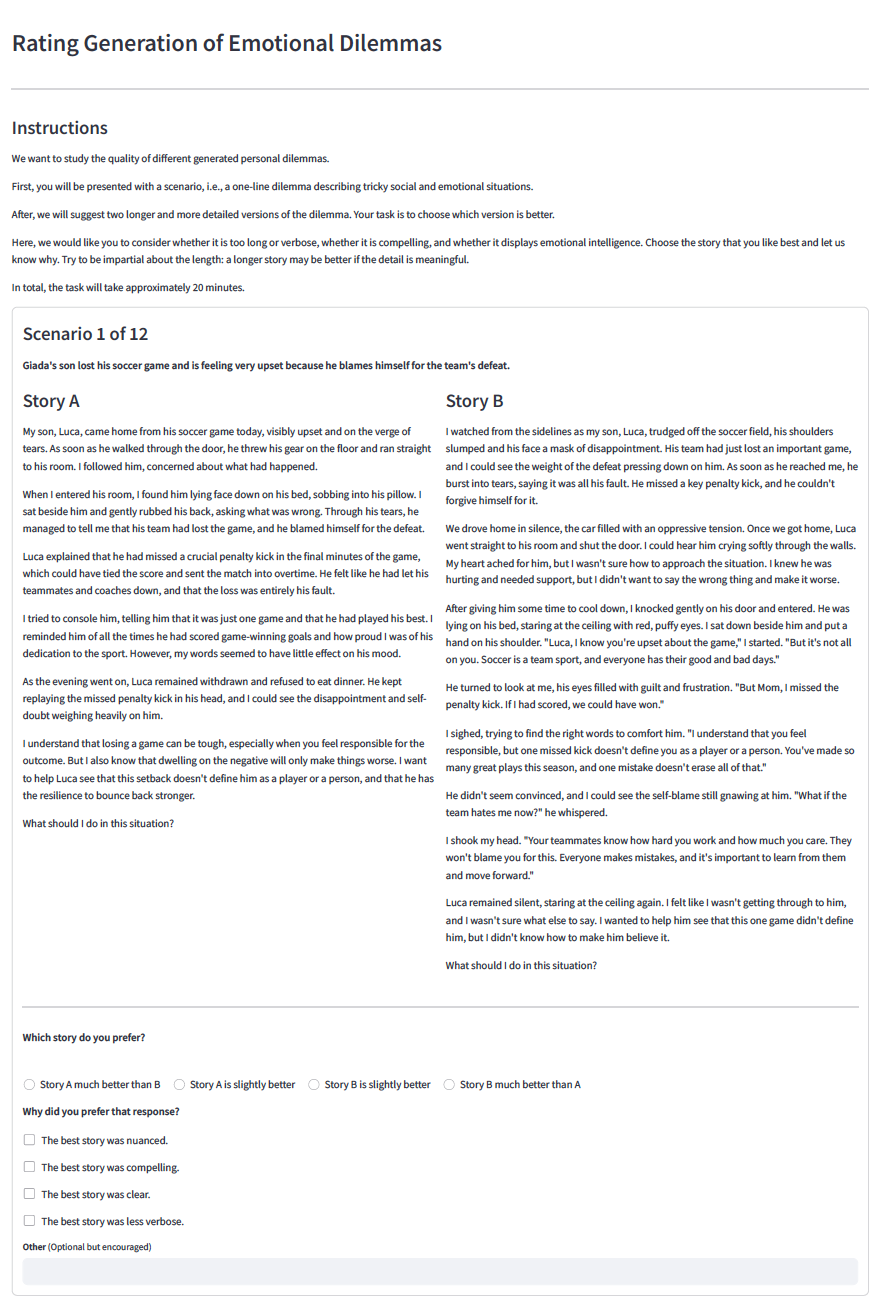}}
\caption{Participant guidelines for rating the generation of dilemmas.}
\label{fig:streamlit_rating_generation_dilemmas}
\end{figure*}

\begin{figure*}[t]
\centering
\frame{\includegraphics[width=0.8\linewidth]{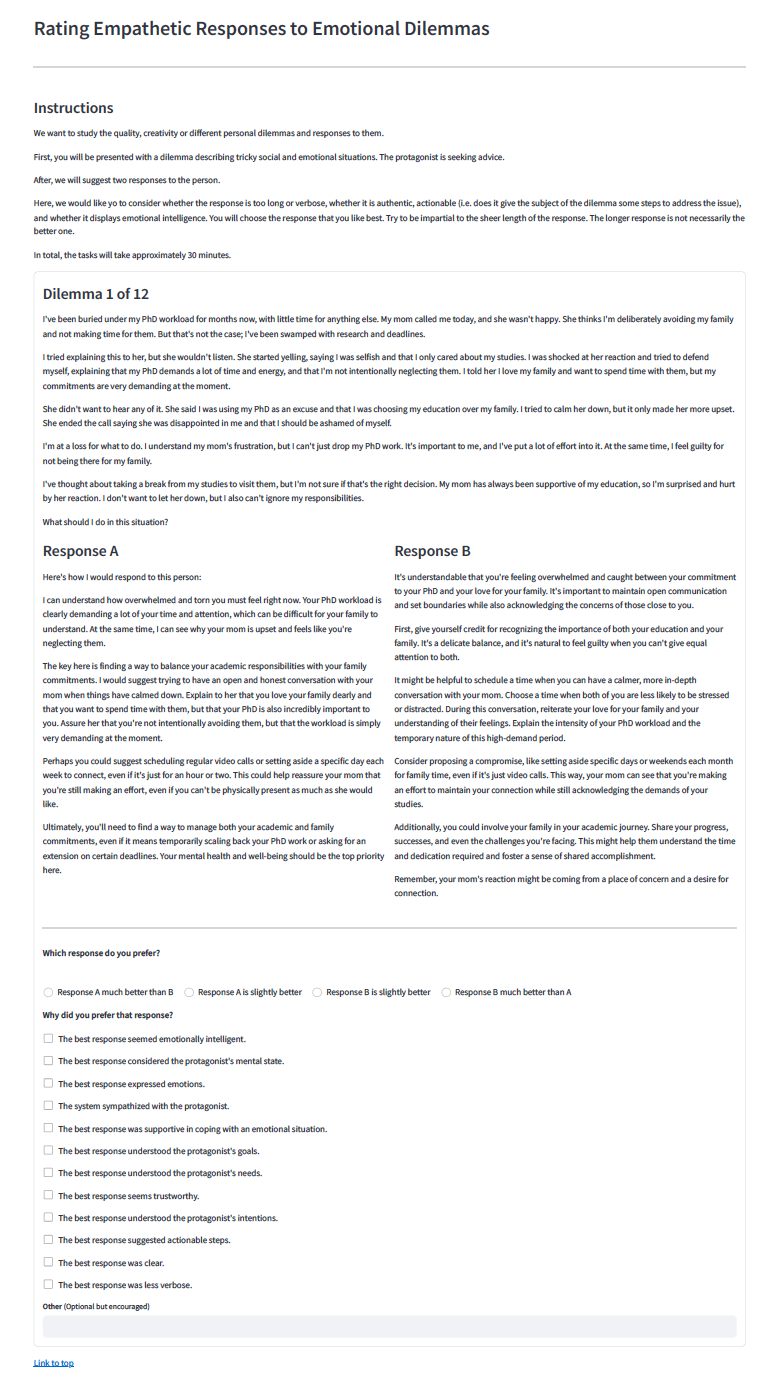}}
\caption{Participant guidelines for rating the responses to dilemmas.}
\label{fig:streamlit_rating_responses}
\end{figure*}

We provide the participant
guidelines in Figures \ref{fig:streamlit_rating_generation_dilemmas} and \ref{fig:streamlit_rating_responses}.

\paragraph{Measuring perceived empathy:} We adapt our feedback from the scale proposed by \cite{schmidmaier2024pets}, which is designed to assess systems with which the users have interacted. We exclude question E5 from the original questionnaire and rephrase them to fit our experiment. The statements are detailed in Table \ref{tab:pets_scale}.

\begin{table*}[ht]
\begin{tabular}{rl}
   E1  & The best response considered the protagonist's mental state. \\
   E2 (EQ)  & The best response seemed emotionally intelligent. \\
   E3  & The best response expressed emotions. \\
   E4  & The best response sympathized with the protagonist. \\
   E5  & The best response was supportive in coping with an emotional situation. \\
   U1  & The best response understood the protagonist's goals. \\
   U2  & The best response understood the protagonist's needs. \\
   U3  & The best response seems trustworthy. \\
    U4 & The best response understood the protagonist's intentions.
\end{tabular}
\caption{Adapted PETS scale for our study.}
\label{tab:pets_scale}
\end{table*}

\begin{figure*}[ht]
    \centering
\includegraphics[width=\linewidth]{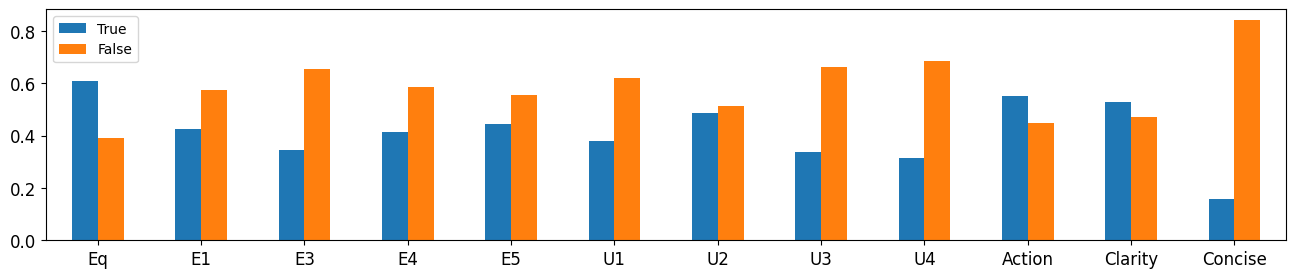}
    \caption{Proportion of times users found the statements in the PETS questionnaire to be true about the winning response. The corresponding statements are shown in Table \protect\ref{tab:pets_scale}. E2 in the questionnaire is equivalent to out EQ question (shown first) so it is not included.}
    \label{fig:human_feedback}
\end{figure*}

\clearpage
\section{More Details on Comparison to Other Leaderboards}
\label{app:comparison_to_other_leaderboards_details}


\begin{figure*}
    \centering
    \includegraphics[width=1.0\linewidth]{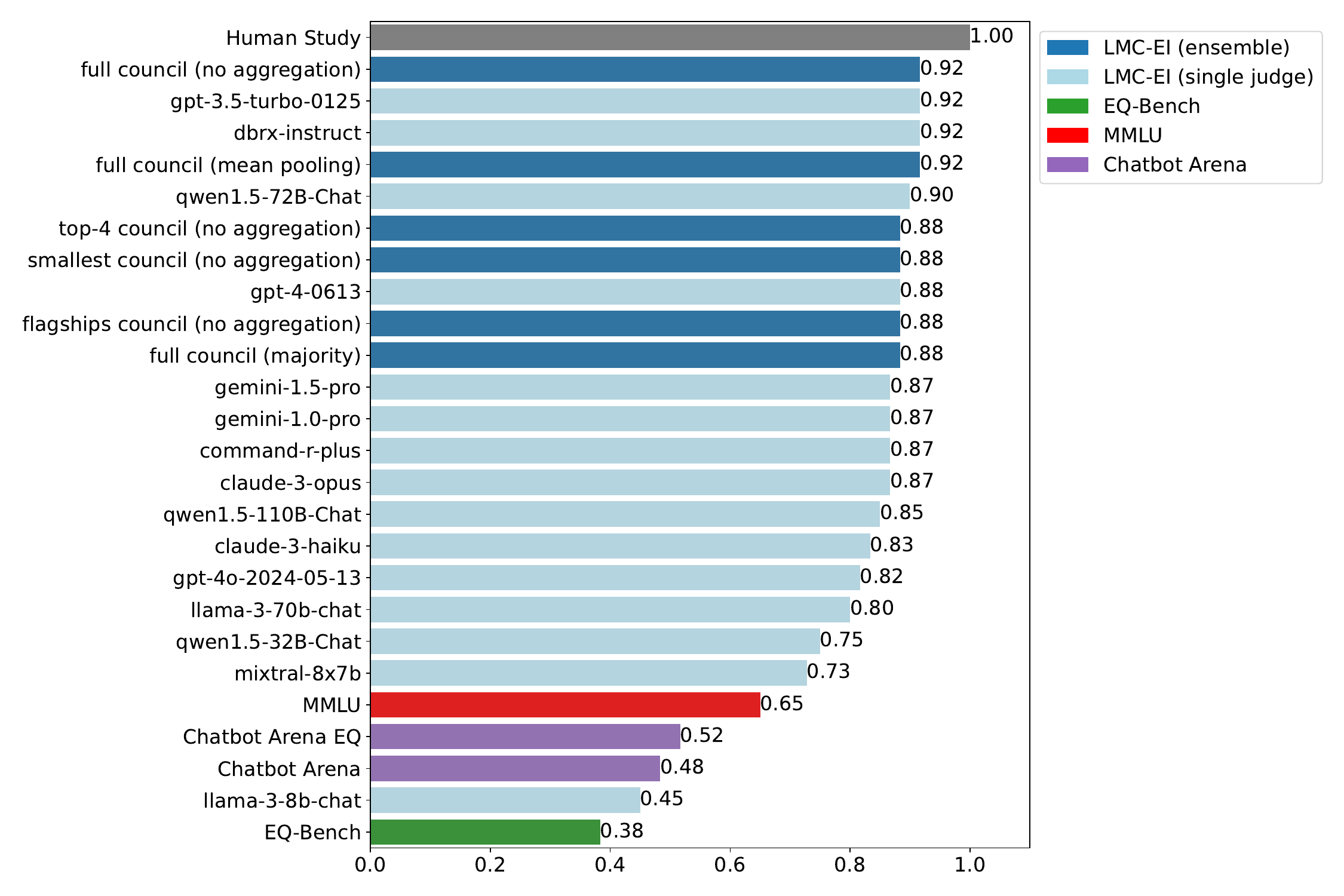}
    \caption{Spearman ranking correlations with our EI Human Study (Appendix \ref{app_sec:human_study}) for 
benchmark scores for 9 LLMs (listed in Figure \ref{fig:human-rankings}). We include the correlation scores for all individual LLM judges on the LMC, as well as correlations for hand-curated sub-councils discussed in Section \ref{sec:oligarchical_councils}.}
    \label{fig:leaderboard_comparison_expanded}
\end{figure*}

\begin{figure*}
    \centering
    \includegraphics[width=1.0\linewidth]{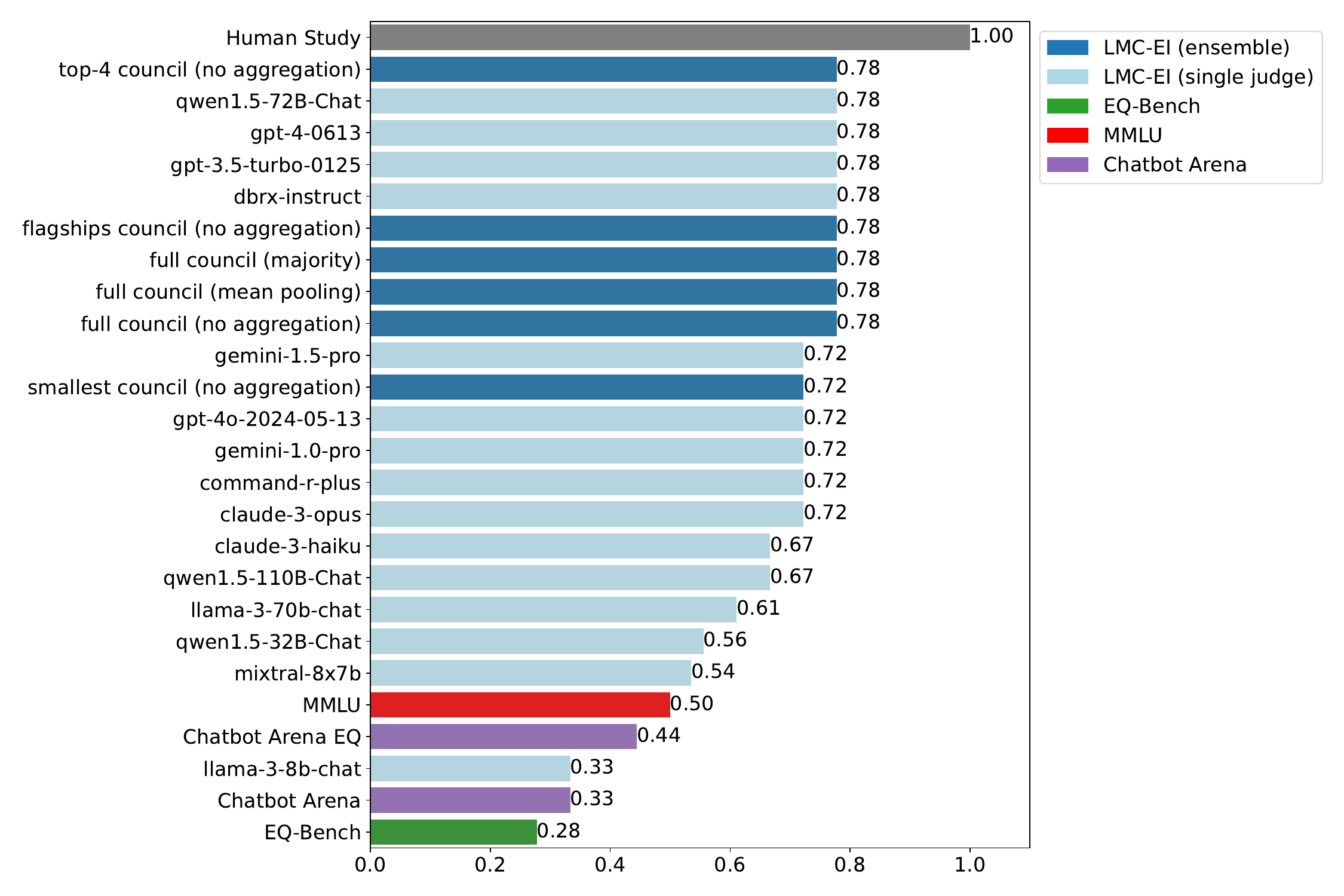}
    \caption{Kendall-Tau ranking correlations with our EI Human Study (Appendix \ref{app_sec:human_study}) for 
benchmark scores for 9 LLMs (listed in Figure \ref{fig:human-rankings}). We include the correlation scores for all individual LLM judges on the LMC, as well as correlations for hand-curated sub-councils discussed in Section \ref{sec:oligarchical_councils}.}
    \label{fig:leaderboard_comparison_expanded_kendall}
\end{figure*}

\paragraph{Mining an EQ subset of Chatbot Arena}

Chatbot Arena includes many prompts that may have little to do with EI, so it may be unsurprising that the overall correlation with our human study is low (0.48) (Figure \ref{fig:leaderboardcomparison}). In this section, we outline a procedure to produce an EI-based re-ranking of LLMs based on an EI-targeted subset of Chatbot Arena prompts.

\begin{figure*}
    \centering
    \includegraphics[width=1.0\linewidth]{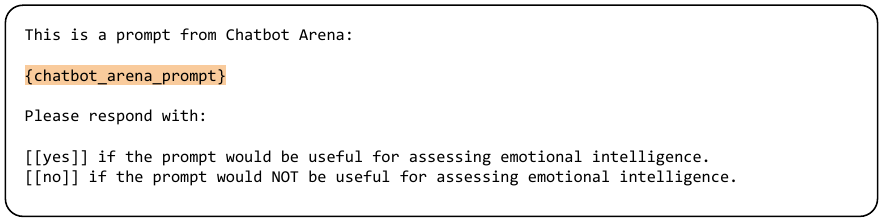}
    \caption{Prompt template used to classify if a prompt would be good for emotional intelligence or not.}
    \label{fig:prompt_template_chatbot_arena_eq_subset}
\end{figure*}

A publicly released subset of Chatbot Arena’s prompts are available online.\footnote{\url{https://huggingface.co/datasets/lmsys/chatbot\_arena\_conversations}} This dataset has 33K unique prompts. Because there are no fine-grained categories that allow us to easily slice the leaderboard by performance on specific questions, we send all 33K unique prompts to a reasonably competent LLM, llama-3.1-8b \cite{llama31}, one prompt at a time, to assess whether the prompt is EI-related or not. The prompt template used to perform a classification of EI-relatedness is in Figure \ref{fig:prompt_template_chatbot_arena_eq_subset}.

This results in 2680 prompts (\textasciitilde8\%) flagged to be potentially useful for assessing EI. While a stronger LLM could be used to do this EI-relatedness flagging, most of the examples we spot-checked looked reasonably related to EI. Here are some examples:

\squishlist
    \item “Why did my parent not invite me to their wedding?”
    \item “Please write an email to a University Professor to tell them that I will not be attending their PhD program.”
    \item “I'm feeling sad. Can you tell me a joke to cheer me up?”
\squishend

The Chatbot Arena dataset with these prompts does not include generated responses from the 9 LLMs that were used in our Human Study. Therefore, we generate new outputs for the 9 models and subsequently score the generated answers with GPT-4o-mini. Due to budget constraints, we only use 100 prompts of the 2680 subset.

Following the same procedure as the LMC, we assess responses in a pairwise fashion with position flipping using Qwen-1.5-32B's responses as the reference and with GPT-4o-mini \cite{gpt4o} as the judge, mimicking the single judge design of Chatbot Arena Hard \cite{11arenahard2024}. This results in 100 x 8 x 2 = 1600 ratings.

\paragraph{Correlation score improves over vanilla Chatbot Arena, but still significantly lower than the Language Model Council.} The Spearman correlation with our human study is listed in Figure \ref{fig:leaderboardcomparison} and Figure \ref{fig:leaderboard_comparison_expanded}. The correlation score improves when using the EI-specific subset of Chatbot Arena (0.48 -> 0.52) compared to vanilla Chatbot Arena scores. However, the overall correlation is still significantly worse than LMC, which has a score of 0.92. This reaffirms the idea that the narrowness of our task may be the dominant basis for high agreement with the human-established ranking of our EI task, as neither EQ-Bench (an EI-centric multiple choice question test) nor re-ranking models with an EI-targeted subset of Chatbot Arena achieve nearly as high of a correlation with our human study.

\clearpage
\section{Qualitative Analysis: What Makes a Response Preferred Over Another?}
\label{sec:qualitativeanalysis}

\begin{table}[ht!]
    \centering
    \scriptsize
    \begin{minipage}{0.47\textwidth}
        \centering
        \begin{tabular}{llr}
        \toprule
        \textbf{Reason 1} &\textbf{Reason 2} &\textbf{Correlation} \\
        \midrule
        less verbose &more succinct &0.650 \\
        better structured &more structured &0.584 \\
        easier to follow &better structured &0.520 \\
        easier to follow &more structured &0.468 \\
        less verbose &more direct &0.450 \\
        more understanding &more empathetic &0.418 \\
        more clear &better structured &0.387 \\
        more direct &more succinct &0.349 \\
        easier to follow &more clear &0.348 \\
        more gentle &more soft &0.337 \\
        \bottomrule
        \end{tabular}
        \vspace{1mm}
        \caption{Top 10 positive correlations.}\label{tab:top_10_positive}
    \end{minipage}
    \hfill
    \vspace{5mm}
    \begin{minipage}{0.47\textwidth}
        \centering
        \begin{tabular}{llr}
        \toprule
        \textbf{Reason 1} &\textbf{Reason 2} &\textbf{Correlation} \\
        \midrule
        more comprehensive &less verbose &-0.276 \\
        less verbose &more detailed &-0.227 \\
        more comprehensive &more direct &-0.202 \\
        more comprehensive &more succinct &-0.197 \\
        more detailed &more succinct &-0.196 \\
        more comprehensive &more focused &-0.161 \\
        more detailed &more direct &-0.148 \\
        more suggestions & less verbose &-0.144 \\
        more understanding &more actionable &-0.139 \\
        less verbose &more nuanced &-0.135 \\
        \bottomrule
        \end{tabular}
        \vspace{1mm}
        \caption{Top 10 negative correlations.}\label{tab:top_10_negative}
    \end{minipage}
\end{table}

\subsection{Motivation}

Several arena-based benchmarks (ours included) have demonstrated that a clear ranking among LLMs \textit{can} be established, but there is not much understanding as to \textit{why} the rankings are the way they are. For example, platforms like Chatbot Arena do not clarify how factors like feel and style are weighed against correctness \cite{wei2024evals}, and while many evaluation systems like AlpacaEval \cite{dubois2024lcalpacaeval} or MT-Bench \cite{zheng2024judging} tout chain-of-thought (CoT) prompting \cite{wei2022cot} to improve the explainability of ratings by LLM judges, these justifications are left unanalyzed. 

We describe a systematic approach to analyzing the CoT reasoning traces from the Language Model Council in our EI case study to better understand the qualitative aspects of what makes a response to an emotional interpersonal conflict more desirable.

\subsection{Reasoning trace themes extraction procedure}
\begin{figure*}[ht]
\centering
\scalebox{0.9}{
\includegraphics[width=\linewidth]{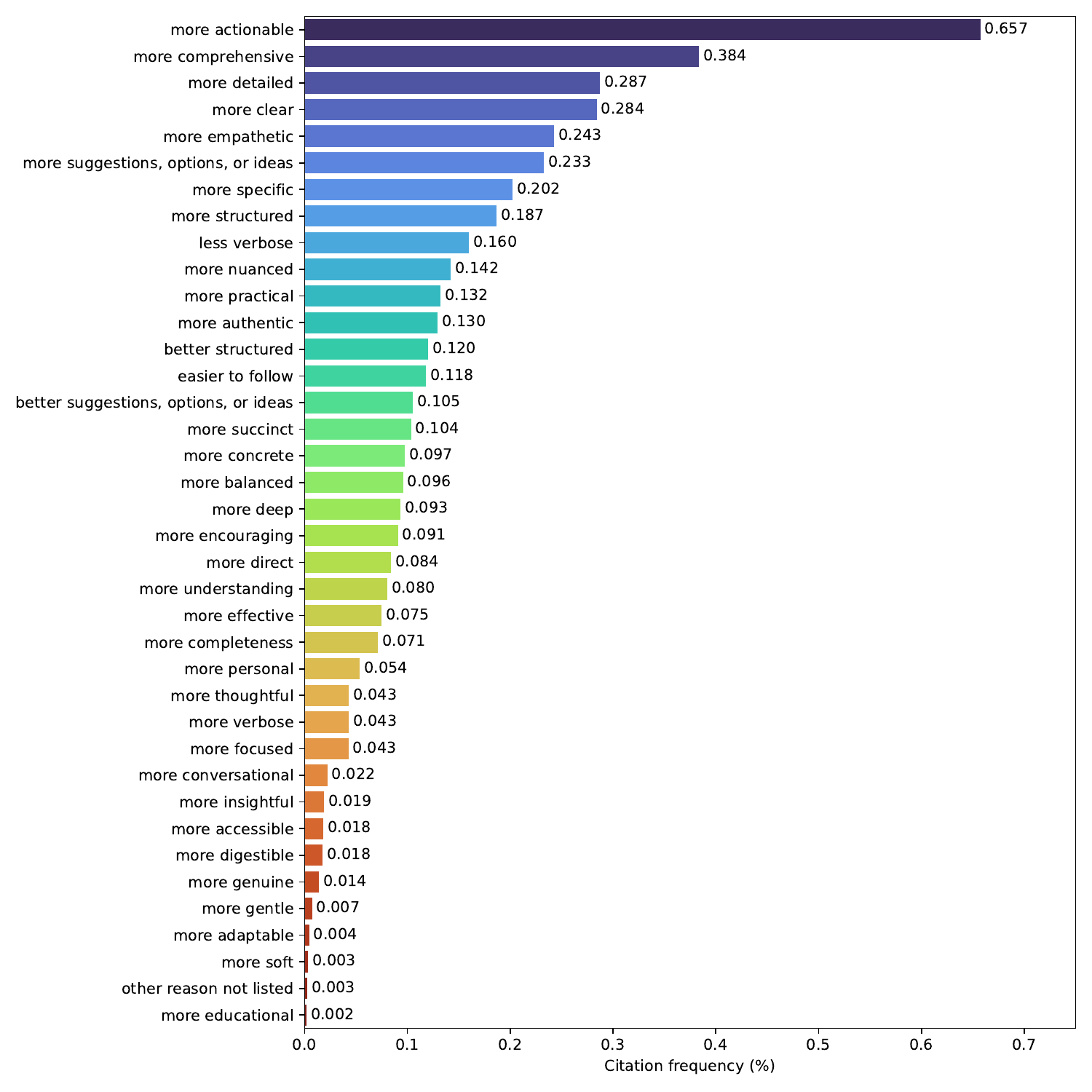}
}
\caption{Citation frequency of 38 qualitative reasons why the winning response was preferred.}
\label{fig:qualitative-reasons-list}
\end{figure*}

\begin{figure*}[ht]
\centering
\includegraphics[width=\linewidth]{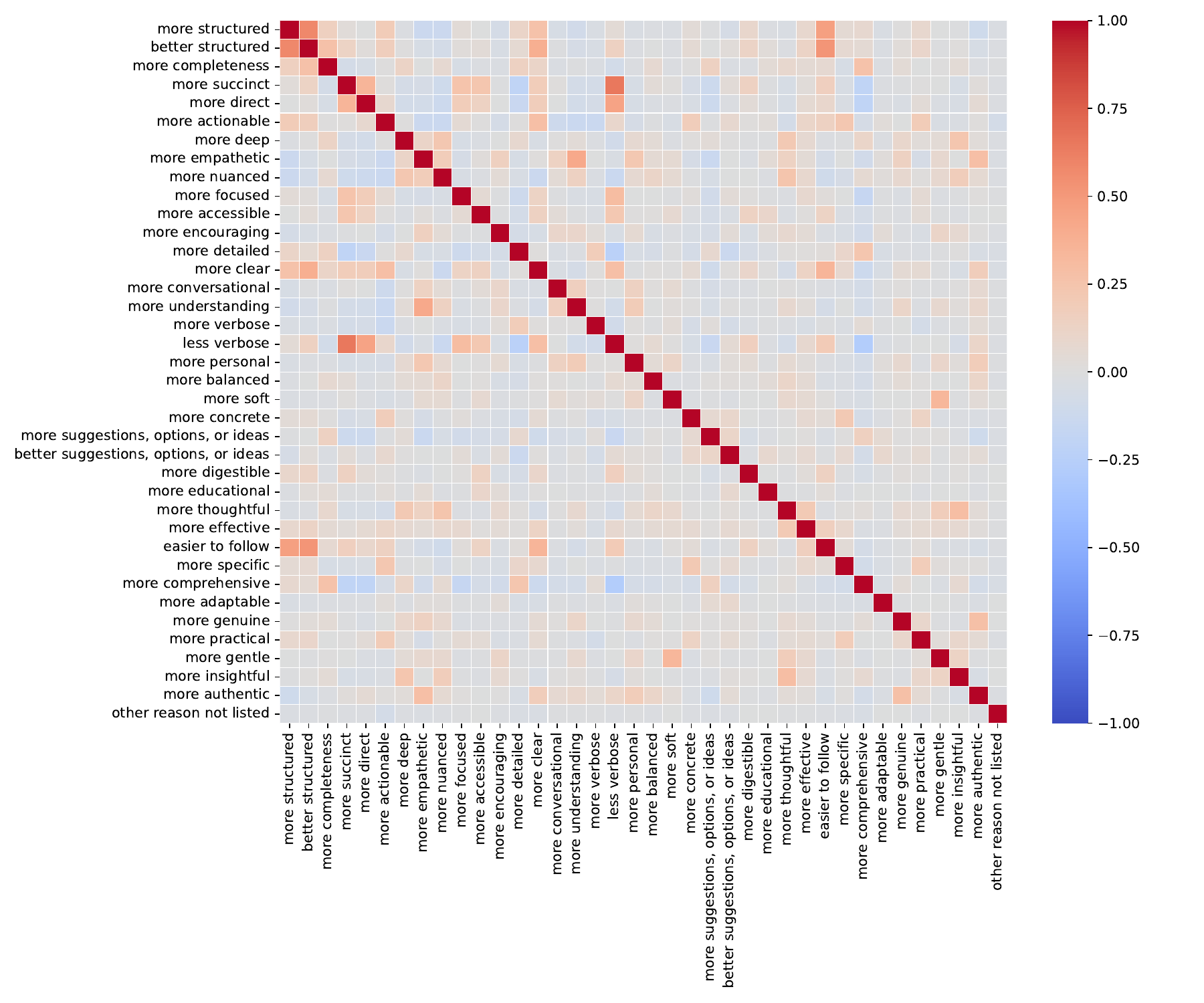}
\caption{Spearman correlation matrix of cited reasons why the winning response was preferred.}
\label{fig:qualitative-reasons-heatmap}
\end{figure*}

\begin{figure*}[ht]
\centering
\includegraphics[width=\linewidth]{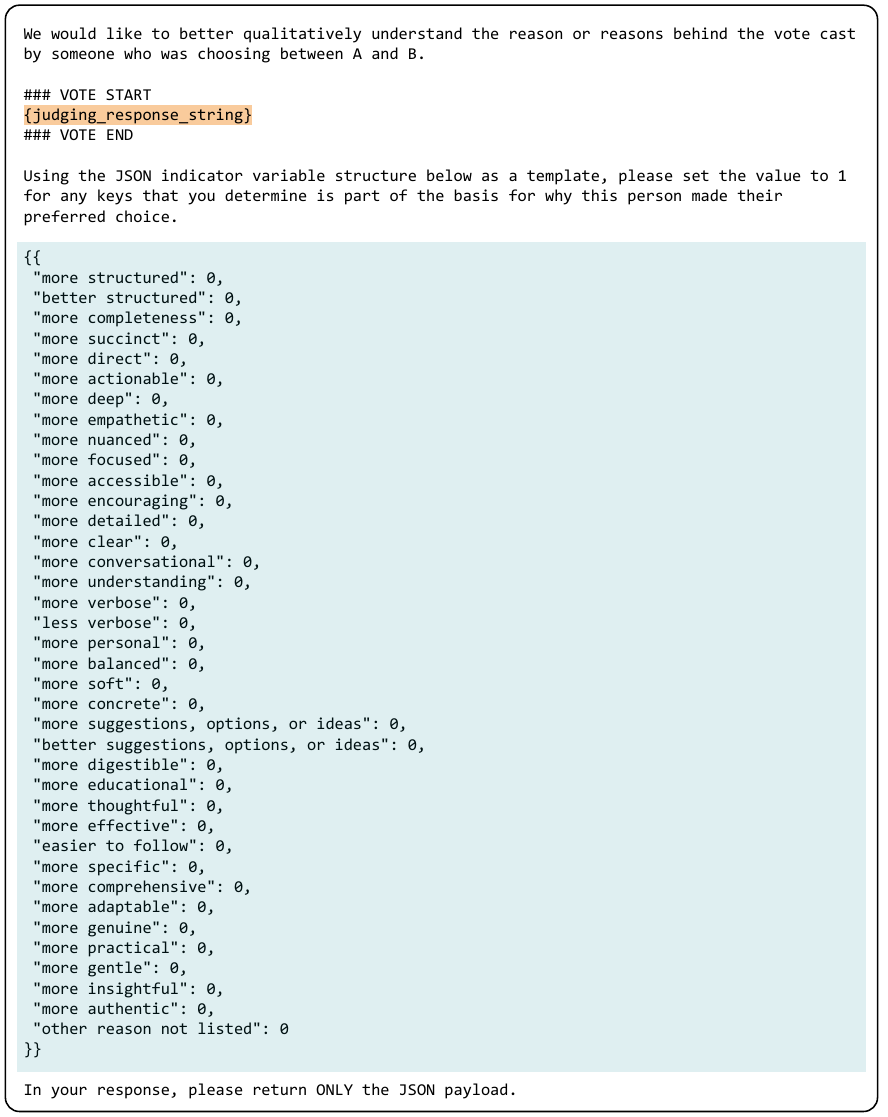}
\caption{Prompt used to map explanations in pairwise ratings to a rich, fixed set of qualitative reasons. The 38 seed qualitative reasons used in the prompt come from manual review of 50 randomly selected pairwise ratings in the main experiment involving the full council of 20 LLMs.}
\label{fig:prompt-qualitative-assessment}
\end{figure*}

First, we manually examine a random sample of 50 reasoning traces, identifying 38 coarse reasons for preferences (e.g., “more practical”). The full list is in Figure \ref{fig:qualitative-reasons-heatmap}. Then, we use a strong LLM (GPT-4o) to map a larger sample of 1K explanations to these predefined reasons (prompt in Figure \ref{fig:prompt-qualitative-assessment}). The 1K sample includes ratings from all 20 LLM judges. Detailed reason citation frequencies are listed in Figure \ref{fig:qualitative-reasons-list}.

\subsection{Subjectivity of defining and assigning themes}

We acknowledge that there is an element of subjectivity in defining coarse reason categories and determining the cutoff for creating new categories. However, this is low risk for several reasons. 

\squishlist
    \item [1.] The categorization process is intended to extract a broad sense of the most frequent themes in the reasoning traces provided by LLM judges, rather than to establish a definitive taxonomy. 
    \item [2.] The actual counting of occurrences for each reason was performed by a separate strong LLM judge, which we do not control.\footnote{We spot checked that the strong LLM judge was reasonable when interpreting a reasoning trace and selecting relevant themes for it. However, we also acknowledge that the act of bucketing is subject to interpretation.}
    \item [3.] A catch-all “other reason not listed” option was provided, though it was rarely used (only 0.3\% of the time).
\squishend

Our primary objective is only to gain a general understanding of the themes driving LLM judges' preferences, so full precision is not required.

\subsection{Results and discussion}

We find that the ratings of LLM judges are almost always based on multiple indicators (4.5 ± 2.4 on average). "More actionable" is the most cited reason, which aligns with the action-oriented framing of our emotional intelligence test. "Structure," "clarity," and "specificity" dominate the top 10 reasons. "More gentle" and "more soft" are cited least, contrasting with "more practical" (\#11) and "more authentic" (\#12). Longer responses ("more comprehensive” \#2, "more detailed" \#3) are more popular than brevity ("less verbose" \#9).

Among the top positively correlated reasons (Figure \ref{tab:top_10_positive}), "better structured" frequently co-occurs with "more structured" (correlation: 0.584) and "easier to follow" (correlation: 0.520), indicating that brevity, structure, and clarity are often evaluated together. The co-occurrence of "more understanding" and "more empathetic" (correlation: 0.418) suggests that judges consider empathy and understanding as closely related, though not identical. In Figure \ref{tab:top_10_negative}, responses that are comprehensive often sacrifice directness and conciseness. "More comprehensive" is inversely correlated with "less verbose" (-0.276), "more succinct" (-0.197), and "more direct" (-0.202). The negative correlation between "more detailed" and "more focused" (-0.161) suggests that providing excessive detail can reduce a response's focus.

We also examine feedback from the human study: we find that users generally find that the best responses display emotional intelligence (60.9\%), are actionable (55.1\%) and clear (52.9\%). In contrast, participants reported the best response is concise only 15.9\% of the time, suggesting language efficiency is less of a determining factor for humans. Moreover, we find little support for empathy: the participants did not find any of the statements in the PETS questionnaire \cite{schmidmaier2024pets} to ring any truer for the winning response. Participants who provided verbal feedback emphasized specificity to the situation, clear examples of how to proceed, and a tone that was not too formal.

\newpage
\subsection{Conclusion}

Our procedure demonstrates a systematic method to drill into reasoning traces from LLMs to better interpret the preferences of LLM judge ratings. For our paper's EI case study, direct feedback from human participants and LLM reasoning trace theme extractions from LLM judge explanations share a consistent theme: longer responses that are clear, detailed, and actionable are more preferred when responding to emotional interpersonal conflicts.



\clearpage
\section{Quantifying the Stability of a Benchmark}
\label{app:merv}

There is natural variance in the ranking of other models, particularly when LLM judges are involved. 

To quantify the robustness of a ranking, we create a new metric, \textbf{M}ean \textbf{E}xpected \textbf{R}ank \textbf{V}ariance (\textbf{MERV}). Conceptually, this can be thought of as the expected ordinal swing of the average respondent’s rank. 

\newpage
\subsection{Mathematical Definition of Mean Expected Rank Variance (MERV)}

Consider a set of \( m \) models evaluated over \( n \) random trials to account for natural variance in model performance. Let \( R_{ij} \) denote the rank of model \( i \) in trial \( j \), where \( i = 1, 2, \dots, m \) and \( j = 1, 2, \dots, n \).

For each model \( i \), the Expected Rank Variance \( \text{ERV}_i \) is defined as the variance of its ranks across the \( n \) trials:

\[
\text{ERV}_i = \frac{1}{n - 1} \sum_{j=1}^{n} \left( R_{ij} - \overline{R}_i \right)^2,
\]

where \( \overline{R}_i \) is the mean rank of model \( i \):

\[
\overline{R}_i = \frac{1}{n} \sum_{j=1}^{n} R_{ij}.
\]

The \textbf{Mean Expected Rank Variance (MERV)} is then defined as the mean of the ERV across all respondents:

\[
\text{MERV} = \frac{1}{m} \sum_{i=1}^{m} {\text{ERV}_i}.
\]

\newpage
\subsection{Understanding MERV}

MERV has an intuitive interpretation. It directly tells us how much the rank of an average respondent is expected to swing, expressed in ordinal positions. A MERV of 3 means that an average respondent’s rank could shift by up to 3 positions in a new trial while a MERV of 0 signifies perfect deterministic-like stability.

MERV is sensitive to changes in relative rankings, providing a good metric for evaluating leaderboard robustness when the primary concern is how consistent relative positions are across different trials.

Because MERV focuses entirely on ranks, it may ignore significant changes in raw performance scores. A small ordinal swing (low MERV) might still hide large variations in actual scores. Conversely, large MERV values could come from minor performance changes, especially if ranks are tightly clustered. If one respondent has highly volatile ranks while others remain stable, MERV might underestimate the overall instability due to averaging.

While MERV provides useful information about rank variability, it says little about the underlying confidence or statistical significance of rank differences.

\newpage
\subsection{Comparison with Separability}

Separability, in contrast, measures the percentage of respondent pairs with completely non-overlapping confidence intervals, often derived from bootstrapping \cite{11arenahard2024}. It quantifies the statistical significance of performance differences, focusing on how distinct the rankings of respondents are in terms of performance intervals.

Both MERV and separability address aspects of reliability, but while MERV is about rank stability, separability is about the stability of the performance margins between respondents. Both give insight into robustness, though from different perspectives.

Separability provides a more nuanced view of the data by considering whether rank changes are statistically significant, whereas MERV gives a more direct sense of how often rankings change.

\clearpage
\section{Prompt Templates}
\label{sec:all-prompts}

In this section, we list all prompts used, including prompts for synthetic expansion, dilemma response, and judging.

\begin{figure*}[ht]
\centering
\includegraphics[width=\linewidth]{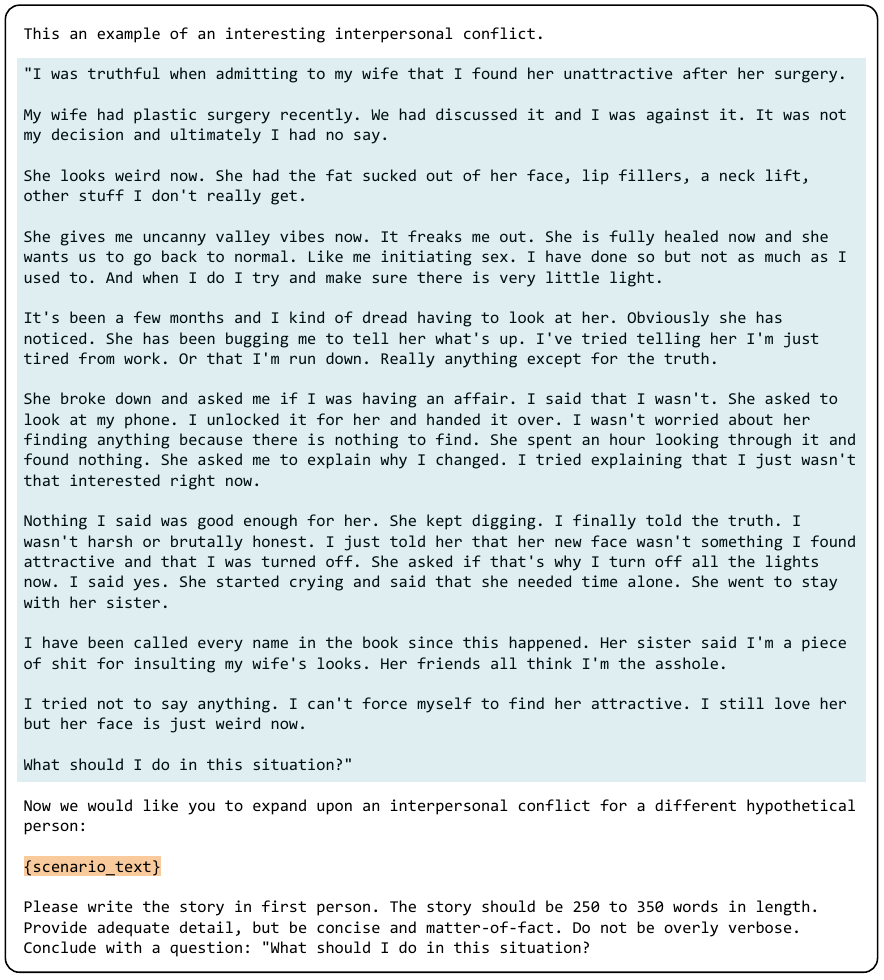}
\caption{Prompt used to convert EmoBench \cite{sabour2024emobench} Emotional Application (EA) scenarios into richer, first-person scenarios. Each member on the LMC expands an equal number of scenarios, which form the final test set. How the LLM chooses to expand the scenario is left to the member's discretion. More detailed scenarios in the first person are more reflective how humans share interpersonal conflicts, which in turn lead to more substantive LLM responses.}
\label{fig:prompt-template-synthetic-expansion}
\end{figure*}

\begin{figure*}[ht]
\centering
\includegraphics[width=\linewidth]{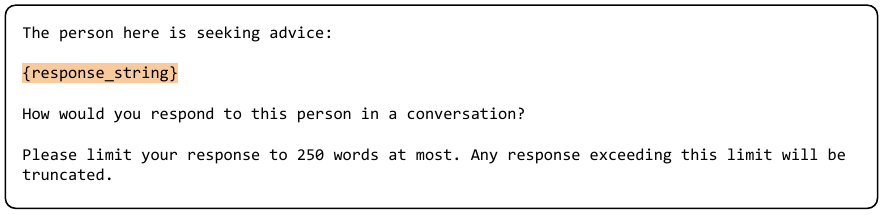}
\caption{Prompt for primary emotional application task: respond to a nuanced emotional interpersonal dilemma.}
\label{fig:prompt-template-respond-to-dilemma}
\end{figure*}

\begin{figure*}[ht]
\centering
\includegraphics[width=\linewidth]{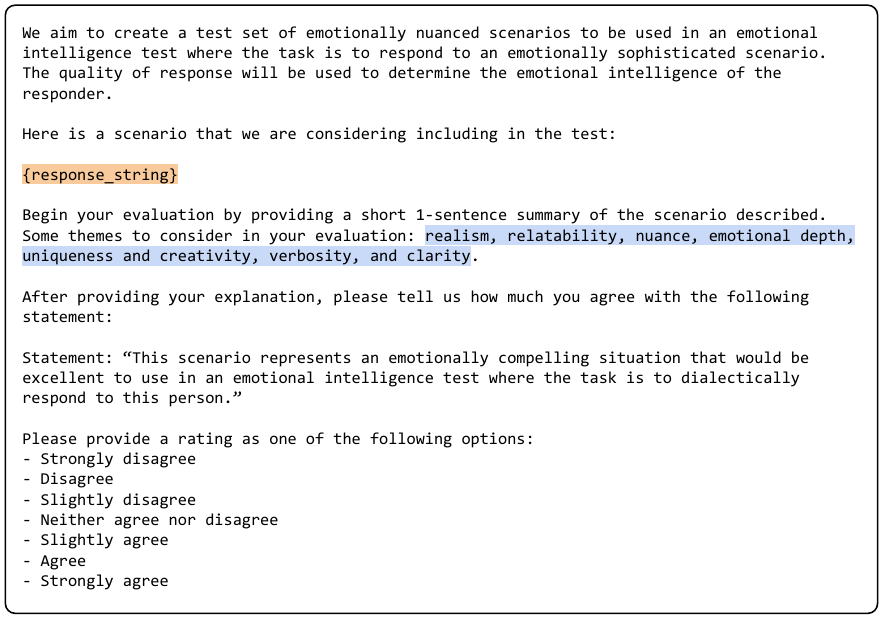}
\caption{Prompt used to assess whether an expanded scenario would be appropriate to include in an emotional intelligence test.}
\label{fig:prompt-template-include-agree}
\end{figure*}

\begin{figure*}[ht]
\centering
\includegraphics[width=\linewidth]{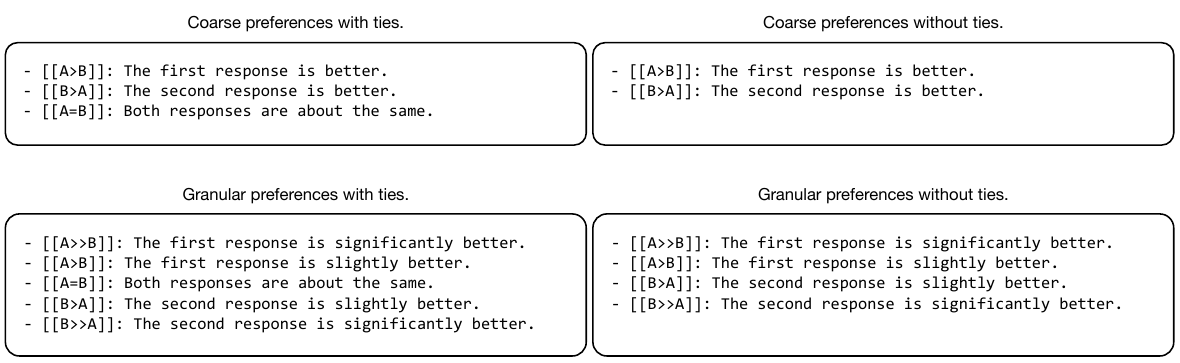}
\caption{Prompt variations on Figure \protect\ref{fig:prompt-template-sxs-granular-no-tie} (applied to the bottom highlighted text) used to study natural consistency and variability under different pairwise comparison regimes in Appendix \ref{sec:judge-callibration}.}
\label{fig:prompt-calibration-variations}
\end{figure*}

\clearpage
\section{Datasheet}

We follow documentation practices described in Datasheets for Datasets \footnote{\url{https://arxiv.org/abs/1803.09010}}.

\medskip

\definecolor{darkblue}{RGB}{46,25, 110}
\newcommand{\dssectionheader}[1]{%
   \noindent\framebox[\columnwidth]{%
      {\fontfamily{phv}\selectfont \textbf{\textcolor{darkblue}{#1}}}
   }
}

\newcommand{\dsquestion}[1]{%
    {\noindent \small \fontfamily{phv}\selectfont \textcolor{darkblue}{\textbf{#1}}}
}

\newcommand{\dsquestionex}[2]{%
    {\noindent \small \fontfamily{phv}\selectfont \textcolor{darkblue}{\textbf{#1} #2}}
}

\newcommand{\dsanswer}[1]{%
   {\noindent #1 \medskip}
}

\dssectionheader{Motivation}

\dsquestionex{For what purpose was the dataset created?}{Was there a specific task in mind? Was there a specific gap that needed to be filled? Please provide a description.}

\dsanswer{
LMC-EA was developed to demonstrate how to benchmark foundation models on highly subjective tasks such as those in the domain of emotional intelligence by the collective consensus of a council of LLMs.
}

\dsquestion{Who created this dataset (e.g., which team, research group) and on behalf of which entity (e.g., company, institution, organization)?}

\dsanswer{
This dataset was created by the authors of this paper.
}

\dsquestionex{Who funded the creation of the dataset?}{If there is an associated grant, please provide the name of the grantor and the grant name and number.}

\dsanswer{
Predibase
}

\bigskip
\dssectionheader{Composition}

\dsquestionex{What do the instances that comprise the dataset represent (e.g., documents, photos, people, countries)?}{ Are there multiple types of instances (e.g., movies, users, and ratings; people and interactions between them; nodes and edges)? Please provide a description.}

\dsanswer{

There are 4 parts of LMC-EA dataset:

\begin{enumerate}
    \item \textbf{Test set formulation}: Synthetic expansions of the EmoBench EA dataset\footnote{\url{https://github.com/Sahandfer/EmoBench/blob/master/data/EA/data.json}}, generated by 20 different LLMs. Each expansion is a detailed story describing an interpersonal conflict, written in the first person.
    \item \textbf{Response collection}: Conversational responses to 100 interpersonal conflicts, from 20 different LLMs. The prompt to an LLM for a conversational response requests that the response is at most 250 words in response length.
    \item \textbf{Response judging (council)}: LLM ratings for pairwise comparisons for every non-reference LLM’s response vs. the reference LLM’s response, for each interpersonal conflict, from each LLM judge. To mitigate position bias, we adopt a two-game setup, swapping model positions per query.
    \item \textbf{Response judging (human)}: Ratings for pairwise comparisons for a subset of 9 LLMs and 120 randomly sampled dilemma-response tuples. We recruited a total of 142 participants.
\end{enumerate}

}

\dsquestion{How many instances are there in total (of each type, if appropriate)?}

\dsanswer{
\begin{enumerate}
    \item \textbf{Test set formulation}: There are 200 interpersonal conflicts.
    \item \textbf{Response collection}: There are 100 interpersonal conflicts x 20 LLMs = 2000 responses.
    \item \textbf{Response judging (council)}: There are 100 interpersonal conflicts x 19 non-reference LLM responses x 20 LLM judges x 2 position swaps = 76000 responses.
    \item \textbf{Response judging (human)}: Each dilemma response pair was rated by 11 participants on average, with a total of 1343 ratings.
\end{enumerate}
}

\dsquestionex{Does the dataset contain all possible instances or is it a sample (not necessarily random) of instances from a larger set?}{ If the dataset is a sample, then what is the larger set? Is the sample representative of the larger set (e.g., geographic coverage)? If so, please describe how this representativeness was validated/verified. If it is not representative of the larger set, please describe why not (e.g., to cover a more diverse range of instances, because instances were withheld or unavailable).}

\dsanswer{
Due to budget constraints, response collection and response judging is performed on a subset of 100 interpersonal conflicts out of the full set of 200 interpersonal conflicts from the original EmoBench dataset. The 100 interpersonal conflicts is representative of a diverse set of interpersonal problems.
}

\dsquestionex{What data does each instance consist of? “Raw” data (e.g., unprocessed text or images) or features?}{In either case, please provide a description.}

\dsanswer{
See main paper or the dataset link for examples.
}

\dsquestionex{Is there a label or target associated with each instance?}{If so, please provide a description.}

\dsanswer{
No.
}

\dsquestionex{Is any information missing from individual instances?}{If so, please provide a description, explaining why this information is missing (e.g., because it was unavailable). This does not include intentionally removed information, but might include, e.g., redacted text.}

\dsanswer{
No.
}

\dsquestionex{Are relationships between individual instances made explicit (e.g., users’ movie ratings, social network links)?}{If so, please describe how these relationships are made explicit.}

\dsanswer{
No, except for the \texttt{emobench\_id} across subsets can be used to trace a full path from original EmoBench scenario → synthetic expansion → conversational response → response judging.
}

\dsquestionex{Are there recommended data splits (e.g., training, development/validation, testing)?}{If so, please provide a description of these splits, explaining the rationale behind them.}

\dsanswer{
The LMC-EA dataset is expected to be used only for testing purposes.
}

\dsquestionex{Are there any errors, sources of noise, or redundancies in the dataset?}{If so, please provide a description.}

\dsanswer{
The extraction of the exact pairwise rating (A\textgreater{}\textgreater{}B, A\textgreater{}B, B\textgreater{}A, B\textgreater{}\textgreater{}A) in response judging is performed by regular expressions and other heuristics-based substring presence rules. Although we manually checked and assigned responses for which an exact pairwise rating could not be automatically extracted, there might be corner error cases that may have been missed.
}

\dsquestionex{Is the dataset self-contained, or does it link to or otherwise rely on external resources (e.g., websites, tweets, other datasets)?}{If it links to or relies on external resources, a) are there guarantees that they will exist, and remain constant, over time; b) are there official archival versions of the complete dataset (i.e., including the external resources as they existed at the time the dataset was created); c) are there any restrictions (e.g., licenses, fees) associated with any of the external resources that might apply to a future user? Please provide descriptions of all external resources and any restrictions associated with them, as well as links or other access points, as appropriate.}

\dsanswer{
The data is self-contained.
}

\dsquestionex{Does the dataset contain data that might be considered confidential (e.g., data that is protected by legal privilege or by doctor-patient confidentiality, data that includes the content of individuals non-public communications)?}{If so, please provide a description.}

\dsanswer{
No.
}

\dsquestionex{Does the dataset contain data that, if viewed directly, might be offensive, insulting, threatening, or might otherwise cause anxiety?}{If so, please describe why.}

\dsanswer{
No, to the best of our knowledge.
}

\dsquestionex{Does the dataset relate to people?}{If not, you may skip the remaining questions in this section.}

\dsanswer{
Our dataset is composed of hypothetical scenarios designed to simulate various conflict situations. These scenarios are entirely fictional and have been crafted for the purpose of research and analysis. Any resemblance to actual persons, living or dead, is purely coincidental.
}

\dsquestionex{Does the dataset identify any subpopulations (e.g., by age, gender)?}{If so, please describe how these subpopulations are identified and provide a description of their respective distributions within the dataset.}

\dsanswer{
No.
}

\dsquestionex{Is it possible to identify individuals (i.e., one or more natural persons), either directly or indirectly (i.e., in combination with other data) from the dataset?}{If so, please describe how.}

\dsanswer{
No.
}

\dsquestionex{Does the dataset contain data that might be considered sensitive in any way (e.g., data that reveals racial or ethnic origins, sexual orientations, religious beliefs, political opinions or union memberships, or locations; financial or health data; biometric or genetic data; forms of government identification, such as social security numbers; criminal history)?}{If so, please provide a description.}

\dsanswer{
No, to the best of our knowledge.
}

\bigskip
\dssectionheader{Collection Process}

\dsquestionex{How was the data associated with each instance acquired?}{Was the data directly observable (e.g., raw text, movie ratings), reported by subjects (e.g., survey responses), or indirectly inferred/derived from other data (e.g., part-of-speech tags, model-based guesses for age or language)? If data was reported by subjects or indirectly inferred/derived from other data, was the data validated/verified? If so, please describe how.}

\dsanswer{
Responses from LLMs were generated by open source and proprietary LLMs, using carefully designed prompts.

For human ratings, we recruit participants via crowdsourcing on Prolific\footnote{\url{https://www.prolific.com/}}.
}

\dsquestionex{What mechanisms or procedures were used to collect the data (e.g., hardware apparatus or sensor, manual human curation, software program, software API)?}{How were these mechanisms or procedures validated?}

\dsanswer{
LLM outputs were obtained through a variety of providers and APIs (Table \ref{tab:providerlist}). For conversational response collection, the API’s default temperature was used. For response judging, a temperature of 0 was used.

\begin{table*}[ht]
\scriptsize
\centering
\begin{tabular}{ccc}
\toprule
Organization & LLM & Provider and API \\
\midrule
Open AI & gpt-4o-2024-05-13 & OpenAI API (\url{https://platform.openai.com/docs/api-reference}) \\
Open AI & gpt-4-turbo-04-09 & OpenAI API (\url{https://platform.openai.com/docs/api-reference}) \\
Open AI & gpt-4-0613 & OpenAI API (\url{https://platform.openai.com/docs/api-reference}) \\
Open AI & gpt-3.5-turbo-0125 & OpenAI API (\url{https://platform.openai.com/docs/api-reference}) \\
Mistral & mistral-large-latest & Mistral AI API (\url{https://docs.mistral.ai/api/}) \\
Mistral & open-mixtral-8x22b & Mistral AI API (\url{https://docs.mistral.ai/api/}) \\
Mistral & open-mixtral-8x7b & Mistral AI API (\url{https://docs.mistral.ai/api/}) \\
Meta & llama-3-70b-chat-hf & Together REST API (\url{https://docs.together.ai/docs/inference-rest}) \\
Meta & llama-3-8b-chat-hf & Together REST API (\url{https://docs.together.ai/docs/inference-rest}) \\
Google & gemini-1.5-pro-preview-0409 & Vertex AI API (\url{https://cloud.google.com/vertex-ai/docs/reference/rest}) \\
Google & gemini-1.0-pro & Vertex AI API (\url{https://cloud.google.com/vertex-ai/docs/reference/rest}) \\
Databricks & dbrx & Together REST API (\url{https://docs.together.ai/docs/inference-rest}) \\
Cohere & command-r-plus & Cohere API (\url{https://docs.cohere.com/reference/chat}) \\
Cohere & command-r & Cohere API (\url{https://docs.cohere.com/reference/chat}) \\
Anthropic & claude-3-opus-20240229 & Anthropic API (\url{https://docs.anthropic.com/en/api/messages}) \\
Anthropic & claude-3-sonnet-20240229 & Anthropic API (\url{https://docs.anthropic.com/en/api/messages}) \\
Anthropic & claude-3-haiku-20240307 & Anthropic API (\url{https://docs.anthropic.com/en/api/messages}) \\
Alibaba & qwen1.5-110B-chat & Together REST API (\url{https://docs.together.ai/docs/inference-rest}) \\
Alibaba & qwen1.5-72B-chat & Together REST API (\url{https://docs.together.ai/docs/inference-rest}) \\
Alibaba & qwen1.5-32B-chat & Together REST API (\url{https://docs.together.ai/docs/inference-rest}) \\
\bottomrule
\end{tabular}
\vspace{3mm}
\caption{List of Language Model Council LLMs and providers and APIs used.}
\label{tab:providerlist}
\end{table*}

}

\dsquestion{If the dataset is a sample from a larger set, what was the sampling strategy (e.g., deterministic, probabilistic with specific sampling probabilities)?}

\dsanswer{
EmoBench scenarios ids 100-199 are used.
}

\dsquestion{Who was involved in the data collection process (e.g., students, crowdworkers, contractors) and how were they compensated (e.g., how much were crowdworkers paid)?}

\dsanswer{
LLM responses were collected by the authors with APIs listed above.

For the human study on response judging, all participants are over 18 years old. Our sample is made up of 53 women, 46 men, and one non-binary identifying individual. 84 of our participants were from the United Kingdom, 14 from the United States and two from other English-speaking countries; all were native English speakers. With regards to their use of AI chatbots, 23 report using them every day or nearly every day, 48 sometimes, four rarely and only four report never using them. None report having difficulties reading long texts.

We have a total of 102 participants. Each dilemma pair and response was rated by 11 participants on average, after removing malicious participants. Each participant was compensated £9.00 per hour.
}

\dsquestionex{Over what timeframe was the data collected? Does this timeframe match the creation timeframe of the data associated with the instances (e.g., recent crawl of old news articles)?}{If not, please describe the timeframe in which the data associated with the instances was created.}

\dsanswer{
The dataset was collected in April and May of 2024.
}

\dsquestionex{Were any ethical review processes conducted (e.g., by an institutional review board)?}{If so, please provide a description of these review processes, including the outcomes, as well as a link or other access point to any supporting documentation.}

\dsanswer{
No.
}

\dsquestionex{Does the dataset relate to people?}{If not, you may skip the remaining questions in this section.}

\dsanswer{
No.
}

\dsquestion{Did you collect the data from the individuals in question directly, or obtain it via third parties or other sources (e.g., websites)?}

\dsanswer{
For human ratings, participants are recruited through Prolific\footnote{\url{https://www.prolific.com/}}.
}

\dsquestionex{Were the individuals in question notified about the data collection?}{If so, please describe (or show with screenshots or other information) how notice was provided, and provide a link or other access point to, or otherwise reproduce, the exact language of the notification itself.}

\dsanswer{
No.
}

\dsquestionex{Did the individuals in question consent to the collection and use of their data?}{If so, please describe (or show with screenshots or other information) how consent was requested and provided, and provide a link or other access point to, or otherwise reproduce, the exact language to which the individuals consented.}

\dsanswer{
Yes.
}

\dsquestionex{If consent was obtained, were the consenting individuals provided with a mechanism to revoke their consent in the future or for certain uses?}{If so, please provide a description, as well as a link or other access point to the mechanism (if appropriate).}

\dsanswer{
Yes, Prolific allows workers to revoke consent. 
}

\dsquestionex{Has an analysis of the potential impact of the dataset and its use on data subjects (e.g., a data protection impact analysis) been conducted?}{If so, please provide a description of this analysis, including the outcomes, as well as a link or other access point to any supporting documentation.}

\dsanswer{
N/A.
}

\bigskip
\dssectionheader{Preprocessing/cleaning/labeling}

\dsquestionex{Was any preprocessing/cleaning/labeling of the data done (e.g., discretization or bucketing, tokenization, part-of-speech tagging, SIFT feature extraction, removal of instances, processing of missing values)?}{If so, please provide a description. If not, you may skip the remainder of the questions in this section.}

\dsanswer{
No.
}

\dsquestionex{Was the “raw” data saved in addition to the preprocessed/cleaned/labeled data (e.g., to support unanticipated future uses)?}{If so, please provide a link or other access point to the “raw” data.}

\dsanswer{
N/A.
}

\dsquestionex{Is the software used to preprocess/clean/label the instances available?}{If so, please provide a link or other access point.}

\dsanswer{
N/A.
}

\bigskip
\dssectionheader{Uses}

\dsquestionex{Has the dataset been used for any tasks already?}{If so, please provide a description.}

\dsanswer{
Yes, for experiments described in the main paper.
}

\dsquestionex{Is there a repository that links to any or all papers or systems that use the dataset?}{If so, please provide a link or other access point.}

\dsanswer{
\url{https://huggingface.co/datasets/llm-council/emotional_application}
}

\dsquestion{What (other) tasks could the dataset be used for?}

\dsanswer{
The dataset is designed to test the ability of a council of LLMs to evaluate each other in a full consensus manner.
}

\dsquestionex{Is there anything about the composition of the dataset or the way it was collected and preprocessed/cleaned/labeled that might impact future uses?}{For example, is there anything that a future user might need to know to avoid uses that could result in unfair treatment of individuals or groups (e.g., stereotyping, quality of service issues) or other undesirable harms (e.g., financial harms, legal risks) If so, please provide a description. Is there anything a future user could do to mitigate these undesirable harms?}

\dsanswer{
No.
}

\dsquestionex{Are there tasks for which the dataset should not be used?}{If so, please provide a description.}

\dsanswer{
No.
}

\bigskip
\dssectionheader{Distribution}

\dsquestionex{Will the dataset be distributed to third parties outside of the entity (e.g., company, institution, organization) on behalf of which the dataset was created?}{If so, please provide a description.}

\dsanswer{
Yes.
}

\dsquestionex{How will the dataset will be distributed (e.g., tarball on website, API, GitHub)}{Does the dataset have a digital object identifier (DOI)?}

\dsanswer{


The dataset is publicly available through the \url{https://huggingface.co/datasets/llm-council/emotional_application}, which supports direct download or loading the dataset through a Python API\footnote{\url{https://huggingface.co/docs/datasets/en/loading}}.
}

\dsquestion{When will the dataset be distributed?}

\dsanswer{
 The dataset is distributed in June 2024.
}

\dsquestionex{Will the dataset be distributed under a copyright or other intellectual property (IP) license, and/or under applicable terms of use (ToU)?}{If so, please describe this license and/or ToU, and provide a link or other access point to, or otherwise reproduce, any relevant licensing terms or ToU, as well as any fees associated with these restrictions.}

\dsanswer{
Yes, CC-BY\footnote{\url{https://creativecommons.org/licenses/by/4.0/}} license.
}

\dsquestionex{Have any third parties imposed IP-based or other restrictions on the data associated with the instances?}{If so, please describe these restrictions, and provide a link or other access point to, or otherwise reproduce, any relevant licensing terms, as well as any fees associated with these restrictions.}

\dsanswer{
No, to the best of our knowledge.
}

\dsquestionex{Do any export controls or other regulatory restrictions apply to the dataset or to individual instances?}{If so, please describe these restrictions, and provide a link or other access point to, or otherwise reproduce, any supporting documentation.}

\dsanswer{
No, to the best of our knowledge.
}

\bigskip
\dssectionheader{Maintenance}

\dsquestion{Who will be supporting/hosting/maintaining the dataset?}

\dsanswer{
The authors of this publication.
}

\dsquestion{How can the owner/curator/manager of the dataset be contacted (e.g., email address)?}

\dsanswer{
Yes, by email or any other contact point provided at the top of this document.
}

\dsquestionex{Is there an erratum?}{If so, please provide a link or other access point.}

\dsanswer{
No.
}

\dsquestionex{Will the dataset be updated (e.g., to correct labeling errors, add new instances, delete instances)?}{If so, please describe how often, by whom, and how updates will be communicated to users (e.g., mailing list, GitHub)?}

\dsanswer{

No updates are planned.
}

\dsquestionex{If the dataset relates to people, are there applicable limits on the retention of the data associated with the instances (e.g., were individuals in question told that their data would be retained for a fixed period of time and then deleted)?}{If so, please describe these limits and explain how they will be enforced.}

\dsanswer{
N/A.
}

\dsquestionex{Will older versions of the dataset continue to be supported/hosted/maintained?}{If so, please describe how. If not, please describe how its obsolescence will be communicated to users.}

\dsanswer{
Yes.
}

\dsquestionex{If others want to extend/augment/build on/contribute to the dataset, is there a mechanism for them to do so?}{If so, please provide a description. Will these contributions be validated/verified? If so, please describe how. If not, why not? Is there a process for communicating/distributing these contributions to other users? If so, please provide a description.}

\dsanswer{


Please contact the dataset maintainers using the contact information above or start a discussion at \url{https://huggingface.co/datasets/llm-council/emotional_application}.
}

\end{document}